\documentclass[twoside,10pt]{article}

\usepackage{jmlr2e}
\usepackage{a4wide}
\usepackage{booktabs}
\usepackage{array}

\usepackage{tikz}

\usetikzlibrary{trees,arrows.meta,positioning,fit,backgrounds}

\usepackage{mathpazo}
\usepackage[T1]{fontenc}

\ShortHeadings{Review of Extreme Classification}{}
\firstpageno{1}

\usepackage{subfiles}
\usepackage{amsmath}
\usepackage{graphicx}
\usepackage{graphicx}
\usepackage{paralist}
\usepackage{wrapfig}

\usepackage{multirow}
\usepackage{mathtools}

\usepackage{amssymb}

\usepackage[utf8]{inputenc}

\usepackage[algo2e,ruled,vlined,linesnumbered]{algorithm2e}

\usepackage{cancel}

\usepackage{algorithmicx}

\usepackage{algorithm}
\usepackage{comment}
\usepackage{tabularx}
\usepackage{nomencl}
\usepackage{subcaption}

\usepackage{color}   
\usepackage{hyperref}

\hypersetup{
    colorlinks=true, 
    linktoc=all,     
    linkcolor=blue,  
}

\makenomenclature

\mathtoolsset{showonlyrefs=true}

\newcommand{\R}{\mathbb{R}}

\begin{document}

\title{Review of Extreme Multi-label Classification}

\author{\name Arpan Dasgupta \email         
      arpan.dasgupta@research.iiit.ac.in \\
      \addr CSTAR, 
      IIIT, Hyderabad, India
      \AND
      \name Preeti Lamba \email         
      preeti.preeti@research.iiit.ac.in \\
      \addr CSTAR, 
      IIIT, Hyderabad, India
      \AND
      \name Ankita Kushwaha \email         
      ankita.kushwaha@research.iiit.ac.in \\
      \addr CSTAR, 
      IIIT, Hyderabad, India
      \AND
      \name Kiran Ravish \email         
      kiran.ravish@research.iiit.ac.in \\
      \addr CSTAR, 
      IIIT, Hyderabad, India
      \AND
      \name Siddhant Katyan \email siddhantkatyan@gmail.com \\
      \addr CSTAR, 
      IIIT, Hyderabad, India
      \AND
       \name Shrutimoy Das \email shrutimoy@gmail.com \\
      \addr CSTAR, 
      IIIT, Hyderabad, India
      \AND
       \name Pawan Kumar \email pawan.kumar@iiit.ac.in \\
      \addr CSTAR,
      IIIT, Hyderabad, India}

\editor{None}

\maketitle

\begin{abstract}
Extreme multi-label classification or XMLC, is an active area of interest in machine learning. Compared to traditional multi-label classification, here the number of labels is extremely large, hence, the name extreme multi-label classification. Using classical one-versus-all classification does not scale in this case due to large number of labels; the same is true for any other classifier. Embedding labels and features into a lower-dimensional space is a common first step in many XMLC methods.
Moreover, other issues include existence of head and tail labels, where tail labels are those that occur in a relatively small number of samples. The existence of tail labels creates issues during embedding. This area has invited application of wide range of approaches ranging from bit compression motivated from compressed sensing, tree based embeddings, deep learning based latent space embedding including using attention weights, linear algebra based embeddings such as SVD, clustering, hashing, to name a few. The community has come up with a useful set of metrics to identify correctly the prediction for head or tail labels.   
\end{abstract}

\begin{keywords}
  extreme classification, head and tail labels, compressed sensing, deep learning, attention
\end{keywords}

\section{Introduction}

\paragraph{}
Extreme multi-label classification (XMLC or XML) is an active area of research. It is multi-label classification problem, where the number of labels is very large going sometimes up to millions. The traditional classifiers such as one-vs-all, SVM, neural networks, etc, cannot be applied directly due to two major reasons. Firstly, the large number of labels creates a major bottleneck as it is not possible to have a simple classifier for each label due to memory constraints. Secondly, the presence of some labels which have very few samples in their support make learning about these labels a challenge. These labels are called {\em tail} labels, and their existence is a major hurdle to achieving good multi-label classification accuracy, consequently, many methods specifically address tail-label prediction. 

\paragraph{}
Extreme multi-label classification (XMLC) problems arise in many settings—for example, assigning tags to a Wikipedia article from its title or full text, recommending ``frequently bought together'' items based on a product’s name or description, and choosing relevant advertisement keywords based on an ad’s text. In all such cases, automated extreme classification is needed because manual labeling is impractical. Because all of these tasks must run in real time, memory and latency constraints are critical when designing XMLC algorithms. Table~\ref{tab:notations} lists common notations used in this paper.



\footnotesize
\begin{table}[t!]
\small 
\centering 
\caption{\label{tab:notations}Table of Common Notations}
\begin{tabular}{|c|l|} 
\hline 
Notation & Meaning \\ 
\hline 
{$\mathbb{R}$} & {The set of real numbers} \\
{$\text{round}(\cdot)$} & {Rounds the argument to 0 or 1} \\
{$\lVert\,{\cdot}\,\rVert_F$} & {Frobenius norm of the argument} \\
{$\lVert\,{\cdot}\,\rVert_\text{tr}$} & {Trace norm of the argument} \\
{$\lVert\,{\cdot}\,\rVert$} & {Norm of the argument} \\ 
{$I$} & {Identity matrix of appropriate dimension}  \\ 
{$I_k$} & {Matrix containing the first $k$ columns of the identity matrix} \\
{$X^{\dagger}$} & {Pseudo-inverse of $X$} \\ 
{$\text{tr}(A)$} & {Trace of the matrix $A$}  \\ 
{$A^T$} & {Transpose of the matrix $A$}   \\ 
{$A^{-1}$} & {Inverse of the matrix $A$}   \\ 
{$\text{ln}(\cdot)$} & {Natural logarithm of the argument}  \\ 
{$\phi$} & {The empty set}  \\ 
{$\nabla_x f$} & {Jacobian of $f$ with respect to $x$}  \\ 
{$\nabla_x^2f$} & {Hessian of $f$ with respect to $x$}  \\ 
{$V_{1:k}$} & {First $k$ columns of matrix $V$}  \\
{$\text{rank}(A)$} & {Rank of matrix $A$}  \\ 
{$\text{card}(S)$} & {Cardinality of the set $S$}  \\
{$\text{vec}(A)$} & {Stack the columns of matrix $A$ to get a vector} \\ 
{$\text{nnz}(A)$} & {Number of non zero entries in the matrix $A$}  \\ 
{$\mathbb{E}[X]$} & {Expected value of the random variable $X$}  \\ 
{$\text{AUPRC}$} & {Area under the precision-recall curve}  \\ 
{$\text{i.i.d}$} & {Independent and identically distributed}  \\
{$\text{CG}$} & {Conjugate Gradient Method} \\ 
\hline 
\end{tabular}
\end{table}
\normalsize

\section{Problem Formulation and Important Definitions}
\paragraph{}
Though different papers in XMLC use different notation, recent papers consistently use the following notation which will be followed in this review paper too. The features are represented by the matrix $X \in \mathbb{R}^{n\times d},$ where $n$ is the number of samples or instances, and $d$ represents the dimension of the features. The label matrix is represented by $Y \in \{0,1\}^{n \times L},$ where $L$ represents the total number of labels. Here $y_{il} = 1$ represents that the $i-$th sample contains $l-$th label as ground truth.   Individual features and labels for a sample are represented by $x_i$ and $y_i,$ respectively, where $i$ represents the sample in consideration. The complete dataset is hence represented by $D = \{x_i,y_i\}_{i=1}^n$.

Some other notation that appear several times are:
\begin{compactenum}
    \item $k$ which generally defines the dimension of the embedded space or the number of children in a tree.
    \item $h$ which generally depicts a linear learner.
    \item $P$ which represents the compression matrix for mapping features from the original or the embedded space. 
\end{compactenum}

These terms have been kept consistent throughout the paper, but have been redefined in the context if it conflicts with a previous definition.

\subsection{Ranking}
The problem of extreme classification requires the model to predict a set of labels, which will be relevant to the given test sample. However, most models do not produce binary results; instead, they provide a ranking of predicted labels by relevance score. The exact labels, if required, can be selected using a thresholding based on the confidence. Thus, XMLC can be conceived of a ranking based problem instead of a classification problem. Even the most common metrics used in XMLC expect a ranking based output from the algorithm. We define those metrics below.

\subsection{Head and Tail Labels}
\begin{figure}[t!]
    \centering
    \includegraphics[width=110mm]{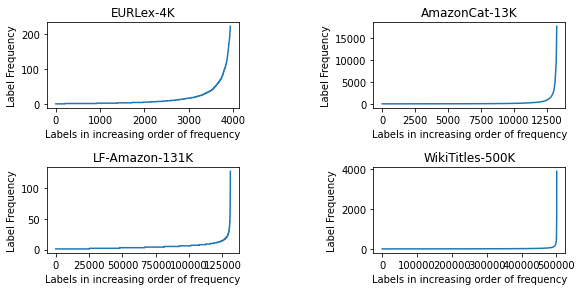}
    \caption{\label{fig:label-frequency}The tail label distribution of $4$ popular XMLC datasets. The tail labels have very less frequency as compared to the most frequent (head) labels. Further, this imbalance grows with an increasing number of labels in the dataset.}

\end{figure}

\paragraph{}
The number of samples are not evenly distributed among the labels in a typical XMLC problem. For example, in the Wiki-500K dataset \citep{Bhatia16}, 98\% of labels have less than 100 training instances. Thus, some of the labels have enough data for the models to learn while most do not have enough data. These are called head and tail labels respectively. In Figure~\ref{fig:label-frequency}, we show plots of label frequency for various datasets. Most models trained directly without keeping the data distribution in mind are likely to develop an implicit bias towards the head labels. Thus, tail-label based metrics are a useful tool for measuring the effectiveness of a model. We discuss some of these metrics below. 

{}

\section{Datasets and Metrics}

\subsection{Datasets}
\paragraph{}
There are different kinds of datasets which are used for bench-marking XMLC results. One set of datasets is obtained from Amazon reviews and titles scraped from internet archives. The reviews, titles and product summary are used for prediction of the correct product tags \citep{mcauley2013hidden}, \citep{mcauley2015image} and \citep{mcauley2015inferring}. Wikipedia based datasets are used for prediction of tags on Wikipedia articles using the article or just the titles \citep{zubiaga2012enhancing}. The dataset EurLex formulates large scale multi-label classification problems for European legislature legal documents \citep{loza2008efficient}. Other proprietary datasets such as that of advertisement bids on bing are used by some methods. For a comprehensive list of dataset, please refer to the XMLC repository \cite{Bhatia16}.

\subsection{Metrics}
\paragraph{}
The following metrics are defined on the predicted score vector $\hat{y} \in R^L$ and ground truth label vector $y \in \{0,1\}^L.$

\subsubsection{Precision@k}
The most commonly used metric for measuring the performance of XMLC algorithms is P@$k$. Here P@$k$ measures what fraction of the top-$k$ predicted labels are present in the actual set of positive labels for the test data point. Formally, P@$k$ is represented by \\ 
$$
P@k := \frac{1}{k} \mathlarger{\sum_{l \in \text{rank}_k(\hat{y})}} y_l.
$$

\subsubsection{DCG@k and nDCG@k}
Discounted Cumulative Gain (DCG) and Normalized Discounted Cumulative Gain (nDCG) are common metrics for measuring the performance of a ranking system. Here DCG@$k$ measures how much the ground truth scores of each of the top $k$ labels predicted by the algorithm add up to. A log term for the positions is added to ensure that better ordering is rewarded. Here nDCG@$k$ is a modified version of the same metric which makes sure that the final score is well bounded for better comparison. 
\begin{align*}
\text{DCG@k} := \mathlarger{\sum_{l \in \text{rank}_k(\hat{y})}} \frac{y_l}{\log (l+1)}, \quad \quad 
\text{nDCG@k} := \frac{\text{DCG}@k}{\sum_{l=1}^{\min(k,||y||_0)}\frac{1}{\log (l+1)}}.
\end{align*}

\subsubsection{Propensity Based Metrics}
\cite{jain2016extreme} proposes a version of all the above metrics which take into account the fact that tail label predictions are measured well. The value $p_l$ represents the propensity score of label $l$ which ensures that the label bias is removed. This prevents models from achieving a high score from just correctly predicting head labels and completely ignoring tail labels. The corresponding versions of the metrics P@$k$, DCG@$k$ and nDCG@$k$ are called PSP@$k$, PSDCG@$k$ and PSnDCG@$k$.

\begin{align*}
\text{PSP}@k &:= \frac{1}{k} \mathlarger{\sum_{l \in \text{rank}_k(\hat{y})}} \frac{y_l}{p_l}. \\
\text{PSDCG}@k &:= \mathlarger{\sum_{l \in \text{rank}_k(\hat{y})}} \frac{y_l}{p_l \log (l+1)}. \\
\text{PSnDCG}@k &:= \frac{\text{PSDCG}@k}{\sum_{l=1}^{\min(k,||y||_0)}\frac{1}{\log (l+1)}}.
\end{align*}

\subsubsection{Macro Metrics}
Propensity score metric is used as performance measure for tail labels. These metrics assign more weight to tail labels to counterbalance their rarity. However, they were primarily developed to handle missing labels rather than focusing explicitly on the long-tail problem, thus not entirely addressing the issue. Macro-averaged metrics and coverage measures are more suitable for evaluating XMLC in the context of long-tail distributions. Macro-averaging treats all labels ``equally'', ensuring that labels with fewer positive examples are not ignored.

\paragraph{}
The paper \cite{schultheis2024generalized} presents methods for optimizing performance measures for tail labels in extreme multi-label classification (XMLC). Assume that we are given a known set of $n$ instances $X=[x_1,\ldots, x_n]^T $ with unknown labels, on which we have to make predictions. Our goal is to assign each instance $x_i$ exactly $k$(out of $L$) labels represented as a $k$-hot vector $\hat{y}_i=\{ y \in \{0,1\}^L=\mathcal{Y}:||y||_1=k\}$ and $Y$ denotes the entire label matrix for $X$. It introduces a multi-label confusion tensor $C(Y,\hat{Y})=[C(y_{:1},\hat{y}_{:1}), \ldots, C(y_{:m},\hat{y}_{:m})] $, here $y_{ij}$ denotes entries of $Y$ and $y_i$ denote rows while $y_{:j}$ denote columns of matrix $Y$, that aggregates binary confusion matrices for all labels, allowing for a structured evaluation of multi-label performance. The paper defines instance-wise weighted utility functions $u_w:\mathcal{Y} \times \mathcal{Y} \rightarrow \mathbb{R}_{\geq 0}$ that assign different weights to various prediction outcomes,\[u_w(\mathbf{y}, \hat{\mathbf{y}}) = \sum_{j=1}^{m} w_{00}^{j} (1 - y_j) (1 - \hat{y}_j) + w_{01}^{j} (1 - y_j) \hat{y}_j + w_{10}^{j} y_j (1 - \hat{y}_j) + w_{11}^{j} y_j \hat{y}_j\] enabling the implementation of measures like precision@k and propensity-scoring. Here, $ w_{00}^{j},  w_{01}^{j},$ $w_{10}^{j},\text{ and } w_{11}^{j}$ to express the utility of true negatives, false positives, false negatives, and true positives, respectively.

\paragraph{}
A significant contribution is the Expected True Utility (ETU) approximation, which simplifies the optimization of utility functions $\Psi(Y, \hat{Y}) = \sum_{i=1}^{n} u_w(y_i, \hat{y}_i)$ by parameterizing confusion matrix entries: 
\[
\Psi(Y, \hat{Y}) = \sum_{j=1}^{m} \left( w_{00}^j c_{00}^j + w_{01}^j c_{01}^j + w_{10}^j c_{10}^j + w_{11}^j c_{11}^j \right).
\] 
This approximation is particularly useful for linear measures and reduces computational complexity. The instance-wise weighted utility functions $u_w$ yields precision@$k$ by taking  $w_{11}^j=1$  and  $ w_{00}^{j},  w_{01}^{j}, w_{10}^{j}$ as $0$. Similarly other metrics can be generated by changing weights. For non-linear macro-measures such as the F-measure, the paper proposes a block coordinate ascent (BCA) algorithm.  It starts with random predictions and uses a utility function to compute gains for each label, selecting the top-k labels that maximize the utility. This process is repeated for each instance until the overall utility improvement falls below a specified threshold, ensuring locally optimal predictions.

\paragraph{}
The paper also addresses the optimization of coverage, a non-linear measure, by reformulating the ETU specifically for this purpose. This reformulation allows for a more efficient direct optimization method compared to the general BCA approach.  The coverage utility can be reformulated as:
    \[
    \Psi_{\text{ETU}}(\hat{Y}) = \mathbb{E}_{Y \mid X} \left[ m^{-1}\sum_{j=1}^{m-1} 1[t_j > 0] \right] = 1 - m^{-1}\sum_{j=1}^{m-1} \prod_{i=1}^{n} \left( 1 - \eta_j(x_i) \hat{y}_{ij} \right),
    \]
    where \( t_j \) is the number of true positives for label \( j \), and \( \hat{y}_{ij} \) is the predicted value of label \( j \) for instance \( i \).

\paragraph{}
Overall, the paper provides a unified framework and efficient algorithms for optimizing XMLC performance, especially for tail labels. These methods are crucial for applications where rare labels are important, such as recommendation systems and medical diagnosis, ensuring more reliable and effective predictions for these critical labels.

\section{Review}

\paragraph{}
There are several categories of methods for performing extreme classification. We have broadly divided them into four categories based on the basic philosophy of the algorithm. In Fig. \ref{fig:xmlc-taxonomy}, we show taxonomy of key papers the methods in XMLC. We do not claim to include all papers, some additional papers are cited in the text, but not mentioned in figure.

\begin{figure}[t!]
\vspace{-4cm}
\centering

\scalebox{0.60}{\begin{tikzpicture}[
  baseline=(current bounding box.center),
  scale = 0.75,
  grow  = right,
  every node/.style = {font=\small, rounded corners,
                       draw, fill=gray!10, inner sep=3pt},
  level 1/.style = {level distance=70mm, sibling distance=75mm},
  level 2/.style = {
      level distance = 45mm,
      sibling distance = 11mm,
      parent anchor = east,
      child  anchor = west,
      anchor = west,
      align = left
  },
  edge from parent/.style = {
      -{Stealth[length=2.5mm]}, very thick, draw
  }
]

\node[font=\small, fill=gray!20] {XMLC}
    child { node {Tree-like}
          child { node {FastXML'14 \cite{FastXML}} }
          child { node {Parabel'18 \cite{prabhu2018parabel},CRAFTML'18 \cite{craftml_siblini} \\ DXML \cite{kumar2021DXML}} }
          child { node {Bonsai'20 \cite{khandagale2020bonsai},XReg'20 \cite{xreg_prabhu} } }}
    child { node {Linear-algebra}
          child { node {MLCSSP'13 \cite{cssp_bi}} }
          child { node {LEML'14 \cite{leml_yu},REML’14 \cite{reml_chang}} }
          child { node {SLEEC'15 \cite{sleec_bhatia}} }
          child { node {PD-SPARSE'16 \cite{pdsparse_ian},REML’16} }
          child { node {Annex'17 \cite{AnnexML}} }
          child { node {DEFRAG'19 \cite{defrag_jalan}} }
          child { node {ReimannXML'21 \cite{naram22ReimannXML}} }
          child { node {MLFM'23 \cite{Pavlovski2023} } }}
    child { node {Multimodal}
        child { node {MUFIN'22 \cite{mittal2022multi}} }}
    child { node {Other Deep-learning}
          child { node {XML-CNN'17 \cite{XML_CNN}} }
          child { node {AttXML'19 \cite{you2019attentionxml}} }
          child { node {DECAF'21 \cite{mittal2021decaf},GalaXC'21 \cite{saini2021galaxc} \\ ECLARE’21 \cite{mittal2021eclare}}} 
          child { node {BoostXML'22 \cite{boostxml},NGAME'23 \cite{dahiya2023ngame} } } 
          child { node {LightDXML'23 \cite{lightDXML}} } 
          }
    child { node {Compressed-sensing}
          child { node {CS'09} }
          child { node {PLST'12,CPLST'12 \cite{cplst_chen}} }
          child { node {R-BF'13} }
          child { node {LTLS'16 \cite{jasinska2016log}} }
          child { node {MACH'19 \cite{mach_medini} } }}
    child { node {One-vs-All}
        child { node {DISMEC'17 \cite{babbar2017dismec} }}
        child { node {SLICE'19 \cite{jain2019slice}} }}
    child { node {Transformer/LLM-assisted}
          child { node {X-Trans'20 \cite{chang2020taming}} }
          child { node {LightXML'21 \cite{LightXML},XR-Trans'21 \cite{zhang2021fast}} }
          child { node {ASTEC'21 \cite{dahiya2021deepxml},SiamXML’21 \cite{pmlr-v139-dahiya21a}} }
          child { node {PINA'23 \cite{chien2023pina}} }
          child { node {MatchXML'24 \cite{matchxml}} }};
\end{tikzpicture}}
\caption{Taxonomy of representative extreme multi-label classification (XMLC) methods. The XMLC methods are boradly classified into 6 classes. Due to popularity of Transformer/LLM based embeddings, we kept them as separate class.}
\label{fig:xmlc-taxonomy}
\end{figure}

\clearpage

\subsection{Compressed-Sensing Based Methods}
\begin{figure}[t!]
    \centering
    \includegraphics[width=70mm]{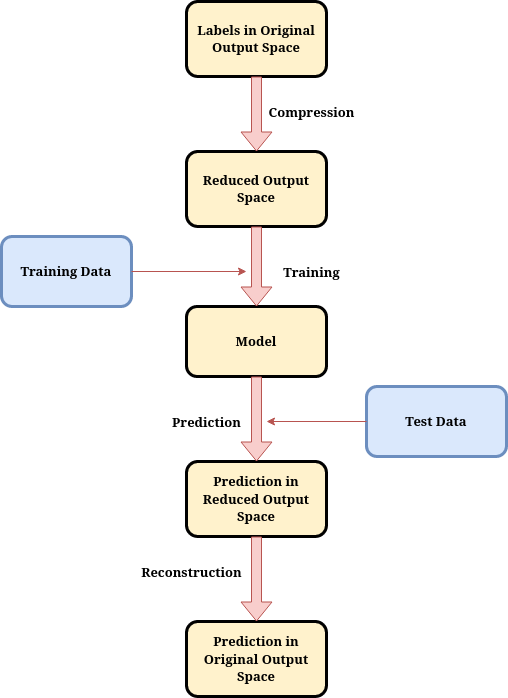}
    \caption{General flow of CS based methods. Workflow of compressed-sensing (CS) methods for extreme multi-label classification: (i) the original, high-dimensional label vectors are first linearly compressed into a compact code space, producing a reduced output space; (ii) using the training instances, a predictor is learned that maps input features to codes in this reduced space, so only the small set of compressed targets is seen during optimisation; (iii) at test time the model outputs a code for each unseen instance; and (iv) a sparse-recovery or learned decoder reconstructs that code back to the full label space, yielding the final predictions—thus the three CS stages, compression $\rightarrow$ learning $\rightarrow$ reconstruction, make extreme-scale label prediction tractable while clearly marking where training data and test data enter the pipeline.}
\end{figure}{}

\paragraph{}
The methods in this category are based on the concept of compressing the label space into a smaller, more manageable embedding space. The idea behind this method comes from signal-processing, where the compressed sensing technique can be used for signal reconstruction from much fewer samples than required, due to the sparsity. In the extreme multi-label scenario, the idea is to recover the original labels from predictions in a smaller label space. Again, The compression operation is valid due to the sparsity of the original label space.

\noindent Broadly, there are 3 steps in this procedure: 
\begin{enumerate}
    \item \textbf{Compression:} The original label space is compressed into a smaller vector space. This method of compression can use a simple linear transformation using an orthogonal matrix, hashing functions, clustering based approaches etc. 
    \item \textbf{Learning:} Since after compression, the output space is much smaller, methods like binary relevance (predicting each element in the output space individually using a binary classifier) become viable.
    \item \textbf{Reconstruction:} During prediction, the output in the compressed space must be converted back to the original label space. This can be done using methods such as solving optimization algorithms, using inverse projection matrices, subset prediction algorithms etc.
\end{enumerate}


\paragraph{}
The paper \cite{cs_hsu} was the first to make use of this technique. The application of compressed sensing in the XMLC problem was motivated from the observation that even though the label space of multi-label classification may be very high dimensional, the label vector for a given sample is often sparse. This sparsity of a label vector is be referred to as the \textit{output sparsity}. The proposed method utilizes the sparsity of $\mathbb{E}[y|x]$ rather than that of $y$. $y$ may be sparse but $\mathbb{E}[y|x]$ may have a large support. This may happen if there are many similar labels. 


\noindent The proposed method proceeds in three steps. First, it compresses $\{(x_i,y_i)\}$ to $\{(x_i,h_i)\},$ where $g_i = P y_i \in R^k$ using a random compression matrix $P \in \mathbb{R}^{k \times L}$ where $k$ is determined by the required sparsity level $s$. It is ensured that $k$ is logarithmic with respect to $L$. Next, for $j = 1 : k$, a  function $h_j(x)$ is learnt to predict $g_i[j]$ for every sample $i$. Finally, during prediction, for an input vector $x$, compute $H(x) = [h_1(x), \dots, h_k(x)].$ Then solve an optimization problem for finding a $s$-sparse vector $\Tilde{y}$ such that $P.\Tilde{y}$ is closest to $H(x)$ by some pre-defined metric.



\paragraph{}
The goal of the problem is to learn a predictor $F : X \rightarrow Y$ such that the error $\mathbb{E}_x ||F(x) - \mathbb{E}(y|x)||_{2}^{2}$ is minimized. The label space dimension $L$ is very large but $\mathbb{E}(y|x)$ of the corresponding label vector $y \in Y$ is $s$-sparse. Given a sample $\{(x_i,y_i)\}_{i=1}^{n},$ we obtain a compressed sample $\{(x_i,Py_i)\}_{i=1}^{n}$ and then learn a predictor $H$ with the objective of minimizing the error $\mathbb{E}_{x}||H(x) - \mathbb{E}(Py|x)||_{2}^{2}.$ Then, the prediction $\mathbb{E}(y|x)$ can be obtained by composing the predictor $H$ of $\mathbb{E}(Py|x)$ with a reconstruction algorithm $R : \mathbb{R}^{K} \rightarrow \mathbb{R}^{L}.$ The algorithm $R$ maps the predictions of compressed labels $h \in \mathbb{R}^{k}$ to predictions of $y \in Y$ in the original output space. This mapping is done by finding a sparse vector $\Tilde{y}$ such that $P\Tilde{y}$ closely approximates the compressed labels $h.$

\paragraph{}
The compression step can be ensured to give a close approximation of the original feature vector if the compression matrix $P$ maintains certain isometry properties as explained in the paper. For the reconstruction algorithm, a greedy sparse compressed sensing reconstruction algorithm called Orthogonal Matching Pursuit (OMP) is used, first proposed in \cite{pati1993orthogonal}.

\begin{wrapfigure}{l}{6.5cm}
\includegraphics[scale=0.6]{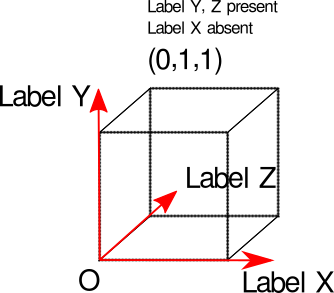}
\caption{\label{fig:hypercube_example}An example of the hypercube representation. Each axis represents whether a specific label is present. The entire label space is represented by the set of vertices.}
\end{wrapfigure} 

\paragraph{}
The previously stated method reduces the feature space significantly, enabling the method to train efficiently on large number of labels, but solving an optimization problem using OMP everytime during prediction is expensive. In the paper \cite{plst_tai}, a method is proposed which performs label space reduction efficiently and handles fast reconstruction. An SVD based approach is adopted for finding an orthogonal projection matrix which captures the correlation between labels. This allows easier reconstruction using the transpose of the projection matrix. The paper also uses a hypercube to model the label space of multi-label classification. It shows that algorithms such as binary relevance (BR, also called one-vs-all), compressive sensing (CS) can also be derived from the hypercube view. 

The hypercube view aims to represent the label set $\mathcal{Y}$ of a given sample using a vector $y \in \{0,1\}^L,$ where $\ell \in y$ iff $\ell^{th}$ component of $y$ is 1. As shown in Figure \ref{fig:hypercube_example} (with $L=3$), each vertex of an $L$-dimensional hypercube represents a label set $\mathcal{Y}$. Each component of $y$ corresponds to one axis of a hypercube, which indicates the presence or absence of a label $\ell$ in $\mathcal{Y}.$





\paragraph{}
One vs All classification (also called Binary Relevance) and the Compressed Sensing Methods discussed can be formulated to be operations on this hypercube. The Binary Relevance method can be interpreted as projecting the hypercube into each of the $L$ dimensions before predicting. Using the hypercube view, each iteration of CS can be thought of as projecting the vertices to a random direction before training. Since $K \ll L,$ this new subspace is much smaller than the original space that the hypercube belongs to.


\paragraph{}
Using the hypercube view, each iteration of CS can be thought of as projecting the vertices to a random direction before training. Since $K \ll L,$ this new subspace is much smaller than the original space that the hypercube belongs to. Thus, the label-set sparsity assumption, which  implies that only a small number of vertices in the original hypercube are of relevance for the multi-label classification task, allows CS to work on a small subspace.


\paragraph{}
A framework called Linear Label Space Transformation (LLST) is proposed by the paper, which focuses on a $K$-dimensional subspace of $\mathbb{R}^L$ instead of the whole hypercube. Each vertex $v$ of the hypercube is encoded to the point $h$ in the $K$-dimensional space by projection. Then, a multi-dimensional regressor $r(x)$ is trained to predict $h$. Then, LLST maps $r(x)$ back to a vertex of the hypercube in $\mathbb{R}^L$ using some decoder $D$. LLST basically gives a more formal definition to compressed sensing based methods in the XMLC context, although it assumes that the compression step must be linear.

\begin{algorithm}
\caption{Linear Label Space Transformation (LLST)}
\begin{algorithmic}[1]
    \State Encode $\{(x_n,y_n)\}$ to $\{(x_n,h_n)\}$ where $h_n = Py_n$ is a point on the $K$-dimensional space obtained using a projection matrix $P \in \mathbb{R}^{K \times L}.$
    \State For $k=1:K,$ learn a function $r_k(x)$ from $\{(x_n,h_n[k])\}_{n=1}^{N}.$
    \State For each input vector $x,$ compute $r(x) = [r_1(x),r_2(x),\cdots, r_K(x)].$
    \State Return $D(r(x))$ where $D:\mathbb{R}^K \rightarrow \{0,1\}^L$ is a decoding function from the $K$-dimensional subspace to the hypercube.
\end{algorithmic}
\end{algorithm}

The choice of the decoder $D$ depends on the choice of projection $P$. For BR, we can simply take $P=I$ as the projection, and $D$ as the component-wise round(.) function. For CS, the projection matrix, $P$, is chosen randomly from an appropriate distribution and $D$, the reconstruction algorithm involves solving an optimization problem for each different $x$.
\newline


\paragraph{}
Taking advantage of the hypercube sparsity in large multi-label classification datasets, we can take $K \ll \ell$ for the LLST algorithm, which in turn will reduce the computational cost. The proposed \textit{Principle Label Space Transformation (PLST)} approach seeks to find the projection matrix $P$ and the decoder $D$ for such an $K$-dimensional subspace through SVD. This method is described below.

\paragraph{}
The label sets of the given examples are stacked together to form a matrix $Y \in \mathbb{R}^{L \times N}$ such that each column of the matrix $Y$ is one of the occupied vertices of the hypercube. The matrix $Y$ is then decomposed using SVD such that

\begin{equation}\label{eqn:svd_plst}
        Y = U \Sigma V^T \Rightarrow U^TY = \Sigma V^T \text{ (since } UU^T = I).
\end{equation}
 Here, $U \in \mathbb{R}^{L \times L}$ is a unitary matrix whose columns form a basis for $Y$, $\Sigma \in \mathbb{R}^{L \times N}$ is a diagonal matrix containing the singular values of $Y$ and $V \in \mathbb{R}^{N \times N}$ is also a unitary matrix.
Assume that the singular values are ordered $\sigma_1 \geq \sigma_2 \geq \cdots \geq \sigma_L.$

\paragraph{}
The second line in \eqref{eqn:svd_plst} can be seen as a projection of $Y$ using the projection matrix $U^T.$ If we consider the singular vectors in $U$ corresponding to the $K$ largest singular values, we obtain a smaller projection matrix $P = U_{K}^{T} = [u_1 \: u_2 \: \cdots \: u_K]^T$ that maps the vertices $y$ to $\mathbb{R}^K.$ The projection matrix $P$ using the principle directions guarantees the minimum  encoding error from $\mathbb{R}^L$ to $\mathbb{R}^K.$ Because  $P = U_K^T$ is an orthogonal matrix, we have $P^{-1} = P^T.$ Thus, $U_K$ can be used as a decoder $D$ to map any vector $r \in \mathbb{R}^{K}$ to $U_K .r \in \mathbb{R}^L.$ Subsequently, a round based decoding is done to get the label vector. To summarize, first PLST performs SVD on $Y$ and obtain $U_{M}^T$, and then runs LLST with $P$ as $U_{M}^T$. The decoder $D$ now becomes $D(r(x)) = \text{round}(U_M\cdot(x)).$ 
\paragraph{}
PLST only considers labl-label correlations during the label space dimensionality reduction (LDSR). In the paper \cite{cplst_chen}, a LSDR approach is explored which takes into account both the label and the feature correlations. The claim is that, such a method will be able to find a better embedding space due to the incorporation of extra information. The paper also provides an upper bound of Hamming loss, thus providing theoretical guarantees for both PLST and the proposed method, conditional principal label space transformation (CPLST). The method uses the concept of both PLST and Canonical Correlaion Analysis (CCA) to find the optimal embedding matrix.

\paragraph{}
Let $Z \in \mathbb{R}^{n \times L}$ such that $z = y - \Bar{y},$  where  $\Bar{y} = \frac{1}{n}\sum_{i=1}^{n}y_i$ is the estimated mean of the label set vectors. Let $V \in \mathbb{R}^{k \times L}$ be a projection matrix, $t = zP^T, t \in \mathbb{R}^k$ is the embedded vector of  $z$. Let $r$ be the regressor to be learned such that $r(x) = t$, for a sample $x \in \mathbb{R}^d.$ Let the predictions be obtained as $h(x) = \text{round}(V^Tr(x) + \Bar{y})$  with $V$ having orthogonal rows. In this setting, for round based decoding and orthogonal linear transformation $V$, the Hamming loss is bounded by 
\begin{equation} \label{eqn:HL_bound}
    HL_{\text{train}} \leq c\bigg(\underbrace{ \| r(X) - ZP^T \|_{F}^{2}}_\text{Prediction error} + \underbrace{ \| Z - ZP^TP \|_{F}^{2}}_\text{Encoding error} \bigg). 
\end{equation}
\paragraph{}
Canonical Correlation Analysis (CCA) is a method generally used for feature space dimensionality reduction (FSDR). In the context of this method, it can be interpreted as feature aware Label Space Dimensionality Reduction (LSDR). The CCA is utilized for analysis of linear relationship between two multi-dimensional variables. It finds two sets of basis vectors $(w_{x}^{(1)},w_{x}^{(2)}, \cdots)$ and $(w_{z}^{(1)},w_{z}^{(2)}, \cdots)$
such that the correlation coefficient between the canonical variables $c_{x}^{(i)} = Xw_{x}^{(i)}$ and $c_{z}^{(i)} = Zw_{z}^{(i)}$ is maximized, where $w_{x}^{(i)} \in \mathbb{R}^{d}, w_{z}^{(i)} \in \mathbb{R}^{L}$.

\paragraph{}
A different version of the algorithm, called Orthogonally Constrained CCA (OCCA) preserves the original objective of CCA and specifies that $W_Z$ must contain orthogonal rows to which round based decoding can be applied. Then, using the hamming loss bound \eqref{eqn:HL_bound}, when $P = W_Z$ and $r(X) = XW_{X}^{T},$ OCCA minimizes $||r(X) - ZW_{Z}^{T}||$ in \eqref{eqn:HL_bound}.  That is, OCCA is applied for the orthogonal directions $P$ that have low prediction error in terms of linear regression.



After some simplification, the objective for OCCA becomes 
\begin{equation} \label{eqn:OCCA_trace}
    \begin{split}
        \underset{PP^T = I}{\min}||HZP^T - ZP^T||_{F}^{2} 
        \approx \underset{PP^T = I}{\min} \text{tr}(PZ^T(I-H)ZP^T),
    \end{split}
\end{equation}
where $H = XX^{\dagger}$. 


\paragraph{}
Thus, for minimizing the objective in \eqref{eqn:OCCA_trace} we only need the eigenvectors corresponding to the smallest eigenvalues of $Z^T(I - H)Z$, or the eigenvectors corresponding to the largest eigenvalues of $Z^T(H-I)Z$.

\paragraph{}
It can be seen that OCCA minimizes the prediction error in the Hamming Loss bound \eqref{eqn:HL_bound} with the  the orthogonal directions $V$ that are relatively simpler to learn in terms of linear regression. In contrast, PLST minimizes the encoding error of the bound \eqref{eqn:HL_bound} with the ``principal" components. The two algorithms OCCA and PLST can be combined to minimize the two error terms simultaneously with the ``conditional principal" directions. Then, the optimization problem becomes
\begin{equation}
    \begin{split}
        \underset{W, \, PP^T=I}{\max} \text{tr} \bigg( PZ^THZP^T \bigg).
    \end{split}
\end{equation}
\paragraph{}
This problem can be solved by taking the eigenvectors with the largest eigenvalue of $Z^THZ$ as the rows of $P$. This $P$ minimizes the prediction error as well as the encoding error simultaneously. This obtained P is useful for the label space dimensionality reduction. The final CPLST algorithm is described below.

\begin{algorithm}[t!]
\caption{ Conditional Principal Label Space Transformation}
\begin{algorithmic}[1]
    \State Let $Z = [z_1, \cdots, z_N]^T$ with $z_n = y_n - \Bar{y}.$
    \State Perform SVD of $Z^THZ$ to obtain $Z^THZ = A \Sigma B$ with $\Sigma = \text{diag}(\sigma_1, \ldots, \sigma_N)$ such that $\sigma_1 \geq \cdots \geq \sigma_N.$ Let $P_K$ contain the top $K$ rows of $B$.
    \State Encode $\{(x_n,y_n)\}_{n=1}^{N}$ to $\{(x_n,t_n)\}_{n=1}^{N},$ where $t_n = P_Kz_n.$
    \State Learn a multi-dimensional regressor $r(x)$ from $\{(x_n,t_n)\}_{n=1}^{N}.$
    \State Predict the label-set of an instance $x$ by $h(x) = \text{round}(P_{K}^{T}r(x) + \Bar{y}).$
\end{algorithmic}
\end{algorithm}

While CPLST provides a good bound on the error and shows theory behind the matrix compression based methods, it does not perform very well and also takes considerable time for prediction. 

\paragraph{}
The paper \cite{bloom_cisse}, takes a novel approach to multi-label classification problem. The concept of Bloom filters \citep{bloom_bloom} is used to reduce the problem to a small number of binary classifications by representing label sets as low-dimensional binary vectors. However, since a simple bloom filter based approach does not work very well, the approach designed in this paper uses the observation that many labels almost never appear together in combination with Bloom Filters to provide a robust method of multi-label classification.

\paragraph{}
Bloom Filters are  space efficient data structures which were originally designed for approximate membership testing.  Bloom Filter (BF) of size $B$ uses $K$ hash functions, where each one maps from number of labels $L$ to $\{1,\cdots,B\}$, which are denoted as $h_k : L \rightarrow \{1, \cdots, B\}$ for $k \in \{1, \cdots, K\}$. The value of  $h_k(\ell)$ is random and chosen uniformly from $\{1, \cdots,B\}$. Each of the hash functions define a representative bit of the entire bit vector of size $B$. The bit vector of a label $\{\ell\}$ is obtained by concatenating all $h_k(\ell)$. There are at most $K$ non-zero bits in this bit vector. A subset $y \subseteq L$ is represented by a bit vector of size $B$, defined by the bitwise-OR of the bit vectors of each of the labels in $y$. The Bloom Filter predicts if the label is present in the set or not, by testing if the encoding of the set contains $1$ at all the representative bits of the label's encoding. If the label is present, the answer is always correct, but if it is not present, there is a chance of a false positive.

\paragraph{}
Bloom filters can be naively applied to extreme classification directly. Let $L$ be the number of labels. Each individual label is encoded into a $K$-sparse bit vector of dimension $B$ such that $B \ll\ell$, and a disjunctive encoding of label sets (bit-wise-OR of the label codes that appear in the label set) represents the encoding of the entire label set. For each of the $B$ bits of the coding vector, one binary classifier is learned. Thus, the number of classification tasks is reduced from $L$ to $B$. If $K > 1$, the individual labels can be encoded unambiguously on far less than $L$ bits. Also, the classifiers can be trained in parallel independently. However, standard BF is not robust to errors in the predicted representation. Each bit in the BF is represents multiple labels of the label set. Thus, an incorrectly prediction of a bit, may cause inclusion (or exclusion) of all the labels which it represents from the label set. 

\paragraph{}
This problem can be tackled by taking into consideration the fact that the distribution of the label sets for real world datasets is not uniform. If it is possible to detect a false positive from the existing distribution of labels, then we can correct it. For example, if $y$ is the set of labels and $\ell \notin y$ is a false positive given $\mathbb{e}(y),$ then $\ell$ can be detected as a false positive if it is known that $\ell$ never appears together with the labels in $y$.  Thus, this method uses the non-uniform distribution of label sets to design the hash functions and a decoding algorithm to make sure that any incorrectly predicted bit has a limited impact on the predicted label set.

\paragraph{}
This paper develops a  new method called the Robust Bloom Filter (RBF) which utilizes the fact that in most real-world datasets, many pairs of labels do not occur together, or co-occur with a very small probability. This allows it to improve over simple random hash functions. This is done in two ways, clustering and error correction.

\paragraph{}
During label clustering, the label set $L$ is partitioned into $P$ subsets $L_1,\cdots,L_P$ such that clusters $(L_1,\cdots,L_P)$ are mutually exclusive , i.e.,  no target set of a sample contains labels from more than one of each of the clusters $L_p, p = 1, \cdots,P$.
If the disjunctive encoding of Bloom filters is used, and if the hash function are designed such that the false positives for every label set can be detected to be mutually exclusive of the other labels. The decoding step can then detect this and perform correction on the bit. If for every bit, the labels in whose encodings it appears is mutually exclusive, then we can ensure a smooth decoding procedure. More clusters will result in smaller encoding vector size $B$, which will  result in fewer number of binary classification problems. Further, the clustering is performed after removing head labels, as they co-occur too frequently to divide into mutually exclusive subsets. For finding the label clustering, the co-occurrence graph is built, and the head labels are removed using the degree centrality measure (\textit{Degree Centrality} is defined as the number of edges incident on a vertex). The remaining labels are then clustered using the Louvain Algorithm \citep{louvain_blondel}. The maximum size of each cluster is fixed to control the number of clusters.


\paragraph{}
As the second step, assuming that the label set $L$ is partitioned into mutually exclusive clusters (after removing the head labels), given a parameter $K$, we can construct the required $K$-sparse encodings by following two conditions: (1) two labels from the same cluster cannot share any representative bit (2) two labels from different clusters can share at most $K-1$ representative bits, ie. they cannot have the exact same encoding.
Let $K$ denote the number of hash functions, $R$ denote the size of the largest cluster, $Q$ denote the number of bits assigned to each label such that $B = R.Q$, $Q \geq K$ and $P \leq \binom{Q}{K}.$ For a given $r\in \{1,\cdots,R\},$ which is the $r$-th batch of $Q$ successive bits, $Q$ bits are used for encoding the $r$-th label of each cluster. In this way, the first condition is satisfied. Also, this encoding can be used for $P$ labels. Now, for a given batch of $Q$ bits, there are $\binom{Q}{K}$ different subsets of $K$ bits. Thus, we can have at most $\binom{Q}{K}$ label clusters. The $P$ labels (one from each cluster) are injectively (one-one) mapped to the subsets of size $K$ to define the $K$ representative bits of these labels. Thus, these encoding scheme satisfies condition 2 as no two subsets in $\binom{Q}{K}$ shares more than $K-1$ indexes. Using a BF of size $B = R.Q,$ we have $K$-sparse label encodings that satisfy the two conditions for $L \leq R.\binom{Q}{K}$ labels partitioned into $P \leq \binom{Q}{K}$ mutually exclusive clusters of size at most $R.$ In terms of compression ratio $\frac{B}{L},$ this encoding scheme is most efficient when the clusters are perfectly balanced, and the number of partitions $P$ is exactly equal to $\binom{Q}{K}$ for some $Q.$ An implementation of this scheme is shown in Table \ref{tab:bloom} (for $Q = 6$ bits, $P = 15$ clusters, $K = 2$ representative bits and $R = 2$).
\small
\begin{table}[t!]
\centering 
    \begin{subtable}{.55\linewidth}
      \centering
        \caption{}
        \begin{tabular}{ |p{0.6cm}| p{2.3cm}|| p{0.6cm} | p{2.55cm}|}
        \multicolumn{4}{c}{} \\
        \hline
        bit index  &  representative for labels &   bit index &  representative for labels \\
        \hline
            $1$ &  $\{1,2,3,4,5\}$ &  $7$ &  $\{16,17,18,19,20\}$ \\
            $2$ &  $\{1,6,7,8,9\}$ &  $8$ &  $\{16,21,22,23,24\}$ \\
          $3$ &  $\{2,6,10,11,12\}$ &  $9$ &  $\{17,21,25,26,27\}$ \\
          $4$ &  $\{3,7,10,13,15\}$ &  $10$ &  $\{18,22,25,28,29\}$ \\
          $5$ &  $\{4,8,11,13,15\}$ &  $11$ &  $\{19,23,26,28,30\}$ \\
          $6$ &  $\{5,9,12,14,15\}$ &  $12$ &  $\{20,24,27,29,30\}$ \\
         \hline
        \end{tabular}
            \end{subtable}%
     \begin{subtable}{.45\linewidth}
      \centering
        \caption{}
        \begin{tabular}{ |p{0.8cm}| p{1.2cm}|| p{0.8cm} | p{1.2cm}|}
 
     \multicolumn{4}{c}{} \\
     \hline
      cluster index  &  labels in cluster &   cluster index &  labels in cluster \\
     \hline
      $1$ &  $\{1,15\}$ &  $9$ &  $\{9,23\}$ \\
      $2$ &  $\{2,16\}$ &  $10$ &  $\{10,24\}$ \\
      $3$ &  $\{3,17\}$ &  $11$ &  $\{11,25\}$ \\
      $4$ &  $\{4,18\}$ &  $12$ &  $\{12,26\}$ \\
      $5$ &  $\{5,19\}$ &  $13$ &  $\{13,27\}$ \\
      $6$ &  $\{6,20\}$ &  $14$ &  $\{14,28\}$ \\
      $7$ &  $\{7,21\}$ &  $15$ &  $\{15,29\}$ \\
      $8$ & $\{8,22\}$ & &\\
     \hline
    \end{tabular}
    \end{subtable}%
    \caption{\label{tab:bloom}Representative bits for 30 labels partitioned into $P= 15$ mutually exclusive label clustersof size $R= 2$, using $K= 2$ representative bits per label and batches of $Q= 6$ bits. The table on the right gives the label clustering. The injective mapping between labels and subsets of bits is defined by $g:\ell \rightarrow \{g_1(\ell) = (1 +\ell)/6, g_2(\ell) = 1 +\ell \, \text{mod} 6\}$ for $\ell \in \{1,\ldots,15\}$ and,  for $\ell \in \{15,\ldots,30\},$ it is defined by $\ell \rightarrow \{(6 +g_1(\ell - 15), \, 6 +g_1(\ell - 15)\}$}
\end{table}
\normalsize


    

\paragraph{}
During the prediction procedure, given an sample $x$ and the encoding predicted $\mathrm{\hat{e}(x)}$, the predicted label set $d(\mathrm{\hat{e}(x)})$ is computed by a two-step process. Firtsly, cluster identification is performed by picking whichever cluster has maximum number of representative  bits, and close to the number of representative bits close to that of $\mathrm{\hat{e}(x)}$. Secondly, the correctl labels within the cluster are found by using probabilistic sampling under the assumption that whichever label in the cluster has the most representative bits in common with $\mathrm{\hat{e}(x)}$ is the most likely to occur.
If logistic regression is used as base learners for binary classification, the posterior probabililty can be used for computing the cluster scores instead of the binary decisions. The cluster which maximizes the cluster score is chosen. The advantage of using a randomized prediction for the labels is that a single incorrectly predicted bit does not result in too many predicted labels.


The paper also provides some theoretical bounds on the error by deriving that each incorrectly predicted bit in the BF cannot imply more than two incorrectly predicted labels. However, R-BF still does not show very good results, probably due to the assumption that labels from different clusters will never co-occur, which highly limits the output space.

\paragraph{}
The paper \cite{jasinska2016log} also takes a similar approach to encoding the labels. A method called Log Time Log Space Extreme classification (LTLS) is proposed which transforms the multi-label problem with $L$ labels into a structured prediction problem by encoding the labels as edges of a trellis graph. This enables
inference and training in order which is logarithmic in the number of labels along with reduced model size.


        



\paragraph{}
First, a trellis graph with $O(\log_2 (L))$ steps is built, each with $2$ states, a root, and a sink vertex as shown in the Fig. \ref{fig:ltls_example}. There are $E$ edges in the graph. A path from the root to the sink $s$ is a vector of length $E$. There are $L$ paths in the graph. Each label $l$, $l \in \{1, \dots, L \}$, is assigned to path $s_{l}$. Additional edges are used if the number of labels is not power of $2$.
This step is equivalent to binary encoding of number of labels. The number of edges $E \leq \log_2 (L) + 1$. Each edge in the above trellis corresponds to a function $f$ we learn. For a label $l$, the encoding vector has $1$ at places where the edge is part of the corresponding path $s_l$, or has $0$ otherwise.


\begin{figure}[t!]
\centering
\begin{tikzpicture}[>=Stealth,node distance=1.8cm,
    every node/.style={circle,draw,fill=blue!10,inner sep=3pt,font=\small},
    edge/.style={->,very thin}]
\node (v0)                {0};

\node (v1) [right=of v0, yshift= 1.2cm] {1};
\node (v2) [right=of v0, yshift=-1.2cm] {2};

\node (v3) [right=of v1] {3};
\node (v4) [right=of v2] {4};

\node (v6) [right=of v3] {6};
\node (v5) [right=of v4] {5};    

\node (v7) [right=of v6, yshift=-1.2cm] {7};

\foreach \a/\b in {v0/v1, v0/v2,
                   v1/v3, v1/v4,
                   v2/v3, v2/v4,
                   v3/v6, v3/v5,
                   v4/v6, v4/v5,
                   v6/v7, v5/v7}
  \draw[edge] (\a) -- (\b);
\end{tikzpicture}
\caption{Trellis encoding used by LTLS \citep{jasinska2016log}.  
Vertex 0 is the source, vertices 1–6 form three successive layers of decision points, and vertex 7 is the sink.  A classifier is attached to every directed edge; at test time the model chooses, at each layer, either the {\em upper} edge (interpret as bit ``1'') or the {\em lower} edge (bit ``0''), so that the complete source→sink route is a 3-bit code that uniquely identifies one label among the \(2^{3}=8\) possibilities: Example 1 – label 7 (111) Path \(0 \!\to\! 1 \!\to\! 3 \!\to\! 6 \!\to\! 7\): the upper edge is taken at all three layers, yielding the binary string 111, i.e. label 7 in a zero-based enumeration. Example 2 – label 4 (100) Path \(0 \!\to\! 1 \!\to\! 4 \!\to\! 5 \!\to\! 7\): upper edge at layer 1, lower edge at layers 2 and 3, producing 100, which encodes label 4. Because the trellis width grows only logarithmically with the number of labels, LTLS needs \(O(\log L)\) classifiers and prediction time, yet can address every label through its unique path.}
\label{fig:ltls_example}
\end{figure}

During prediction, given an input $x$, the algorithm predicts $E$ values from each of the binary classifiers. 
The value at each edge $e_{i}$ is a function $f_{i}: \mathcal{X} \rightarrow \R$ which can be assumed to be a simple linear regressor. The score $F(x, s_{l})$ of label $l$ is the sum of values in the path $s_{l}$. 
\begin{align}
    F(x, s_{l}) = \sum_{i \in s_{l}} f_i(x). 
\end{align}

\paragraph{}
For prediction of the final labels from the label scores, we can use the scheme where top-k labels correspond to $k$ longest paths in the trellis graph (which can be found using DP based list Viterbi algorithm \citep{seshadri1994list}). 

Despite the unique approach, the algorithm failed to give very high scores compared to emerging tree based and linear algebra based methods. Subsequent work was done by \cite{evron2018efficient} which attempted different graph structures, but it was only applicable for multi-class classification. 

\paragraph{}
Although \cite{bloom_cisse} uses hashing based label space compression technique, it makes assumptions about the co-occurrence of label which leads to reduction in performance. The paper \cite{mach_medini}, proposes  Merged Averaged Classifiers via Hashing (MACH) which is a hashing based approach to XMLC which utilizes something similar to a count sketch data structure instead of bloom filters for performing universal hashing. This reduces the task of extreme multi-label classification to a small number of parallelizable binary classification tasks.

Sketch \citep{sketch_moses} is a data structure that stores a summary of a dataset in situations where the whole data would be prohibitively costly to store. The count-sketch is a specific type of sketch which is used for counting the number of times an element has occurred in the data stream.
Let $S = q_1, q_2,\cdots,q_n$ be a stream of queries, where each $q_i \in \mathcal{O}$ which is the set of objects queried. Let $h_1,\cdots, h_t$ be $t$ hash functions from $\mathcal{O}$ to $\{1, \cdots,b\}$.
Count sketch data structure consists of these hash functions along with a $t \times b$ array of counters $C$, where $t$ is the number of hash functions and $b$ is the number of buckets for each hash table. There are two supported operations: (1) Insert new element $q$ into sketch $C$, which is performed by adding $1$ for each hash function $t$ at position $C[t][h_t(q)]$ (2) Count the number of times element $q$ has occurred, performed by $F_{t=1...T} (C[t][h_t(q)])$ where $F$ can be mean, median or minimum.


If we use only one hash function, then we get the desired count value in expectation but the variance of the estimates is high. To reduce the variance, $t$ hash functions are used and their outputs are combined. This concept is utilized by the proposed method.

\paragraph{}
The method Merged-Averaged Classifiers via Hashing (MACH) randomly merges $L$ classes into $B$ random-meta-classes or buckets $(B \ll L)$ using randomized hash functions. This merging of labels into meta classes gives rise to a smaller multi-label classification problem and classifiers such as logistic regression or a deep network is used for this meta-class classification problem. Let $R$ be the number of hash functions. Then, the classification algorithm is repeated independently for $R = O(\text{log }L)$ times using an independent 2-universal hashing scheme each time. During prediction, the outputs of each of the $R$ classifiers is aggregated to obtain the prediction class (The aggregation scheme may be median, mean or min). This is the step where an estimate like method of the count sketch data structure is used. The advantage of MACH stems from the fact that since each hash function $h_j$ is independent, training the $R$ classifiers are trivially parallelizable.  The parameters $B$ and $R$ can be tuned to trade accuracy with both computation and memory.
The algorithms are described below.
\newline

\small 
\begin{minipage}{0.46\textwidth}
\begin{algorithm}[H]
\centering
\caption{ Train}
\begin{algorithmic}[1]
    \State \textbf{Data : } $X \in \mathbb{R}^{N \times d}, Y \in \mathbb{R}^{N \times L}$
    \State No. of buckets, $B$, no of hash functions, $R$
    \State \textbf{Output : } $R$ trained classifiers
    \State initialize $R$ 2-universal hash functions $h_1,h_2, \cdots,h_R$
    \State initialize \textit{result} as an empty list
    \State for $i = 1 : R$ do
    \State \quad $Y_{h_i} \leftarrow h_i(Y) $
    \State \quad $M_i = {\tt trainClassifier}(X,Y_{h_i})$
    \State \quad Append $M_i$ to \textit{result}
    \State end for
    \State \textbf{Return } \textit{result}
\end{algorithmic}
\end{algorithm}
\end{minipage}
\hfill
\begin{minipage}{0.46\textwidth}
\begin{algorithm}[H]
\centering
\caption{ Predict}
\begin{algorithmic}[1]
    \State \textbf{Data : } $X \in \mathbb{R}^{N \times d}, Y \in \mathbb{R}^{N \times L}$
    \State \textbf{Input : }$M = M_1,M_2,\cdots,M_R$
    \State \textbf{Output : }$N$ predicted labels
    \State load $R$ 2-universal hash functions $h_1,h_2,\cdots,h_R$ used in training
    \State initialize $P$ as an empty list
    \State initialize $G$ as a $(|N|*L)$ matrix
    \State for $i = 1 : R$ do
    \State \quad $P_i = \text{getProbability}(X,M_i)$
    \State \quad Append $P_i$ to $P$
    \State end for
    \State for $j = 1 :\ell$ do
    \State \quad /*$G[:,j]$ indicates the $j^{th}$ column in matrix $G$*/
    \State \quad $G[:,j] = (\sum_{r=1}^{R}P_r[:,h_r(j)])/R$
    \State end for
    \State \textbf{Return }argmax(G,axis=1)  //whichever label has the highest probability for a given sample
\end{algorithmic}
\end{algorithm}
\end{minipage}
\normalsize

\paragraph{}
The paper is able to prove a bound on the error due to the theorem that if a pair of classes is not indistinguishable, then there is at least one classifier which provides discriminating information between them.
If the loss function used is softmax, then we get the multiclass version of MACH. For a multi-label setting, the loss function can be binary cross-entropy.






\paragraph{Comparison between above stated methods}
\begin{enumerate}
    \item Compressed Sensing \citep{cs_hsu}, even though it provides theoretical bounds, does not perform very well in the results and prediction times. It is consistently outperformed by PLST \citep{plst_tai}.
    \item CPLST \citep{cplst_chen} outperforms PLST but is much slower and less scalable.
    \item R-BF \citep{bloom_cisse} and LTLS \citep{jasinska2016log} outperform the results of CPLST but are outperformed by \cite{mach_medini}.
\end{enumerate}

\begin{table}[t!]
\centering 
\footnotesize
\begin{tabular}{ |p{3cm}||p{3cm}|p{3cm}|p{3cm}|  }
 \hline
 \multicolumn{4}{|c|}{Compressed Sensing Based Methods} \\
 \hline
  Algorithm & Compression & Learning & Reconstruction\\
 \hline
 CS   & Compression Matrix obtained from compression methods such as OMP & Simple regressors in the reduced space & Solving an optimization problem \\ \hline 
 PLST & Compression matrix obtained from SVD & Simple regressors in the reduced space & Round based decoding after multiplying with transpose of compression matrix\\ \hline 
 CPLST & Compression matrix obtained from a combination of OCCA and PLST, considering both feature-label and label-label correlations & Simple regressors in the reduced space & Round based decoding after multiplying with transpose of compression matrix\\ \hline 
 R-BF & Ecodes every subset of labels into a bit encoding by `OR' operation on individual labels & Binary Classifiers for each bit in encoding & An algorithm based on identification of the closest labels included within the predicted encoding \\ \hline 
 LTLS & Each label is assigned to a path in a trellis graph and a corresponding bit vector is obtained & Regressors at each edge predict the weight of the edges & Longest path algorithms are used to find the top k paths and corresponding labels are predicted\\ \hline 
 MACH & Universal hash functions are used to assign each label a bucket. Multiple hash functions are used to iterate the process. & Multi label classifiers are used to predict the corresponding meta labels for each hash function & A sketch-like aggregation function is used to find the predicted labels from the predicted meta-labels\\
 \hline
\end{tabular}
\end{table}
\normalsize

\subsection{Linear-Algebra Based Methods}
\paragraph{}
The methods in this category do not follow any specific structure, but are heavily reliant on simple Linear Algebra based optimizations, which allow the methods to gain an advantage over simply embedding in a lower dimensional space like in compressed sensing based methods. The methods may end up working like compressed sensing ones, but generally aim to do some improvement over them. For example, LEML converts the entire process of compression, prediction and decoding in CPLST to one step directly and proves the equivalence. SLEEC on the other hand refutes the usage of the low-rank assumption used in most of the embedding methods and develops a KNN based distance preserving embeddings. PD-SPARSE develops a method for optimizing the XMLC problem in both primal and dual spaces and obtain sparse embeddings.  A summary of each method is provided at the end of the section.

\paragraph{}
Subset-selection is one of the most common methods utilized to make the extreme classification problem tractable. The core idea was to find a good representative subset of labels on which a simple classifier can be learnt and the predictions can be scaled back to the full label set. The paper by \cite{cssp_bi} performs a subset selection on the labels to form a smaller label space which approximately spans the original label space. The labels are selected via a randomized sampling procedure where the probability of a label being selected is proportional to the leverage score of the label on the best possible subset space.

\paragraph{}
\cite{moplms_balasubramanian} has previously attempted the label selection procedure. This paper assumes that the output labels can be predicted by the selected set of labels. It solves an optimization problem for selecting the subset. However, the size of the label subset cannot be controlled explicitly and also the optimization problem quickly becomes intractable. To address this issue, this paper models the label subset selection problem as a Column Subset Selection Problem (CSSP).

\paragraph{}
The method Column Subset Selection Problem (CSSP) seeks to find exactly $k$ columns of $Y$ such that these columns span $Y$ as well as possible, given a matrix $Y \in \mathbb{R}^{n \times L}$ and an integer $k$. For this purpose, an index set $C$ of cardinality $k$ is required such that $||Y - Y_C Y_{C}^{\dagger}Y||_F$ is minimized. Here, $Y_C$ denotes the sub-matrix of $Y$ containing the columns indexed by $C$ and $Y_C Y_{C}^{\dagger}$ is the projection matrix on the $k$-dimensional space spanned by the columns of $Y_C.$ In \cite{cssp_boutsidis}, this problem is solved in two stages : a \textit{randomized stage} and a \textit{deterministic stage}. Let $V_{k}^{T}$ denote the transpose of the matrix formed by the top $k$ right singular vectors of $Y.$ In the randomized stage, a randomized column selection algorithm is used to select $\theta(k \log k)$ columns from $Y.$ Each column $i$ of $V_{k}^{T}$ is assigned  a probability $p_i$ which corresponds to the leverage score of the $i^{th}$ column of $Y$ on the best rank $k$ subspace of $Y.$ It is defined as $p_i = \frac{1}{k}||(V_{k}^{T})_{i}||_{2}^{2}.$ The randomized stage of the algorithm samples from the columns of $Y$ using this probability distribution. In the deterministic stage, a rank revealing QR (RRQR) decomposition is performed to select exactly $k$ columns from a scaled version of the columns sampled from $Y.$

\paragraph{}
In the extreme classification setting, the objective of the label subset selection problem can be stated as 
\begin{equation} \label{eqn:cssp_obj}
    \underset{C}{\min} \, ||Y - Y_C Y_{C}^{\dagger}Y||_F.
\end{equation}
The approach proposed in this paper directly selects the $k$ columns of the label matrix $Y.$ The algorithm is given in Algorithm \ref{algo:cssp}. First, a partial SVD of $Y$ is computed to pick the top $k$ right singular vectors, $V_k \in \mathbb{R}^{n \times k}.$ Then, the columns in $Y$ are sampled with replacement, with the probability of selecting the $i^{th}$ column being $p_i = \frac{1}{k}||(V_k)^{T}_{i}||_{2}^{2}.$ However, instead of selecting $\Theta(k \log k)$ columns, the sampling procedure is continued until $k$ different columns are selected. It has been shown that if the number of trials is $T = \frac{2c_0^2 k}{\epsilon^2} \log \frac{c_0^2 k}{\epsilon^2}$ (which is in $O(k \log k)$) for some constants $c_0$ and $\epsilon,$ then the algorithm gives a full rank matrix $(V_k)^T_C$ with high probability. Once the $k$ columns have been obtained, $k$ classifiers are learned.

\small 
\begin{algorithm}[t!] 
\caption{Multi-label classification via CSSP (ML-CSSP)}
\label{algo:cssp}
\begin{algorithmic}[]
    \State Compute $V_k,$ the top $k$ right singular vectors of $Y.$
    \State Compute the sampling probability $p_i$ for each column in $Y$ as defined above.
    \State $C \leftarrow \Phi$
    \State \textbf{while } $|C| < k$ \textbf{ do}
    \State \quad Select an integer from $\{1,2,\ldots, L\}$ where the probability of selecting $i$ is equal to $p_i.$
    \State \quad \textbf{if } $i \notin C$ \textbf{ then}
    \State \quad \quad $C \leftarrow C \cup \{i\}$
    \State \quad \textbf{end if}
    \State \textbf{end while}
    \State Train the classifier $f(x)$ from $\{x^{i}, y_{C}^{i}\}_{i=1}^{n}.$
    \State Given a new test point $x,$ obtain its prediction $h$ using $f(x)$ and return $\hat{y}$ by rounding $h^{T}Y_{C}^{\dagger}Y$
\end{algorithmic}
\end{algorithm}
\normalsize 

\paragraph{}
The minimizer of the objective \eqref{eqn:cssp_obj} satisfies 
\begin{equation} \label{eqn:cssp_min}
Y \approx Y_C Y_C^{\dagger}Y.
\end{equation}
From \eqref{eqn:cssp_min}, it can be seen that each row of $Y$ can be approximated as the product of each row of $Y_C$ with $Y_C^{\dagger}Y.$ Now, given a test sample $x \in \mathbb{R}^{d},$ let $h \in k$ denote the prediction vector. Then, using \eqref{eqn:cssp_min}, the $L$-dimensional prediction vector can be obtained as $h^TY_{C}^{\dagger}Y.$ The elements of the $L$-dimensional output vector are further rounded to get a binary output vector.

\paragraph{}
Empirical results show that ML-CSSP gives lower RMSE compared to other methods available at that time. The given CSSP algorithm is also compared to the two-stage algorithm of \citep{cssp_boutsidis} and the results show that the number of sampling trials and encoding error is better for the proposed algorithm. As the number of selected labels $k$ is increased, training error increases as the number of learning problems increase while the encoding errors decrease as $k$ is increased, as expected. ML-CSSP is able to achieve lower training errors than PLST \citep{plst_tai} and CPLST \citep{cplst_chen} as selected labels are easier to learn than transformed labels.

\paragraph{}
In the paper \cite{leml_yu}, the problem of extreme classification with missing labels (labels which are present but not annotated) is addressed by formulating the original problem as a generic empirical risk minimization (ERM) framework. It is shown that the CPLST \citep{cplst_chen}, like compression based methods, can be derived as a special case of the ERM framework described in this paper.

\paragraph{}
The predictions for the label vector are parameterized as $f(\mathrm{x};Z) = Z^T\mathrm{x},$ where $Z \in \mathbb{R}^{d \times L}$ is a linear model. Let $\ell(\mathrm{y},f(\mathrm{x};Z)) \in \mathbb{R}$ be the loss function.
The loss function is assumed to be decomposable, i.e., $$\ell(\mathrm{y},f(\mathrm{x};Z)) = \sum_{j=1}^{L}\ell(y^j,f^j(\mathrm{x};Z)).$$

\paragraph{}
The motivation for the framework used in this paper comes from the fact that even though the number of labels is large, there exists significant label correlations. This reduces the effective number of parameters required to model them to much less than $d \times\ell,$ which in turn constrains $Z$ to be a low rank matrix.

\paragraph{}
For a given loss function $\ell(\cdot)$ and no missing labels, the parameter $Z$ can be learnt by the canonical ERM method as
\begin{equation} \label{eqn:erm_without_missing_labels}
    \begin{split}
        \hat{Z} &= \underset{Z \, : \, \text{rank}(Z) \leq k}{\text{argmin}}J(Z) = \sum_{i=1}^{n}\sum_{j=1}^{L}\ell(Y_{ij},f^j(x_i;Z)) + \lambda r(Z), \\
    \end{split}
\end{equation}

where $r(Z): \mathbb{R}^{d \times L} \rightarrow \mathbb{R}$ is a regularizer. If there are missing labels, the loss is computed over the known labels as
\begin{equation} \label{eqn:erm_with_missing_labels}
    \begin{split}
        \hat{Z} &= \underset{Z \, : \, \text{rank}(Z) \leq k}{\text{argmin}}J(Z) = \underset{(i,j)\in \Omega}{\sum}\ell(Y_{ij},f^j(x_i;Z)) + \lambda r(Z), \\
    \end{split}
\end{equation}
where $\Omega \subseteq [n] \times [L]$ is the index set that represents known labels. Due to the non-convex rank constraint, this \eqref{eqn:erm_with_missing_labels} becomes an NP-hard problem. However, for convex loss functions, the standard alternating minimization method can be used. For L2-loss, eq. \eqref{eqn:erm_without_missing_labels} has a closed form solution using SVD. This gives the closed form solution for CPLST.

\paragraph{}

Most of the compressed sensing based methods and previous embedding based methods assume that the label matrix is low rank and aims to hence predict in a smaller dimensional embedding space rather than the original embedding space. The paper \cite{sleec_bhatia} claims that the low-rank assumption is violated in most practical XMLC problems due to the presence of the tail labels which act as outliers and are not spanned by the smaller embedding space. Thus, this paper proposes a method called Sparse Local Embeddings for Extreme Multi-label Classification (SLEEC) which utilizes the concept of "distance preserving embeddings" to predict both head and tail labels effectively. 



\paragraph{}
The global low rank approximations performed by embedding based methods do not take into account the presence of tail labels. 
It can be seen that a large number of labels occur in a small number of documents. 
Hence, these labels will not be captured in a global low dimensional projection.

    

    

\paragraph{}
In SLEEC embeddings $z_{i}$ are learnt which non-linearly capture label correlations by preserving the pairwise distances between only the closest label vectors,  i.e. $d(z_i,z_j) = d(y_i,y_j)$ only if $ i \in \text{kNN}(j),$ where $d$ is a distance metric. Thus, if one of the label vectors has a tail label, it will not be removed by approximation (which is the case for low rank approximation). Regressors $V$ are trained to predict $z_{i} = Vx_{i}.$ During prediction SLEEC uses kNN classifier in the embedding space, as the nearest neighbours have been preserved during training. Thus, for a new point $x$, the predicted label vector is obtained using $y = \sum_{i : Vx_i \in \text{kNN}(Vx)}y_i.$ For speedup, SLEEC clusters the training data into $C$ clusters, learns a separate embedding per cluster and performs kNN only within the test point's cluster. Since, clustering can be unstable in large dimensions, SLEEC learns a small ensemble where each individual learner is generated by a different random clustering.

\paragraph{}
Due to the presence of tail labels, the label matrix $Y$ cannot be well approximated using a low-dimensional linear subspace. Thus it is modeled using a low-dimensional non-linear manifold. That  is,  instead  of  preserving  distances of a given label vector to all the training points, SLEEC attempts to preserve the distance to only a few nearest neighbors. That is, it finds a $\hat{L}-$dimensional matrix $Z = [z_1,z_2, \ldots,z_n] \in \mathbb{R}^{\hat{L} \times n}$ which minimizes the following objective:

\begin{equation}
    \underset{Z \in \mathbb{R}^{\hat{L}\times n}}{\min} ||P_{\Omega}(Y^TY) - P_{\Omega}(Z^TZ)||_{F}^{2},
    \label{eqn:SVP}
\end{equation}
where  the index set $\Omega$ denotes the set of neighbors that we wish to preserve, i.e., $(i,j) \in \Omega$ iff $j \in N_i.$ $N_i$ denotes the set of nearest neighbors of $i$.
$P_\Omega : \mathbb{R}^{n \times n} \rightarrow \mathbb{R}^{n \times n}$ is defined as :

\begin{equation}
    (P_{\Omega}(Y^TY))_{ij} = 
    \begin{cases}
    \langle y_i,y_j \rangle, & \text{if } (i,j) \in \Omega\\
    0, & \text{otherwise}.
    \end{cases} \label{eqn:pomega}
\end{equation}




The above equation, after simplification can be solved by using Singular Value Projection (SVP) \citep{svp_jain} using the ADMM \citep{admm_sprechmann} optimization method. The training algorithm is shown in Algorithm \ref{algo:sleec_train}.

\small 
\begin{algorithm}[t!]
\caption{ {\tt SLEEC}: Training} 
\label{algo:sleec_train}
\begin{algorithmic}[]
    \State \textbf{Require : } $\mathcal{D} = \{(x_1,y_1), \ldots, (x_n,y_n)\},$ embedding dimensionality $\hat{L},$ number of neighbors $\bar{n},$ number of clusters $C,$ regularization parameter $\lambda, \mu$, L1 smoothing parameter $\rho$
    \State Partition $X$ into $Q^1, \ldots, Q^C$ using $k$-means
    \State \textbf{for} each partition $Q^j$
    \State \quad Form $\Omega$ using $\Bar{n}$ nearest neighbors of each label vector $y_i \in Q^j$
    \State \quad $[U\quad \Sigma] \leftarrow \text{SVP}(P_{\Omega}(Y^{jY^{j^T}}), \hat{L})$
    \State \quad $Z^j \leftarrow U\Sigma^{\frac{1}{2}}$
    \State \quad $V^j \leftarrow ADMM(X^j, Z^j, \lambda, \mu, \rho)$
    \State \quad $Z^j = V^j X^j$
    \State \textbf{end for}
    \State \textbf{return }$\{(Q^1, V^1, Z^1),\ldots, (Q^C, V^C, Z^C)\}$
\end{algorithmic}
\end{algorithm}
\normalsize 
\paragraph{}
The paper \citep{rembrandt_mineiro} proposes a method called REMBRANDT, which utilizes a correspondence between rank constrained estimation and low dimensional label embeddings to develop a fast label embedding algorithm.

The assumption made by the paper is that with infinite data, the matrix being decomposed is the expected outer product of the conditional label probabilities. In particular, this indicates that two labels are similar  when  their  conditional  probabilities  are  linearly  dependent  across  the dataset.


The label embedding problem is modeled as a rank constrained estimation problem, as described below. Given a data matrix $X$ and a label matrix $Y$, let the weight matrix to be learned have a low-rank constraint. In matrix form 

\begin{equation}
W^{*} = \underset{\text{rank}(W) \leq k}{\text{argmin}} ||Y - XW||^{2}_{F}. \label{eqn:rankest}    
\end{equation}

The solution to \eqref{eqn:rankest} is derived in \cite{rank_friedland}. The equation is further simplified to obtain a low rank matrix approximation problem, which is solved using a randomized algorithm proposed in \cite{random_halko}.

After the predictions are done in the embedding space, a decoding matrix is used to project the predicted label matrix back to the $\mathbb{R}^{d \times L}$ space. 

\paragraph{}
   \cite{pdsparse_ian} shows that for the extreme classification problem, a margin-maximizing loss with $l_1$ penalty, yields a very sparse solution both in primal and in dual space without affecting the predictor's expressive power. The proposed algorithm, PD-Sparse incorporates a Fully Corrective Block Coordinate Frank Wolfe (FC-BCFW) \cite{lacoste2015global} algorithm that utilizes this sparsity to achieve a complexity sublinear to the number of primal and dual variables. The main contribution of this method to the field is the use of the separation ranking loss in XMLC which was further used in several subsequent methods.
  
  Instead of making structural assumption on the relation between labels, for each instance, there are only a small number of correct labels and the feature space can be used to distinguish between labels. Under this assumption, a simple margin-maximizing loss can be shown to yield sparse dual solution in the setting of extreme classification. When this loss is combined with $l_1$ penalty, it gives a sparse solution both in the primal as well as dual for any $l_1$ parameter $\lambda$. 
  
  Let  $P(y) = \{l \in [L]| y_l = 1\}$ denote positive label indexes and $N(y) = \{l \in [L]| y_l = 0\}$ denote negative label indexes. In this paper, it is assumed that $nnz(y)$ is small and does not grow linearly with $L$.
  
\paragraph{}
  The objective of separation ranking loss is to compute the distance between the output and the target. The loss penalizes the prediction on an input $x$ by the highest response $(z)$ from the set of negative labels minus the lowest response from the set of positive labels.
  
  \begin{equation} \label{eqn:loss_function}
      L(z,y) = \underset{k_n \in N(y),\, k_p \in N(p)}{\max} (1 + z_{k_n} - z_{k_p})_{+},
  \end{equation}
where an instance has zero loss if all positive labels $k_p \in P_i$ have higher positive responses that that of negative labels $k_n \in N_i$ plus a margin. Here $(\, \cdot \,)_{+} = \max(0,.)$.

\begin{wrapfigure}{l}{6.5cm}
\includegraphics[scale=0.4]{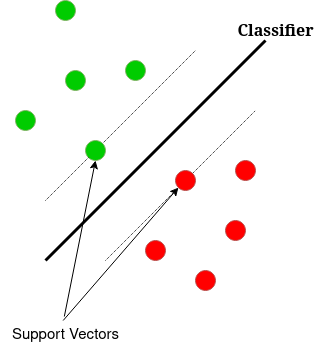}
\caption{Support Labels in the context of an SVM.}\label{fig:support_labels}
\end{wrapfigure} 
\paragraph{}
In a  binary classification scenario, the support labels are defined as 
 \begin{equation} \label{eqn:supp_label}
     (k_n,k_p) \in \underset{k_n \in N(y), \, k_p \in P(y)}{\text{argmax}}(1 + z_{k_n} - z_{k_p})_{+}.
 \end{equation}
Here, the labels that maximize the loss in \eqref{eqn:loss_function}  form the set of support labels. The basic inspiration for use of loss \eqref{eqn:loss_function} is that for any given instance, there are only a few labels with high responses and thus, the prediction accuracy can be boosted by learning how to distinguish between these labels. In the extreme classification setting, where $L$ is very high, only a few of them are supposed to give high response. These labels are taken as the support labels. 

The problem now becomes to find a solution to the problem 
\begin{equation} \label{eqn:final_loss}
     W^{*} = \underset{W}{\text{argmin}} \sum_{i=1}^{N} L(w^Tx_i, y_i) + \lambda\sum_{k=1}^{K} ||W_{k}||_{1}. 
\end{equation}
It can be shown that the solutions in both the primal and dual space is sparse in nature. Thus, sparsity assumptions on $W^*$ are not hindrances but actually a property of the solution. The above formalized problem is solved by separating blocks of variables and constraints to use the Block Coordinate Frank Wolfe Algorithm \citep{lacoste2013block}. The only bottleneck in the entire process comes down to how to find the most violating label in each step, for which an importance sampling based heuristic method is used.








\paragraph{}
In this paper \cite{reml_chang}, the idea of using low-rank structure for label correlations is re-visited keeping in mind the tail labels. The claim made in \cite{sleec_bhatia} which says that low-rank decompositions are not applicable in extreme classification is studied and a claim is made which says that low-rank decomposition is applicable after removal of tail labels from the data. Hence, in addition to low-rank decomposition of label matrix, the authors add an additional sparse component to handle tail labels behaving as outliers. 

Assuming the label matrix as $Y \in \{-1,+1\}^{n \times L},$ it is decomposed as 
$$ Y = \hat{Y}_L + \hat{Y}_S,$$
where $\hat{Y}_L$ is low rank and depict label correlations and $\hat{Y}_S$ is sparse, and captures the influence of tail labels. This can be modeled as 
\begin{equation}
     \begin{split}
        \underset{\underset{\underset{\text{card}(\hat{Y}_S) \leq s}{\text{rank}(\hat{Y}_L) \leq k,} }{\hat{Y}_L, \hat{Y}_S,} }{\min} \quad  &||Y - \hat{Y}_L - \hat{Y}_S||^{2}_{F}. \\    
     \end{split}
\end{equation}{}

Given $\hat{Y}_L$ and $\hat{Y}_S$, we wish to learn regression models such that $\hat{Y}_L \equiv WX$ and $\hat{Y}_S \equiv HX,$ where $W,H \in \mathbb{R}^{d \times L}$. 
Let $W = UV$ where $U \in \mathbb{R}^{d \times k}, V \in \mathbb{R}^{k \times L}.$ We impose the constraints by
 \begin{equation} \label{eqn:reml_obj}
     \begin{split}
         \underset{U,V,H}{\min}\quad ||Y - XUV - XH||_{F}^{2} + \lambda_1||H||_{F}^{2} +  \lambda_2(||U||_{F}^{2} + ||V||_{F}^{2}) + \lambda_3 ||XH||_{1}, \quad
         \lambda_1 , \lambda_2, \lambda_3 > 0.
     \end{split}
 \end{equation}
The objective \eqref{eqn:reml_obj} can be divided into three subproblems: 

\begin{align} 
    &V = \underset{V}{\text{argmin}} \: ||Y - XUV - XH||_{F}^{2} + \lambda_2||V||_{F}^{2}.  \label{eqn:reml_obj_1} \\
    &U = \underset{U}{\text{argmin}} \: ||Y - XUV - XH||_{F}^{2} + \lambda_2||U||_{F}^{2}. \label{eqn:reml_obj_2} \\
    &H = \underset{H}{\text{argmin}} \: ||Y - XUV - XH||_{F}^{2} + \lambda_1||H||_{F}^{2} +  \lambda_3||XH||_{1}.  \label{eqn:reml_obj_3}
\end{align}    
Each of these sub problems can be solved by equating the partial differentiation to zero. 

\paragraph{}
The sub problems \eqref{eqn:reml_obj_1},\eqref{eqn:reml_obj_2} and \eqref{eqn:reml_obj_3} can be efficiently solved in parallel and their solutions can be combined to achieve the final solution. This is a divide and conquer strategy. 

\paragraph{}
For the divide step, given the label matrix $Y \in \mathbb{R}^{n \times L},$ it is randomly partitioned into $t$ number of $m$-column sub matrices $\{(Y)_i\}_{i=1}^{t},$ where we suppose $L = tm$ and each $(Y_i) \in \{-1,1\}^{n \times m}.$ Thus, the original problem is divided into $t$ sub-problems regarding $\{(Y)_1, \cdots, (Y)_t\}.$ The basic optimization methods discussed in the previous section can now be adopted to solve these sub-problems in parallel, which outputs the solutions as $\{((\hat{W})_1,(\hat{H})_1),\cdots,((\hat{W})_t,(\hat{H})_t)\}.$

\paragraph{}
The conquer step exploits column projection to integrate the solutions of sub-problem solutions. The final approximation $W$ to problem \eqref{eqn:reml_obj} can be obtained by projecting $[(\hat{W}_1),\cdots,(\hat{W}_t)]$ onto the column space of $(\hat{W}_1).$ After obtaining $\hat{W},$ the objective can be optimized over each column of $\hat{H}$ in parallel to obtain the sparse component of the resulting multi-label predictor.


\small 
\begin{algorithm}[t!]
\caption{ REML}
\begin{algorithmic}[1]
    \State \textbf{Input:} $X,Y,t \geq 1$
    \State for $i = 1, \cdots, t$ do in parallel
    \State \quad Sample $(Y)_i \subseteq Y$
    \State \quad repeat
    \State \quad \quad Solve $(\hat{V})_i$ from problem \eqref{eqn:reml_obj_1}
    \State \quad \quad Solve $(\hat{U})_i$ from problem \eqref{eqn:reml_obj_2}
    \State \quad \quad Solve $(\hat{H})_i$ from problem \eqref{eqn:reml_obj_3}
    \State \quad until Convergence
    \State \quad $(\hat{W})_i = (\hat{U})_i(\hat{V})_i$
    \State end
    \State {\tt ColumnProjection} $([(\hat{W})_1,\cdots,(\hat{W})_t],(\hat{W})_1)$
\end{algorithmic}
\end{algorithm}
\normalsize 
\paragraph{}
In \cite{AnnexML}, the author highlights some major problems that the SLEEC \citep{sleec_bhatia} algorithm has. SLEEC uses only feature vectors for $k$-means clustering. It does not access label information during clustering. Thus data points having the same label vectors are not guaranteed to be assigned to the same partition. As the number of partitions increase, the data points having the same tail labels may be assigned to different partitions.
Further, in the prediction step, SLEEC predicts labels of the test point by using the nearest training points in the embedding space. Hence, the order of distance to the neighbors of the test point is considered, and not the values themselves. This may cause an isolated point which may be far from a cluster center to belong to that cluster.


\paragraph{}
The proposed method, AnnexML is able to solve these issues by learning a multi-class classifier, which partitions the approximate $k$-Nearest Neighbor Graph (KNNG) \citep{knng_dong} of the label vectors in order to preserve the graph structure as much as possible. AnnexML also projects each divided sub-graph into an individual embedding space by formulating this problem as a ranking problem instead of a regression one. Thus the embedding matrix $V_c$ is not a regressor like in SLEEC.

\paragraph{}
There are mainly three steps to the training of the AnnexML algorithm. Firstly, AnnexML learns to group the data points that have similar label vectors to the same partition, which is done by utilizing the label information.  It forms a $k$-nearest neighbor graph (KNNG) of the label vectors, and divides the graph into $K$ sub-graphs, while keeping the structure as much as possible. After the KNNG has been computed and partitioned, we learn the weight vector $w_c$ for each partition $c$ such that the data points having the same tail labels are allocated in the same partition while the data points are divided almost equally among partitions. Thirdly, the learning objective of AnnexML is to reconstruct the KNNG of label vectors in the embedding space. For this task, it employs a pairwise learning-to-rank approach. That is, arranging the $k$ nearest neighbors of the $i$-th sample is regarded as a ranking problem with the $i$-th point as the query and other points as the items. The approach used is similar to DSSM \citep{dssm_yih}.

\paragraph{}
The prediction of AnnexML mostly relies on the $k$-nearest neighbor search in the embedding space. Thus, for faster prediction, we need to speed up this NN search. An approximate $k$-nearest neighbor search method is applied, which explores the learned KNNG in the embedding space by using an additional tree structure and a pruning technique via the triangle inequality. This technique gives AnnexML a much faster prediction than SLEEC.

\paragraph{}

In the paper by \cite{defrag_jalan}, a new method for reducing the dimensionality of the feature vectors is explored which is especially useful when the feature vector is already sparse. This is not a specific XMLC method but a {\em add-on method} for any XML algorithm which can be used to provide speedups when the number of features is very high with minimal loss of accuracy.

The reduction in dimensionality is performed by a method called feature agglomeration (see Figure~\ref{fig:feature_agg}), where clusters of features are created and then a sum is taken to create new features. If $F$ is partition of the features $[d]$, then corresponding to every cluster $F_K \in F$ we create a single feature by summing up the features within cluster $F_k$. Thus, for a given vector $z \in \mathbb{R}^d,$ we create an agglomerated vector $\Tilde{z} \in \mathbb{R}^K$ with just $K$ features using the clustering $F$. The $k-$th dimension of $\Tilde{z}$ will be $\Tilde{z}_k = \Sigma_{j \in F_k}z_j$ for $k = 1 \ldots K.$

\begin{figure}[t!]
 \centering
\begin{tikzpicture}[
    >=Stealth,
    every node/.style={font=\footnotesize},
    feat/.style={minimum width=0.9cm,minimum height=0.55cm,draw},
    seg/.style={minimum width=1.8cm,minimum height=0.55cm,draw},
    arr/.style={->,very thin}
]

\foreach \i/\txt/\col in {
      0/{1}/blue!30, 1/{2}/blue!30, 2/{}/blue!15, 3/{}/blue!15,
      4/{}/green!35,5/{}/green!35,6/{}/green!35,
      7/{}/yellow!35,8/{}/yellow!35,9/{}/yellow!35,
      10/{}/red!30,11/{N}/red!30}
  \node[feat,fill=\col] (t\i) at (\i*0.9,0) {\txt};

\node[font=\large] (p1) at ( 0.9*1.5,-1.2) {$+$}; 
\node[font=\large] (p2) at ( 0.9*5.0,-1.2) {$+$}; 
\node[font=\large] (p3) at ( 0.9*8.0,-1.2) {$+$}; 
\node[font=\large] (p4) at ( 0.9*10.5,-1.2) {$+$};

\foreach \k in {0,1,2,3}   \draw[arr] (t\k.south)  -- (p1);
\foreach \k in {4,5,6}     \draw[arr] (t\k.south)  -- (p2);
\foreach \k in {7,8,9}     \draw[arr] (t\k.south)  -- (p3);
\foreach \k in {10,11}     \draw[arr] (t\k.south)  -- (p4);

\begin{scope}[xshift=2.2cm] 

  \node[seg,fill=blue!30]   (s1) at ( 0.0,-2.4) {1};
  \node[seg,fill=green!35]  (s2) at ( 1.8,-2.4) {};
  \node[seg,fill=yellow!35] (s3) at ( 3.6,-2.4) {};
  \node[seg,fill=red!30]    (s4) at ( 5.4,-2.4) {K};

  \node[draw,rounded corners,fit=(s1)(s4),inner sep=0pt] {};
\end{scope}

\draw[arr] (p1) -- (s1.north);
\draw[arr] (p2) -- (s2.north);
\draw[arr] (p3) -- (s3.north);
\draw[arr] (p4) -- (s4.north);
\end{tikzpicture}
 \caption{\label{fig:feature_agg}%
    Illustration of the adaptive feature-agglomeration step in
    DEFRAG \citep{defrag_jalan}. The coloured bar across the top is the original
    sparse TF-IDF vector with \(N\) coordinates. DEFRAG groups adjacent features
    with highly correlated activations and repeatedly merges the two most-similar
    groups by summing them until only \(K\ll N\) groups remain (lower bar).
    For example, starting from the 10-dimensional vector
    \([f_1,f_2,\dots ,f_{10}]\), the algorithm successively fuses
    \(\{f_1,f_2\}\), then \(\{f_3,f_4,f_5\}\), then \(\{f_6,f_7,f_8\}\), and
    finally \(\{f_9,f_{10}\}\), yielding the four-dimensional surrogate
    \([f'_1=f_1+f_2,\; f'_2=f_3+f_4+f_5,\; f'_3=f_6+f_7+f_8,\;
      f'_4=f_9+f_{10}]\).
    These \(K\) aggregates capture local co-occurrence patterns, lower memory and
    training cost, and feed the downstream one-vs-all classifiers; unlike SLICE,
    which retains the raw features but samples negatives aggressively, DEFRAG
    achieves scalability primarily through this dimensionality reduction.}

\end{figure}


\paragraph{}
The proposed method, DEFRAG, clusters the features into balanced clusters of size not more than $d_0$. If this results in $K$ clusters, then we sum up the features in these clusters to obtain $K$-dimensional features which are subsequently used for training and testing using any XMLC method. It can be theoretically proven that with a linear classification method, feature agglomeration cannot give worse performance than the original method.


\begin{algorithm}[t!] 
\caption{ DEFRAG: Make Tree}
\label{algo:defrag}
\begin{algorithmic}[1]
    \State If $|S| \leq d_0$ then
    \State \quad $n \leftarrow \text{Make-Leaf}(S)$
    \State else
    \State \quad $n \leftarrow \text{Make-Internal-Node}()$
    \State \quad $\{S_{+},S_{-}\}\leftarrow \text{Balanced-Split}(S,\{z^i, i \in S\})$ // Balanced Spherical kmeans or nDCG split
    \State \quad $n_{+} \leftarrow \text{Make-Tree}(S_{+},\{z^i,i \in S_{+}\}, d_0)$
    \State \quad $n_{-} \leftarrow \text{Make-Tree}(S_{-},\{z^i,i \in S_{-}\}, d_0)$
    \State \quad $n.\text{left\_child} \leftarrow n_+$
    \State \quad $n.\text{right\_child} \leftarrow n_-$
    \State \textbf{return } Root node of the tree
\end{algorithmic}
\end{algorithm}

    

DEFRAG first creates a representation vector for each feature $j \in [d]$ and then performs hierarchical clustering on them. At each internal node of the hierarchy, features at that node are split into two children nodes of equal sizes by solving either a balanced spherical 2-means problem, or by minimizing nDCG. This process is continued until we are left with less than $d_0$ features at a node, in which case the node is made a leaf. The representative vectors can be either based on features only or labels and features together which can perform different functions.



\paragraph{}
The use of DEFRAG for dimensionality reduction has several advantages over the classical dimensionality reduction techniques such as PCA or random projections for high-dimensional sparse data. 
Feature agglomeration in DEFRAG involves summing up the coordinates of a vector belonging to the same cluster. This is much less computationally expensive than performing PCA or random projections.
Also, the feature agglomeration done in DEFRAG does not densify the vector. If the sparsity of a vector is $s$, then the $K$-dimensional vector resulting from the agglomeration will have at most $s$ non-zeros ($s \leq K, s = K$ if one feature from each cluster has a non-zero). This is unlike PCA or random projections which densify the vectors which in turn cause algorithms such as SLEEC or LEML to use low dimensional embeddings. This results in a loss of information. Thus, using DEFRAG preserves much of the information of the original vectors and also offers speedup due to reduced dimensionality.

\paragraph{}
The recent paper \cite{Pavlovski2023} introduces the Multi-Label Factorization Machine (MLFM) model (which can be seen as low rank decomposition), designed to handle Extreme Multi-Label Classification problems effectively, with a specific focus on {\em Behavioral Ad Targeting}. The model addresses the issue of feature interactions and aims to reduce the original time complexity associated with factorization machines \citep{rendle2010}. The authors propose an innovative lightweight formulation to make the computation more efficient. Additionally, the paper discusses the incorporation of Field-weighted Factorization Machines (FwFM) \citep{pan2018} and field-level interactions to enhance the model's capability.

\paragraph{}
The MLFM model starts by embedding features into a lower-dimensional space, which is essential for managing the high-dimensional data typical in behavioral ad targeting. Each feature is represented by a binary variable and assigned an embedding vector, initialized randomly. This step transforms raw features into a form that can be more efficiently processed, significantly enhancing the model's capacity to learn from large datasets.

\paragraph{}
The MLFM model considers second-degree feature interactions to manage a large number of features. The decision function $\phi(\mathbf{x})$ of a second-degree MLFM is defined as:
\begin{align}
\mathbf{w}_0+\left[\sum_{i=1}^D w_i^l x_i\right]_{l=1}^L+\left[\sum_{i=1}^D \sum_{j=i+1}^D r_{F(i) F(j)}^l\left\langle\mathbf{v}_i, \mathbf{v}_j\right\rangle x_i x_j\right]_{l=1}^L, \label{eqn:second-degree}
\end{align}
where $w_0 \in \mathbb{R}^L$ is a bias vector, $W = \left[w_i^l\right]_{D \times L}$ are the feature weights over $L$ class labels; $F(i)$ and $F(j)$ denote the fields that features $x_i$ and $x_j$ belong to, respectively, thus $R = \left[r_{F(i) F(j)}^l\right]_{\hat{D} \times \hat{D} \times L},$ where $\hat{D}$ denotes the number of different fields; and $V = \left[v_{i m}\right]_{D \times M}$ is the feature embedding matrix shared among all labels. $\left\langle v_i, v_j\right\rangle$ represents the dot product between the embeddings of two features, for each $i, j=1, \ldots, D$. This extends the capacity of the linear formulation given by the first two terms in Eqn \eqref{eqn:second-degree}, and allows for modeling between-feature as well as between-field interactions.

\paragraph{}
One of the significant contributions of the paper is the analysis of the model's complexity. The original MLFM model has a quadratic time complexity with respect to the number of features, which can be prohibitive for large datasets. The authors address this by proposing a lightweight formulation that reduces the time complexity to linear in the number of features. This improvement is achieved by factorizing the field interactions, transforming the interaction terms into a more computationally efficient form as follows
\begin{align*}
\left[\frac{1}{2} \sum_{k=1}^{H * M}\left(\left(\sum_{i=1}^D q_{i k}^l x_i\right)^2-\sum_{i=1}^D\left(q_{i k}^l x_i\right)^2\right)\right]_{l=1}^L, 
\end{align*}
where $q_{i k}^l$ are the interaction factors for the embeddings. This decomposition simplifies the computation significantly. The time complexity of the new formulation is linear in terms of the number of features \(D\) and the dimensions of the feature and field embeddings \(M\) and \(H\), specifically \(O(DMH)\). Considering all \(L\) labels, the complexity becomes \(O(LDMH)\). Given that \(H\) is typically of the same order of magnitude as \(M\) or lower, the total complexity can be approximated and further reduced in sparse multi-field categorical data scenarios to \(O(L \hat{D} MH)\), where \(\hat{D}\) is the reduced number of active features per field.

\paragraph{}
The paper describes the parameter learning process using a labeled dataset. The categorical cross-entropy loss function is employed, with a sigmoid function to handle the multi-label nature of the problem. The optimization process involves minimizing this loss over all data points, using gradient-based methods to iteratively refine the model parameters. This systematic approach ensures that the model converges to an optimal set of parameters, improving its predictive accuracy.\\
\paragraph{Comparison between above stated methods}
\begin{enumerate}
    \item Method LEML outperforms all of the compressed sensing based methods in terms of performance and subsequent methods based on LEML hence outperform compressed sensing based methods.
    \item Method SLEEC was the first to find the issue with low-rank decomposition based methods proposed as compressed sensing based methods. Using a distance preserving embedding, SLEEC achieved a large improvement over previous methods. AnnexML improved on SLEEC further improving both performance and prediction times.
    \item Method REML revisits the idea of low-rank embeddings by separating out the tail labels and learning different predictors for head and tail labels.
\end{enumerate}

\begin{table}[t!]
\footnotesize
\centering 
\begin{tabular}{ |p{3cm}||p{3cm}|p{7cm}|  }
 \hline
 \multicolumn{3}{|c|}{Linear Algebra Based Methods} \\
 \hline
  Algorithm & Type of Method & Summary\\
 \hline
 MLCSSP & Subset Selection & The maximum spanning subset of the label matrix is chosen i.e., a label subset is chosen which best represents the entire label space. Prediction is performed in this label space and the results are then extrapolated on the entire label set. \\ \hline 
 LEML   & Low rank decomposition & The low-rank decomposition of the label matrix is assumed and a general ERM framework is proposed to solve it. Attempts to generalize compressed sensing based methods like CPLST and apply them for missing labels too. \\ \hline 
 SLEEC & Distance preserving embeddings & A method for generating embeddings which preserve the distance between labels in the generated embedding space. The prediction is done via KNN in the embedding space. \\ \hline 
 REMBRANDT & Low rank decomposition & Uses a randomized low-rank matrix factorization based technique to perform rank-constrained embedding. It can be shown to be equivalent to CPLST in the simplest case. \\ \hline 
 PD-SPARSE & Separation Ranking Loss based Frank Wolfe & Utilizes the dual and primal sparsity of a solution to the problem with separation ranking loss and $L1$ regularization. The problem is solved via a Fully Corrective Block Coordinate Frank Wolfe Algorithm. \\ \hline 
 REML & Low rank decomposition & Re-visits the Low rank decomposition technique and tries to resolve the issue tail labels create by separating head and tail labels. This makes the low-rank assumption valid once again and head and tail labels are predicted separately. \\ \hline 
 AnnexML &Distance preserving embedding & Distance preserving embeddings and improvement over the SLEEC algorithm by using a KNN graph for partitioning labels, using a ranking based loss and using an ANN based prediction technique. \\ \hline 
 DEFRAG & Feature Agglomeration Supplement & An algorithm for dealing with large feature spaces in extreme classification by performing feature agglomeration. Can also be used for feature imputation and label re-ranking.\\
 \hline
\end{tabular}
\end{table}
\normalsize

\subsection{Tree-Based Methods}
\paragraph{}
Tree Based methods rely on the inherent hierarchy of the data to speed up the multi-label learning and prediction. The main goal of these methods is to repeatedly divide either the label or sample space in order to narrow down the search space during prediction. The leaf nodes generally contain a one-versus-all or a simple averaging based technique for final prediction.

Broadly, these techniques can be categorized into two types :
\begin{enumerate}
    \item \textbf{Label Partitioning based.} These methods rely on dividing the labels into clusters at each level, thus forming a sort of meta-label structure. Each of the levels is then equipped with a multi-label classifier to determine the most relevant meta labels. The leaves contain a small number of labels which can be dealt with by a one-vs-all classifier.
    \item \textbf{Sample/Instance Partitioning Based.} These methods group together the train samples and cluster them into branches at each level. This forms a sort of decision tree structure which can be used to group the incoming test point into the correct branches and finally reach the leaf. The branch decisions are made by a multi-class classifier and the final predictions on the leaf are then made by combining the labels of the instances at that leaf.
\end{enumerate}

\begin{figure}[t!]
    \centering
    \includegraphics[width=80mm]{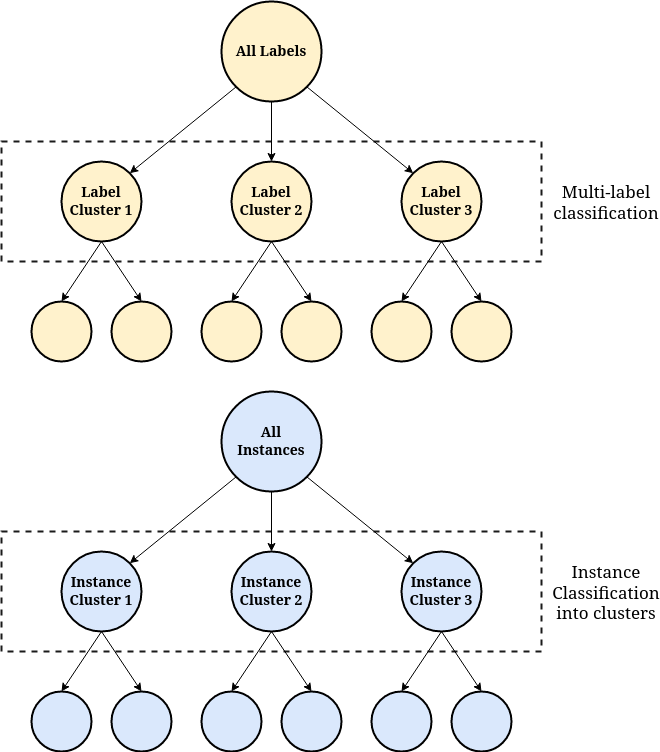}
    \caption{Comparison of label and instance partitioning methods. Label partitioning methods (top left) cluster labels to convert XMLC into a series of multi-label classification problems. Instance partitioning methods cluster samples and allow KNN-like approaches at the leaf.}
\end{figure}{}

\paragraph{}
Tree Based methods in general provide faster training and prediction than the embedding and deep-learning based methods due to the reduction of training and search space provided by the tree structure.


\paragraph{}
The paper \cite{prabhu2014fastxml} proposes a method called FastXML which improves upon the state of the art tree based multi label classification methods of the time, Multi-Label Random Forest Classifier (MLRF) \citep{liu2015mlrf} and Label Partitioning by Sub-linear Ranking (LPSR) \citep{weston2013label}. Both these methods learn a hierarchy of the labels to deal with the large number of labels. The LPSR learns relaxed integer programs to partition the sample into the clusters learnt by k-means like algorithm. MLRF on the other hand learn an ensemble of randomized trees but uses a brute force partitioning method based on Gini index or entropy. In contrast, FastXML learns to partition samples or data points rather than labels.

\paragraph{}
Both methods MLRF and LPSR did not perform competitively on XMLC datasets when compared with existing methods. Significantly higher accuracy is achieved by FastXML by optimizing a nDCG loss along with learning the ranking of the labels in each partition at the same time. An alternating minimization algorithm for efficiently optimizing the problem formulation is also developed. 

\paragraph{}
FastXML proposes to learn the hierarchy by directly optimizing the ranking loss function. This way of learning the hierarchy is better as nDCG is a measure which is sensitive to both relevance and ranking ensuring that the relevant positive labels are predicted with the highest possible ranks. This is not guaranteed by rank insensitive measures such as Gini-index in MLRF or clustering error in LPSR.

\paragraph{}
FastXML partitions the current node feature space by learning a linear separator $w$ such that
\begin{align*}
    \min \, {||w||}_{1} + \sum_{i} C_{\delta}(\delta_{i}) \log(1+e^{\delta_{i}w^{T}x_{i}}) &- C_{r}\sum_{i} \frac{1}{2}(1+\delta_{i}) L_{\text{nDCG}_{@L}(r^{+},y_{i}} \\
    &-  C_{r}\sum_{i} \frac{1}{2}(1-\delta_{i}) L_{\text{nDCG}_{@L}(r^{-},y_{i}}),
\label{fastxmlcost}
\end{align*}
where $w \in \mathcal{R}^{D}, \delta \in \{-1,+1\}_{L}, r^{+}, r^{-} \in \Pi (1,L) $ where $i$ represents indices of all the training points present at the node being partitioned. $\delta_{i} \in \{-1,+1\}$ is indicative of whether point $i$ was assigned to the negative r positive partition . The positive and negative partition are represented by variables $r^{+}$ and $ r^{-}$ respectively. User defined parameters $C_{\delta}$ and $C_{r}$ are used to determine the relative importance of the three terms.

\paragraph{}
There are three types of parameters in the equation, $w$, $\delta$ and $r^{+}, r^{-}$. Though the other two parameters can be obtained from $w$ itself, it is easier to optimize with separate parameters. The first term in the loss is an $l_{1}$ term on $w$ which ensures sparsity. The second term is the log loss of $\delta_{i}w^{T}x_{i}$. This term optimizes the parameters $\delta$ and $w$ together, as the solution to be obtained is $\delta_{i}^{*}= $ sign$(w^{T}x_{i})$. This tries to make sure that a point which is assigned to the positive (negative) partition, i.e points for which $\delta_{i} = +1(\delta_{i}=-1),$ will have positive (negative) values of $w^{T}x_{i}$. The third and fourth terms in the equation maximize the $\text{nDCG}_{@L}$ of the rankings predicted for the positive and negative partitions, $r^{+}$ and $r_{-}$ respectively. These terms relate $r^{+}$ to $\delta$ and thus to $w$.


\paragraph{}
Maximizing nDCG makes it likely that the relevant positive labels for each point are predicted as high as possible. As a result, points within the same partition are more likely to have similar labels whereas points in different partitions are more likely to have dissimilar labels.


\paragraph{}
The loss also allows a label to be assigned to both partitions if some of the points which contain the label are assigned to the positive partition and some to the negative. This makes the FastXML trees more robust as the error which is propagated from the parents may be corrected by the children.


\paragraph{}
Once we have learned the hyperplane during the training phase, then we know how to partition a test instance into a left or a right child. So, now we know how to traverse the tree starting at the root and keep applying this procedure recursively until we get to a leaf. 

Given a new data point for prediction, $x \in \R^{D}$, \textsc{FastXML's} top ranked $k$ predictions are given by 

\begin{equation}
    r(x) = \text{rank}_k \left( \frac{1}{T} \sum_{i=1}^{T} P_{t}^{\text{leaf}}(x) \right), 
\end{equation}
where $T$ is the number of trees in the \textsc{FastXML} ensemble and $P_{t}^{\text{leaf}}(x)$ is the distribution of points in the lead nodes containing $x$ in tree $t$. It is found empirically, that the proposed \textsc{nDCG} based objective function learns balanced which leads to fast and accurate predictions in logarithmic time. The main reason behind the accurate prediction is the novel node partitioning formulation which optimized an \textsc{nDCG} based ranking loss over all the labels. Such a loss turns is more suitable for XMLC than the Gini index used by \textsc{MLRF} or the clustering error used by \textsc{LPSR} and efficiently scales to problems with more than millions of labels.


\paragraph{}
In the paper \cite{prabhu2018parabel}, a label tree partitioning based method, Parabel, is proposed which aims to achieve the high prediction accuracy achieved by one-vs-all methods such as DiSMEC \citep{babbar2017dismec} and PPDSparse \citep{PPD_Sparse} while having fast training and prediction times. Parabel overcomes the problem of large label spaces by reducing the number of training points in each one vs all classifier such that each label's negative training examples are those with most similar labels. This is performed using the tree structure learnt.

    
    


In Parabel, labels are recursively partitioned into two groups to create a balanced binary tree. After the partitioning, similar labels end up in a group. Partitioning is stopped when the leaf node contains at-least $M$ labels. After the tree construction, two binary classifiers have to be trained at each internal node. These two classifiers help in determining whether the test point has to be sent to left or the right child or both. At the leaf node One vs All classifiers are trained for each label associated with that leaf node. At the time of prediction, a test sample is propagated through the tree using classifiers at internal nodes, after reaching leaf nodes, ranking over the labels is assigned using the one vs all classifiers present at the leaf.


\paragraph{}
The labels are represented by the normalized sum of training instances to in which it appears. This label representation is used to cluster the labels using a $2$-means objective along with an extra term to ensure that the label splits are uniform. Further, the distance between labels is measured using the cosine function, which basically leads to a spherical k-means objective with $k=2$.
This objective is solved using the well-known alternating minimization. 

\paragraph{}
Using this splitting procedure recursively for each internal node of the tree starting from the root, we can partition the labels to create a balanced binary tree. Partitioning is stopped when there are less than $M$ number of labels in a cluster and they become leaf nodes. In the final tree structure we end up with, the internal nodes represent meta-labels, and the problem at each node becomes a smaller multi-label classification problem.

\paragraph{}
After the tree structure is learnt, we need to train a classifier at each edge for predicting the probability of the child being active given that the parent is active. Let $ z_n $ be a binary variable associated with  each node $n$ for each sample. For a given sample , if $ z_n  = 1 $ then the sub-tree rooted at node $n$ has a leaf node which contains a relevant label of the given sample. Thus we need to predict $P(z_{iCn}=1 \mid z_{in}=1, x)$ at each internal node, where $C_n$ represents a child of node $n$. This can be done by optimizing the MAP estimate 
\begin{equation}
    \underset{\Theta_{I_n}}{\min} \quad -\log P(\Theta_{I_n}) - \sum_{i:z_{in}} \log P \left (  z_{iCn} \mid z_{in} = 1, x_i, \Theta_{I_n},    \right ),
\end{equation}
which is parameterized by $\Theta_{I_n}$.


\paragraph{}
Similarly, at the leaf nodes one has to minimize the loss 
\begin{equation}
    \underset{\Theta_{L_n}}{\min} \quad -\log P(\Theta_{L_n}) - \sum_{i:z_{in}} \log P \left (  y_{in} \mid z_{in} = 1, x_i, \Theta_{L_n}    \right ),
\end{equation}
which basically trains a one-vs-all classifier for each label present in that leaf.


\paragraph{}
Highly parallel and distributed versions of Parabel are possible because all the optimization problems are independent. The independence does not imply that model does not recognize label to label dependency or correlation. In fact, the model does take into account label-label correlation when clustering the labels. At the time of clustering similar labels are grouped into a cluster, thus model is not truly independent with respect to labels.

\paragraph{}
During prediction, the goal is to find the top $k$ labels with the highest value of 
\begin{equation}
    P(y_l \mid x) = P(y_l=1 \mid z_n=1,x) \, \Pi_{n_0 \in A_n}P(z_{n_0}=1 \mid z_{P_{n_0}}=1,x),
\end{equation}
where $P_n$ is the parent of node $n.$ This means that we have to find the path to the leaf from the root with the maximum value of product of probabilities on its nodes. We can use a beam search (greedy BFS traversal) to find the top $k$ labels with the highest probabilities.

\paragraph{}
The paper \cite{craftml_siblini} proposes a random forest based algorithm with a fast partitioning algorithm called CRAFTML (Clustering based RAndom Forest of predictive Trees for extreme Multi-label Learning). The method proposed is also an instance/sample partitioning based method like FastXML but it uses a k-means based classification technique in comparison to nDCG optimization in FastXML. Another major difference with FastXML is that it uses $k$-way partition in the tree instead of 2-way partitions. CRAFTML also exploits a random forest strategy which randomly reduces both the feature and the label spaces using random projection matrices to obtain diversity.


\paragraph{}
CRAFTML computes a forest $F$ of $m_F$ $k$-ary instance trees which are constructed by recursive partitioning. The base case of the recursive partitioning is either of the three $(i)$ the number of instances in the node is less than a given threshold $n_{\text{leaf}}, (ii)$ all the instances have the same features, or $(iii)$ all the instances have the same labels. After the tree training is complete, each leaf stores the average label vector of the samples present in the leaf. 

\paragraph{}
The node training stage in CRAFTML is decomposed into three steps. First, a random projection of the label and feature vectors into lower dimensional spaces is performed. This is done using a projection matrix which is randomly generated from either Standard Gaussian Distribution or sparse orthogonal projection. Next, $k$-means is used for recursively partitioning the instances into $k$ temporary subsets using their projected label vectors. We thus obtain $k$ cluster centroids. The cluster centroids are initialized with the $k$-means++ strategy which improves cluster stability and algorithm performances against a random initialization. Finally, the cluster centroids are re-calculated by taking the average of feature vectors assigned to the temporary clusters. We now have a $k$-means classifier which is capable of classification by finding the closest cluster to any instance. The instances are partitioned into $k$ final subsets (child nodes) by the classifier.


The training algorithm is given below.
\small 
\begin{algorithm}[t!]
\caption{ trainTree}
\begin{algorithmic}[t!]
    \State \textbf{Input:} Training set with a feature matrix $X$ and a label matrix $Y.$
    \State Initialize node $v.$
    \State $v.$isLeaf $\leftarrow$ testStopCondition($X,Y$)
    \State if $v.$isLeaf = false then
    \State \quad $v.$classif $\leftarrow$ trainNodeClassifier($X,Y$)
    \State \quad $(X_{child_i},Y_{child_i})_{i=0,\cdots,k-1}$ $\leftarrow$ split($v.$classif,$X,Y$)
    \State \quad for $i$ from $0$ to $k-1$ do
    \State \quad \quad $v.child_i \leftarrow$ trainTree($X_{child_i},Y_{child_i}$)
    \State \quad end for
    \State else
    \State \quad $v.\hat{y} \leftarrow $ computeMeanLabelVector($Y$)
    \State end if
    \State \textbf{Output:} node $v$
\end{algorithmic}
\end{algorithm}
\normalsize 

\small 
\begin{algorithm}[t!]
\caption{trainNodeClassifier}
\begin{algorithmic}[t!]
    \State \textbf{Input:} feature matrix $(X_v)$ and label matrix $(Y_v)$ of the instance set of the node $v.$
    \State $X_s,Y_s \leftarrow$ sampleRows($X_v,Y_v,n_s$)   \Comment{$n_s$ is the sample size}
    \State $X_{s}^{'} \leftarrow X_sP_x$   \Comment{random feature projection}
    \State $Y_{s}^{'} \leftarrow Y_sP_y$   \Comment{Random label projection}
    \State $c \leftarrow k$-means($Y_{s}^{'},k$)  \Comment{$c \in \{0,\cdots,k-1\}^{\min(n_v,n_s)}$} \\
    \Comment{$c$ is a vector where the $j^{th}$ component $c_j$ denotes the cluster idx of the $j^{th}$ instance associated to $(X_{s}^{'})_{j,.}$ and $(Y_{s}^{'})_{j,.}.$}
    \State for $i$ from $0$ to $k-1$ do
    \State \quad $(classif)_{i,.} \leftarrow $computeCentroid(${(X_{s}^{'})_{j,.}| c_j = i}$)
    \State end for
    \State \textbf{Output:} Classifier $classif(\in \mathbb{R}^{k \times d_{x}^{'}})$
\end{algorithmic}
\end{algorithm}
\normalsize

\paragraph{}
During prediction, we traverse from the root to the leaves using the $k$-means classifier decisions at each layer. Once a leaf is reached, prediction is given by the average of the label vectors of instances present at that leaf. Using the forest, we aggregate the predictions which largely helps increase the accuracy.

\paragraph{}
\cite{khandagale2020bonsai} introduces a method Bonsai, which improves on Parabel \citep{prabhu2018parabel}. The authors make the observation that in Parabel, due to its deep tree structure there is error propagation down the tree, which means that the errors in the initial classifiers will accumulate and cause higher classification error in leaf layers. Also, due to the balanced partitioning in the tree layers, the tail labels are forced to be clustered with the head labels, which causes the tail labels to be subsumed. Hence, in Bonsai, a shallow tree structure with $\ge 2$ children for each node is proposed which ($i$) prevents ``cascading effect" of error in deep trees. ($ii$) allows more diverse clusters preventing misclassification of tail labels.

\paragraph{}
Another contribution of the paper is to introduce different types of label representations based on either the feature-label correlations, label-label co-occurrence or a combination of both. 

Let $ {(x_i,y_i)_{i=1}^{N}} $ be the given training samples with $ x_i \in R^D $ and $ y_i \in \{ 0,1 \}^L $. 
$ X_{N \times D} $, $ Y_{N \times L }  $ are data and label matrices respectively. Let $ v_l $ be the representation of label $l.$

There are three ways to represent the labels before clustering.

 In the input space representation, $ v_l = \sum_{i=1}^N y_{il}x_i $, this can be compactly written as $ V = Y^{T} X $, where each row $ V_{p:} $ corresponds to the label representation of the label, i.e. $ p^{th} $ row is the label representation of label number $p$ ($ V_{p:} = v_p $). 
\begin{align}
    \nu_i = Y^TX  = \begin{bmatrix}
                        v_1^T\\
                        v_2^T\\
                        \vdots \\
                        v_L^T\\
                     \end{bmatrix}_{L \times D}, \quad 
                     \text{where} \quad
            X = \begin{bmatrix}
                    x_1^T\\
                    x_2^T\\
                    \vdots\\
                    x_N^T\\
                \end{bmatrix}_{N \times D}
            Y = \begin{bmatrix}
                    y_1^T\\
                    y_2^T\\
                    \vdots\\
                    y_L^T\\
                \end{bmatrix}_{N \times L}.
\end{align}

In the output space representation, the Label representation matrix $ \nu_o $ is given by 
\begin{align}
    \nu_o = Y^TY & = \begin{bmatrix}
                        v_1^T\\
                        v_2^T\\
                        \vdots \\
                        v_L^T\\
                     \end{bmatrix}_{L \times L}, \quad \text{where} \quad
            Y = \begin{bmatrix}
                    y_1^T\\
                    y_2^T\\
                    \vdots\\
                    y_L^T\\
                \end{bmatrix}_{N \times L}.
\end{align}


In the joint input-output representation, the Label representation matrix $ \nu_{io} $ is given by
\begin{align}
    \nu_{io} = Y^TZ = \begin{bmatrix}
                            v_1^T \\
                            v_2^T \\
                            \vdots\\
                            v_L^T \\
                      \end{bmatrix}_{L \times (D+L)}, \quad \text{where} \quad
                Z = \begin{bmatrix}
                        z_1^T\\
                        z_2^T\\
                        \vdots\\
                        z_L^T\\
                    \end{bmatrix}_{N \times (D+L)}
                Y = \begin{bmatrix}
                        y_1^T\\
                        y_2^T\\
                        \vdots\\
                        y_L^T\\
                    \end{bmatrix}_{N \times L}.
\end{align}

Here the matrix $Z$ is obtained by concatenating the vector $X$ and vector $Y,$ hence the dimension $ N \times (D+L).$ 
%
%
%
%
The training and prediction algorithms for Bonsai remain the same as Parabel, except that any of the label representations may be used, the tree is created by dividing into $k$ clusters, and the balancing constraint is not enforced.

\paragraph{}
Extreme Regression (XReg) \citep{xreg_prabhu} is proposed as a new learning paradigm for accurately predicting the numerical degree of relevance of an extremely large number of labels to a data point. For example, predicting the search query click probability for an ad. The paper introduces new regression metrics for XReg and a new label-wise prediction algorithm useful for Dynamic Search Advertising (DSA). 

The extreme classification algorithms treat the labels as being fully relevant or fully irrelevant to a data point but in XReg, the degree of relevance is predicted. Also, the XMLC algorithms performs point-wise inference i.e. for a given test point, recommend the most relevant labels. In this paper, an algorithm is designed for label-wise inference, i.e., for a given label, predict the most relevant test points. This improves the query coverage in applications such as Dynamic Search Advertising.

\paragraph{}
XReg is a regression method which takes a probabilistic approach to estimating the label relevance weights. All the relevant weights are normalized to lie between 0 and 1 by dividing by its maximum value, so as to treat them as probability values. These relevant weights are treated as marginal probabilities of relevance of each label to a data point. Thus, XReg attempts to minimize the KL Divergence between the true and predicted marginal probabilities for each label w.r.t each data point. This is expensive using the naive $1$-vs-All approach. So, XReg uses the trained label tree from Parabel.

\paragraph{}
Each internal node contains two $1$-vs-All regressors which give the probabilities that a data point traverses to each of its children. Each leaf contains $M$ $1$-vs-All regressors which gives the conditional probability of each label being relevant given the data point reaches its leaf.

\paragraph{}
For point-wise prediction, beam search is used as in Parabel. The top ranked relevant labels are recommended based on a greedy $BFS$ traversal strategy. For label-wise prediction, estimate what fraction of training points visit the node of a tree, say $f.$ For some factor $F$, allot $f \times F$ test points to respective nodes. This ensures all the non-zero relevance points for a label end up reaching the labels' leaf node. Finally, the topmost scoring test points that reach a leaf are returned.
\begin{table}[t!]
\footnotesize
\centering
\begin{tabular}{ |p{2cm}||p{1.5cm}|p{1.5cm}|p{1.5cm}|p{3cm}|p{3cm}|}
 \hline
 \multicolumn{6}{|c|}{Tree Based Methods} \\
 \hline
  Algorithm & Type of Tree & No of children & Explicit Balancing & Partitioning Method & Classification Method in Leaf \\
 \hline
  FastXML & Instance Tree / Sample Tree & Binary Tree ($2$ children) & No Balancing & nDCG based loss optimization problem solved at each internal node to learn partitioning and label ranking at the same time & Ranking function learnt at the leaves used for direct label prediction \\ 
 \hline
  Parabel & Label Tree & Binary Tree ($2$ children) & Balancing done by partitioning into same sized clusters & $2$-means based partitioning into balanced clusters & Labels with highest probability predicted by a MAP estimate learnt during training \\ 
 \hline
  CRAFTML & Instance Tree / Sample Tree & $k$-ary Tree ($k$ children) & No Balancing & $k$-means clustering on projected label and feature vectors for each sample & $k$-means classifier and label averaging for prediction \\ 
 \hline
  Bonsai & Label Tree & $k$-ary Tree ($k$ children) & No Balancing & $k$-means based partitioning into clusters & Labels with highest probability predicted by a MAP estimate learnt during training \\ 
 \hline
  XReg & Label Tree & Binary Tree ($2$ children) & Balancing done by partitioning into same sized clusters & $2$-means based partitioning into balanced clusters & Labels with highest probability predicted by a set of regressor on the path to the leaf node is given as final output.\\ 
 \hline
\end{tabular}
\caption{Comparison of Tree Based Methods for XMLC\label{table:compare-tree-based}}
\end{table}
\normalsize

\paragraph{Comparison between above stated methods}
In Table \ref{table:compare-tree-based} above, we compare tree based methods for XMLC. We highlight some key differences below. 

\begin{enumerate}
    \item Most tree based methods do not outperform the baseline one-vs-all scores, however, they have lesser prediction and training times compared to some compressed sensing-based, linear-algebra-based and deep-learning-based methods. For example, CRAFTML is $20\times$ faster at training than SLEEC.
    \item Label partitioning based methods perform slightly better than the instance partitioning methods in experimentation. Bonsai outperforms CRAFTML in most datasets. 
    \item Label partitioning is a more scalable approach due to the number of instances possibly being arbitrarily large. Instance partitioning is however, faster during prediction due to no need for traversal of multiple paths in the tree.
\end{enumerate}

\subsection{One-versus-All Methods}
To make one-versus-all methods scalable fore extreme classification problems, there were two prominant methods, namely, DISMEC~\citep{babbar2017dismec} and SLICE~\citep{jain2019slice}. We discuss these key two papers. 

DISMEC employs a double layer of parallelization, and by exploiting as many cores as available, it can gain significant training speedup over SLEEC and other SOTA methods. DISMEC achieves 10\% improvement in precision and nDCG measures over SLEEC. By explicitly inducing sparsity via pruning of spurious weights, models learnt by DISMEC can have up to three orders of magnitude smaller size.

DISMEC considers the following optimization problem: Squared Hinge-loss and L2-Regularization
\begin{equation}
    \min_{w_{l}} \quad \left[\|w\|^{2}_{2}  + C \sum_{i=1}^{N}(\max(0, 1-s_{l_{i}} w^{T}_{l} x_{i} ))^{2} \right],
\end{equation}
where $C$ is the parameter to control the trade-off between the empirical error and model complexity.
For large scale problems, the above optimization problem is solved in the primal using the trust region Newton method, which requires the gradient and Hessian. To tackle the issue of learning million of weight vectors $w_{l},$ they consider the following.

\begin{itemize}
    \item \textbf{Double layer of parallelization} this enables learning thousands of $w_{l}$ in parallel and hence obtain significant speed up in training for such scenarios.

    \item \textbf{Model sparsity by restricting ambiguity} to control the growth in model sizes without sacrificing prediction accuracy.
\end{itemize}

\paragraph{The highlights of double layer of parallelization is the following:} 
\begin{itemize}
    \item On top level, labels are separated into batches of say $1000$, which are then sent to separate nodes each of which consists of $32$ or $64$ cores depending on cluster hardware.
    
    \item On each node, parallel training of a batch of $1000$ labels is performed using OpenMP.
    \begin{itemize}
             \item For a total of \textbf{L} labels, it leads to \textbf{ $B = \frac{L}{1000} + 1 $} batches of label.
             \item For instance, if one has access to $M$ nodes with $32$ cores, then $32 \times M$ classifier weights can be trained in parallel.
    \end{itemize}
    \item In addition to scalable and parallel training, learning models in batches of of labels has another advantage, the resulting weight matrices are stored as individual blocks, each consisting of weight vectors for the number of labels in the batch. These block matrices can be exploited for distributed prediction to achieve real-time prediction which is close to the performance of the tree based classifiers such as FastXML.
\end{itemize}

The models may still aggregate up to 1TB due to large datasets. This is due to the use of L2-Regularization which squares the weight values \& encourages the presence of small weight values in the neighborhood of $0$.
In fact, approximately 96\% of the weights lie in the interval $[-0.01,0.01]$. Let $w_{\Delta} = \{ w_{d,l}, \|w_{d,l}\| < \Delta, 1 \leq d \leq D, 1 \leq \ell < L $. For small values of $\Delta$, the weight values in $w_{\Delta}$ represent the ambiguous weights which carry very little discriminative information of distinguishing one label against another. These ambiguous weights occupy enormous disk-space rendering the learnt model useless for prediction purposes.
In principle, $\Delta$ can be thought of as an ambiguity control hyperparameter  of DISMEC algorithm. It controls the trade-off between the model size and prediction speed on one hand, and reproducing the exact L2-Regularization model on the other hand. However, in practice, we fixed $\Delta = 0.01$, and it was observed to yield good performance.

It is also important to note that sparse models can also be achieved by using L1-Regularization but leads to under-fitting and worse prediction performance compared to L2-Regularization followed by the pruning step.

Leading extreme classifiers like DISMEC\cite{}, PDSparse and PPDSparse\cite{} scale to about a million labels. Their training and prediction costs grow linear with the number of labels making them prohibitively expensive when the number of labels go up to 100 million labels. In the papers authors where embedding each data point into a low (64) dimensional space and they observe that Parabel do not give a good accuracy for low dimensional embedding, also parable do-not scale beyond 10 million labels.
    
In \cite{jain2019slice}, authors proposed a new method called SLICE, where they use 1-vs-All approach. In usual 1-vs-All classifiers, classifier for each label will be trained on the positive training samples and associated negative samples for that label. Usually the negative samples will be in large numbers since there are very large number of labels. Authors in this paper propose a novel negative sampling technique which is demonstrated to be significantly more accurate for  low dimensional dense feature representations , using which they train each classifier only the most confusing negative samples. Thus reducing the complexity.

In the SLICE paper authors use a 1-vs-All model to solve the extreme classification model. Let $ \{ (x_i,y_i)_{i=1}^N \} $ be the set of N independent and identically distributed training points with unit normalized features i.e $ ||x_i||_2^2 = 1 $, lying on the $D$ dimensional unit hyper sphere $ x_i \in S^D $, $ y_i \in \{ 0, 1 \}^L $. 

In the 1-vs-All model, one can model the posterior probability directly by using 
\begin{equation}
    P(y_i|x_i;\textbf{W}^D , \, \textbf{b}^D) = \Pi_{l=1}{L} (y_il|xi;w_l^D,b_l^D) = \Pi_{l=1}^{L} \left (
    1+e^{-y_{il}({w_l^D}^T x_i+b_l^D)} \right)^{-1}
\end{equation}
To maximize the this posterior probability we can take logarithm and add regularization, we get 
\begin{equation}
    \underset{w_l,b_l}{\text{argmin}} \quad \frac{1}{2}w_l^{T}w_l + C\sum_{i=1}^{N} \text{log} \left (
    1+e^{-y_{il}({w_l^D}^T x_i+b_l^D)} \right)^{-1}
\end{equation}

By differentiating this equation with respect to $ w_l $ and equating the resulting term to zero, we get the $w_{\ell}$ which minimizes the above expression
\begin{equation}
    w_{\ell}^D = C \sum_{i=1}^{N} P \left ( -y_{i\ell} \, \mid \, x_i; w_{\ell}^D , b_{\ell}^D \right )
    \label{minimum_Weight_Equation_Slice}
\end{equation}
Authors say that the training points which contribute very less to this sum, i.e., the training points $ x_i $ for which the value $ \left (1+e^{-y_{i\ell}({w_{\ell}^D}^T x_i+b_{\ell}^D)} \right)$ is small. Searching for the points $ x_i $ which contribute very less will need computing the value for all the points. This approach is not useful since the computation for all the points is costly and we want to compute only less points to get a very good estimate of minimum $ w_{\ell}^D.$ 

To cheaply compute the negative examples for each classifier, authors in this paper, model the likelihood and prior probabilities, such that the resulting posterior $ P \left (y_{i\ell}|x_i;w_{\ell}^G , b_{\ell}^G \right ) $ will be similar to the one that is learned directly from the data $ P \left (y_{il}|x_i;w_{\ell}^D , b_{\ell}^D \right ) $. Here $D$ represents that posterior model is directly learned from data, where as $G$ represent that posterior is obtained using a generative model.

Using the generative model, one can see which examples will give the highest value for $ P \left (y_{i \ell}|x_i;w_{\ell}^G , b_{\ell}^G \right ) $ and take that samples as negative samples (if cheaply available), then along with the positive examples for that label train the classifier.

Lets look at generative model described in the paper. Each label is assumed to be independently activated $ (y_{i\ell} = +1) $or deactivated $ (y_{i\ell} = -1) $ with the probabilities $ \pi_{\ell} $ and $ 1 - \pi_{\ell} $ respectively.
So the probability of vector $ y_i $ of $L$ labels is written as $ p(y_i;\pi)$
\begin{equation}
    p(y_i;\pi) =  \Pi_{i=1}^{L} \quad \pi_{l}^{\frac{1}{2}(1+y_il) } (1-\pi_{l})^{\frac{1}{2}(1-y_il) }
\end{equation}
feature vector $ x_i $ is assumed to be sampled from the distribution 
\begin{equation}
\begin{split}
    P(x_i \, | \, y_i) = 
        \begin{cases} 
            e^{\frac{1}{2}\sum_{\ell=1}^{L} (1+y_{i\ell}) \left ( \gamma_{\ell}^+ x_{\ell}^T \mu_{\ell}^+ + Z_{\ell}^+ \right ) + ( 1-y_{i\ell} ) \left ( \gamma_{\ell}^- x_{\ell}^T \mu_{\ell}^- + Z_{\ell}^- \right )    } 
            & \text{if} \quad \textbf{x} \in S^D  \\
            0 & \text{if} \quad \textbf{x} \notin S^D.  
        \end{cases}
\end{split}
\end{equation}

Where parameters for each label $\ell$ are $ \gamma_{\ell}^+ , \gamma_{\ell}^-, \mu_{\ell}^+ , \mu_{\ell}^- $. Seeing this distribution one can see that
\begin{equation}
\begin{split}
    P(x_i|y_{i\ell}) = 
        \begin{cases} 
            e^{ \left ( \gamma_{\ell}^+ x_{\ell}^T \mu_{\ell}^+ + Z_{\ell}^+ \right )} 
            & if \quad y_{i\ell} = +1 , x_i \in S^D \\
            e^{ \left ( \gamma_{\ell}^- x_{\ell}^T \mu_{\ell}^- + Z_{\ell}^- \right )    } 
            & if \quad y_{i\ell} = -1  ,x_i \in S^D.
        \end{cases}
\end{split}
\end{equation}
The above distributions are Von Mises Fisher Distributions, In the paper authors are modeling probabilities of positive and negative points for each label with this distribution.

Computing the posterior $ P(y_i|x_i; W^G,b^G) $ using bayes theorem will give 
\begin{equation}
    \begin{split}
        P(y_i|x_i; W^G,b^G) =   \quad P(x_i|y_i)P(y_i;\pi)/P(x_i)
       =  \quad \prod_{l=1}^{L} \left (  1+e^{-y_{il}(w_l^G x_i + b_l^G)}   \right )^{-1},
    \end{split}
\end{equation}
where $w_\ell^G = \gamma_l^+ \mu_l^+ - \gamma_l^- \mu_l^-  $ and $ b_l^G = Z_l^+ - Z_l^- + \log(\frac{\pi_l}{1-\pi_l}) $. It is said in the paper that when training data grows asymptotically, $ w_l^G, b_l^G $ will converge to $ w_l^D,b_l^D $ respectively and hence sampling the negative training points from generative model posterior will be as effective as sampling from discriminative posterior model.

It is stated in the paper that the maximum likelihood estimates of concentration parameters $ \gamma_l^{\pm} $ cannot be estimated efficiently. Hence other parameters ( $ w_l^D,b_l^D $ ) which are dependent on $ \gamma_l^{\pm} $ cannot be calculated efficiently. But they stressed on observation that concentration parameter for the positive samples will generally be higher than that of negative samples, since negative samples are from all possible labels. Using this observation, it is assumed that $ \gamma_l^- \mu_l^- $ is negligible when compared to $ \gamma_l^+ \mu_l^+  $
and $ w_l^G = \gamma_l^{+} \mu_l^{+}  $, this leads to 
\begin{equation}
    \begin{split}
        P(y_i|x_i; W^G,b^G) = &  \quad P(x_i|y_i)P(y_i;\pi)/P(x_i)\\
       = & \quad \prod_{l=1}^{L} \left (  1+e^{-y_{il}(w_l^G x_i + b_l^G)}   \right )^{-1} 
       =  \quad \prod_{l=1}^{L} \left (  1+e^{-y_{il}(\gamma_l^{+} \mu_l^{+} x_i + b_l^G)}   \right )^{-1} \\
       P(-y_i|x_i; W^G,b^G) = &  \quad \prod_{l=1}^{L} \left (  1+e^{y_{il}(\gamma_l^{+} \mu_l^{+} x_i + b_l^G)}   \right )^{-1} \\
       P(-y_{il}|x_i; W^G,b^G) = &  \quad  \left (  1+e^{y_{il}(\gamma_l^{+} \mu_l^{+} x_i + b_l^G)}   \right )^{-1} 
       =   \quad  \left (  1+e^{-1(\gamma_l^{+} \mu_l^{+} x_i + b_l^G)}   \right )^{-1} \quad \text{if} \quad y_{il} = -1.
    \end{split}
\end{equation}

\paragraph{}
From the above equation we can see that the term $P(-y_i|x_i; W^D,b^D) = P(-y_i|x_i; W^G,b^G)$ in equation \eqref{minimum_Weight_Equation_Slice} is equal to $\left(1+e^{-1(\gamma_l^{+} \mu_l^{+} x_i + b_l^G)} \right)^{-1}$ for negative samples $x_i ~ (\text{which have}~ y_{il} = -1 $). Hence negative samples which contribute very large $P(-y_i \mid x_i; W^G,b^G) $ are those with high values of $ \mu_l^{+} x_i.$ As both $ \mu_l^{+} ,x_i $ are normalized, the higher the dot-product the lesser the distance between those. The insight from this is that the most confusing negative samples for a given label $l$ are those $ x_i $ ( $\quad \text{with} \quad y_{il} = -1$) which are most nearer to $\mu_l^+ .$ 

\paragraph{}
In this paper authors use an algorithm called HNSW ( Hierarchical Navigable Small World Models) to construct Approximate nearest neighbor search(ANNS) data structure, which can be used to query the nearest neighbors efficiently (in almost log time). In this paper, ANNS data structure is built using $ \mu_l^+ \left( =  \frac{\sum_{i:y_{il}=1} x_i}{||\sum_{i:y_{il}=1} x_i ||_2} \right ) $, and then queried for all the training points $ x_i $ to get the set of labels $ S(x_i) = \{  l| 1\leq \ell \leq L, \mu_l^+ \in ANNS(x_i)   \} $ having the highest values of dot product  $ \mu_l^+ x_i $. Then for each label l, the set of negative training samples $ N_l^{\mu} $ can be selected such that $ N_l^{\mu} = \{  i|1\leq i \leq N, y_{il} = -1, \, l \in S(x_i) \}.$

\paragraph{}
Similar to computing the label representations in Parabel, here in slice, the $ \mu_l^+ $ vectors for all labels are computed with $ O(ND\log(L)) $ cost. Average number of Positive samples for a label are $ O(\frac{N}{L}\log L ) $. Same number of most confusing negative samples are considered for training the classifier for that label. i.e  $ |P_l | = | N_l^{\mu} | = O(\frac{N}{L}\log L ).$ To construct HNSW data structure for negative samples cost is $ O( LDlogL  ).$ To select $ N^{\mu} $ cost is $ O(ND\log L) .$ Cost to train $ O(\frac{N}{L}\log L ) $ examples is of $ O(\frac{N}{L}(logL) DL),$ i.e., $ O( ND\log L ).$ Thus the total cost is $ O(ND\log L) $, since $ N>L $ in this problem.

\subsection{Deep-Learning-Based Methods}

\paragraph{}
Deep learning as a field has grown rapidly in the recent past, and has started to become a dominating method in most machine learning problems, where there is a large amount of available data. In extreme multi label classification, deep learning methods for XMLC did not appear until later, for several reasons. Firstly, the extremely large output space would require correspondingly large neural network models (in terms of parameters) to output all labels. Secondly, the prevalence of tail labels (many labels with very few training examples) means deep models struggle to learn those rare labels. For example, in the Wiki-500K dataset \citep{Bhatia16}, 98\% of labels have less than 100 training instances. Early attempts at deep learning in XMLC were not very successful compared to linear algebra or tree-based methods that dominated the benchmarks at the time. However, deep learning proved extremely effective at extracting context-dependent features from text. With improved architectures, these methods eventually challenged the earlier bag-of-words approaches in XMLC. 

\paragraph{}

The paper \cite{liu2017deep} was the first to try out application of Deep Learning in XMLC, using a family of CNN's. This followed the success of deep learning in multi-class text classification problems by methods like FastText \citep{joulin2016bag}, the CNN text-classifier of Kim ({CNN-Kim}) \citep{chen2015convolutional} and a bag-of-words CNN ({Bow-CNN}) by \cite{johnson2014effective}. These methods inspired the XML-CNN architecture, which applies similar concepts to multi-label setting. 

\paragraph{}
In \cite{kim2014}, authors create a document embedding from concatenation of word embeddings. $t$ CNN filters are applied to obtain a $t$ dimensional vector which is passed to a soft-max layer after max pooling. Bow-CNN \citep{liu2017deep} creates a $V$ dimensional vector for each region of the text, where $V$ is the vocabulary size, indicating whether each word is present in the region. It also uses dynamic max pooling for better results. XML-CNN combines these ideas and proposes the following architecture - various CNN filters similar to CNN-Kim, dynamic max pooling similar to Bow-CNN, a bottleneck fully connected layer, and binary cross entropy loss over a sigmoid output layer. 

\paragraph{}
Let $e_i \in R^k$ be the word embedding of the $i-$th word of the document, then $e_{i:j} \in R^{k(j-i+1)}$ is the concatenation of embeddings for a region of the document. A convolution filter $v$ is applied to a region of $h$ words to obtain $c_i = g_c(v^Te_{i:i+h-i})$ which is a new feature, where $g_c$ is a non-linear activation. $t$ filters like this are used with varying $h$ to obtain a set of new features. 

\paragraph{}
Dynamic Max pooling is then used on these newly obtained features $c$. The usual max pooling basically takes the maximum over the entire feature vector to obtain a single value. However, this value does not sufficiently represent the entire document well. Thus, a max-over-time pooling function is used to aggregate the vector into a smaller vector by taking max over smaller segments of the initial vector. Thus, pooling function $P$ is given by 
\begin{align*}
P(c) = \left[\max \left(c_{1:\frac{m}{p}} \right), \, \dots , \, \max \left(c_{m-\frac{m}{p}+1:m} \right) \right] \in \R^p.
\end{align*}
This pooling function can accumulate more information from different sections of the document. These poolings from different filters are then concatenated into a new vector.

\begin{figure}[t!]
    \centering
    \includegraphics[scale=0.4]{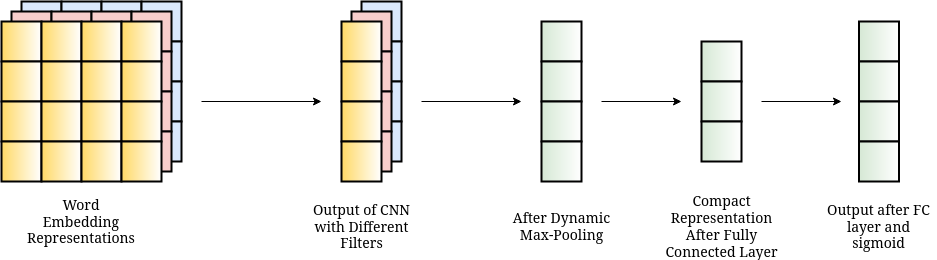}
    \caption{\label{fig:CnnXML} The XML-CNN model }
\end{figure}

\paragraph{}
In Figure \ref{fig:CnnXML}, we show an illustration of XML-CNN. The output of the pooling layer is now fed into a fully connected layer with fewer number of neurons, also called a bottleneck layer. This bottleneck layer has two advantages. Firstly, it reduces the number of parameters from $pt \times L$ to $h \times (pt + L)$ where $L$ is the number of labels, $t$ is the number of filters, $p$ is the pooling layer hyperparameter and $h$ is the number of hidden layers in the bottleneck layer. This allows the model to fit in memory as L is often large. Secondly, another non-linearity after the bottleneck layer leads to a better model. Thirdly, this encourages the learning of a more compact representation of the data.

\paragraph{}
Finally, the output of the bottleneck layer is passed through another fully connected layer with output size $L$. The binary cross entropy loss is observed to work the best used for this method. 

The ablation studies performed in the paper show that BCE loss, bottleneck layer, and dynamic max pooling each contribute to the improvement of the model. Though XML-CNN was slower than the tree based methods, it clearly outperformed them in terms of performance. It also performed better than linear algebra based methods, showing the potential deep learning held for the XMLC field.


\paragraph{}

However, XML-CNN was unable to capture the most relevant parts of the input text to each label, because the same text representation is given for all the labels. Sequence to Sequence based methods like MLC2Seq \citep{MLC2Seq} were also not suited for the task since the underlying assumption that labels are predicted sequentially is not reasonable in extreme classification. The paper \cite{you2019attentionxml} proposed \textbf{AttentionXML}, which used a BiLSTM (bidirectional long short-term memory) to capture long-distance dependency among words and a multi-label attention to capture the most relevant parts of texts for each label. Since application of attention to train on each label with the full dataset is not possible for large datasets, there must be a method for training on only the most relevant samples for each label. For this, AttentionXML uses a shallow and wide Probabilistic Label Tree (PLT) built on the labels.

\begin{figure}[t!]
    \centering
    \includegraphics[scale=0.3]{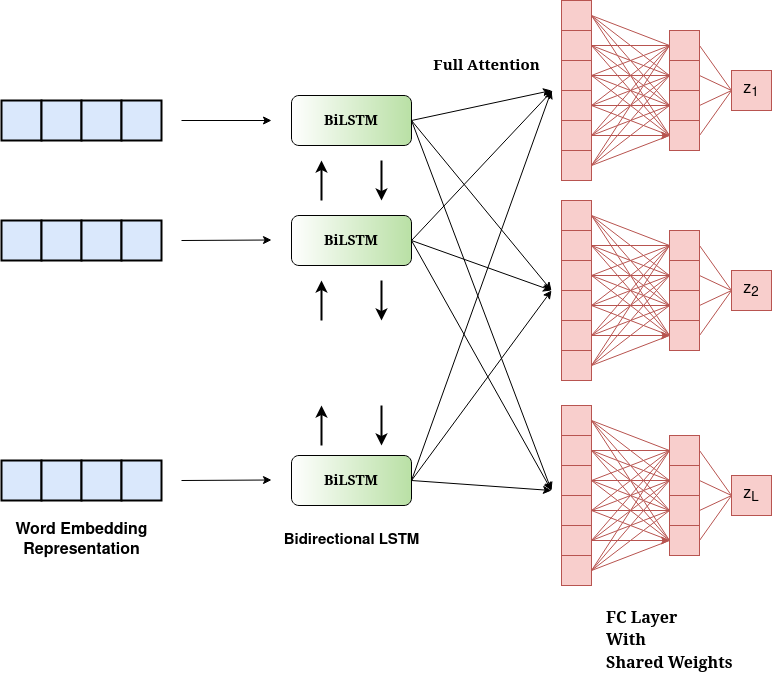}
    \caption{\label{fig:AttnXML} An Illustration of Attention Aware Deep Model in AttentionXML }
\end{figure}

\paragraph{}
The first step is to create the PLT, which is be done by hierarchical clustering algorithm followed by a compression algorithm to reduce depth. This step is similar to that of Parabel by \cite{prabhu2018parabel} where we create a label tree, but here the tree is constructed to be wide and shallow. In effect, the label tree resembles that of Bonsai by \cite{khandagale2020bonsai}. The reason for creation of a wide tree instead of a deep one is to prevent error propagation through the levels. The leaves correspond to an original label while the internal nodes correspond to meta or pseudo labels. Each node predicts $P(z_n | z_{Pa(n)}=1, x)$, ie the probability that label or pseudo-label $n$ is present, given that parent of $n$ is present. Thus, the marginal probability for each label being present is given by the product of probabilities in its path in the PLT i.e.,
\begin{align*}
    P(z_n =1 \mid x) = \Pi_{i \in \text{Path}(n)} P(z_i \mid z_{Pa(i)}=1, x).
\end{align*}
The value of the marginal probabilities is predicted by the attention based deep learning network. As depicted in \ref{fig:AttnXML}, the architecture consists of a bidirectional LSTM which finds embeddings for the words and then followed by a 2-layer fully connected neural network which finally predicts the probability that the current label is active given the input. There is one block of such network per layer of the PLT. 

\paragraph{}
The input to the fully connected layers is given by using a attention based mechanism which can find if a word embedding is relevant to the label. Finally, the parameters of the fully connected layers are shared among the different labels in each layer, as they perform the same function. This also largely helps reduce model size. Further, to reduce training cost, the model weights of each layer is initialized with the weights of the previous layer.

\paragraph{}
It is to be noted that even if this network is directly used, the training time will be infeasible as each label would have to train on each sample. However, we do not need to train on every sample for a particular label. Hence, only the samples which contain the label and a few more negative samples are used for training.

\paragraph{}
AttentionXML outperforms XML-CNN and most other methods in terms of P@1 scores. Further, the good PSP scores show that AttentionXML performs well on tail labels as well. The utility of multi-head attention is proven in the ablation study. However, AttentionXML is not easy to train as the training and prediction times are much larger compared to other methods and hence does not scale well to larger datasets.

\paragraph{}

Despite the usage of attention by AttentionXML, a direct attempt at using the pretrained transformers for XMLC was not yet made. The transfer learning of pretrained transformers for this task could be the solution to the large training times required by AttentionXML. The paper by \cite{chang2020taming} made the first attempt at using pretrained transformers for the XMLC task in the model {\em XTransformer}. Transformers like BERT \citep{devlin2018bert} RoBERTa \citep{liu2019roberta}, and XLNet \citep{yang2019xlnet} are highly successful in the NLP domain where they outperformed most other methods on tasks like question answering, POS tagging, and sentence classification due to their ability to generate contextual embeddings for words and sentences which are applicable to downstream tasks. However, application of these models for the XMLC problem was not so straightforward even though XMLC is primarily a text classification problem. 

\paragraph{}
The difficulty was because of the large number of labels and the output space sparsity. Many of the labels do not have enough samples to be able to be able to train large deep neural networks and hence direct application of pretrained transformers led to bad results and memory issues. Xtransformer handles this by reducing the larger problem into smaller subproblems, like in tree based methods, and then training the transformers to predict in this smaller problem. Each of the subproblems are eventually solved to get the final predictions.

\begin{figure}[t!]
    \centering
    \includegraphics[scale=0.32]{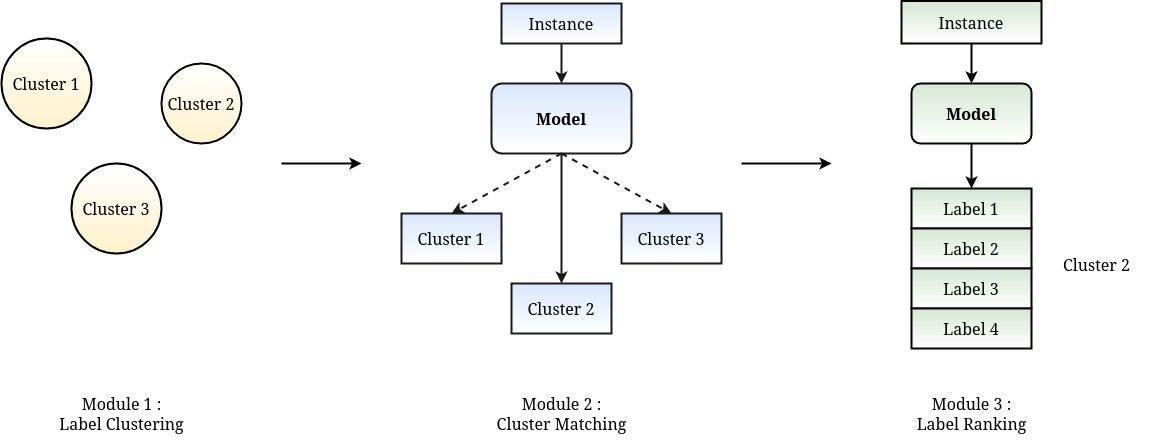}
    \caption{\label{fig:Xtransformer} The XTransfomer method }
\end{figure}

\paragraph{}
In Figure~\ref{fig:Xtransformer}, we show an illustration of Xtransformer. More concretely, the XTransformer model consists of three parts or modules. The first module is called Semantic Label Indexing, which is responsible for the division of the larger problem into smaller subproblems. For this, a clustering of the labels is performed, creating $K$ clusters where $K \ll L$. The clusters are created by finding a label to cluster assignment matrix $C \in \{0,1\}^{L\times K}$ where $C_{lk} = 1$ if label $l$ belongs to cluster $k$. This matrix can be created by using hierarchical clustering algorithms and the clusters are represented based on the mean of transformer embeddings of the label text for the labels present in the cluster. 

\paragraph{}
These clusters are then used as meta-labels for the problem of fine tuning the transformer models in the second module called the Neural Matcher. Since the number of meta-labels is much smaller, the  label sparsity issue due to tail labels is avoided. The transformers are trained to map from instances to the relevant clusters. Embeddings $\phi(x)$ are thus trained for each instance $x$. 

\paragraph{}
In the third module after the matching step, a small subset of label clusters is selected for each instance from which we are required to find a relevance score of each label. For performing this final step which is crucial to prediction, a one-vs-all classifier is trained for each label. The instances on which each label trains must be limited in number to allow efficient training. Techniques like picking teacher-forcing negatives (for each label $l$, picking only the instances instances with label $l$ or with any of the labels in the same cluster as label $l$) allow picking only a small subset of ``hard" instances.

\paragraph{}
During prediction, a subset of clusters are selected by the Matcher. All the labels selected from each of the clusters are then passed the test instance to which each one-vs-all classifier gives a relevance score to decide if the label is relevant to the instance, based on which the label ranking is done. XTransformer is successfully able to perform well on all datasets and was only outperformed by AttentionXML on some datasets at the time.

\paragraph{}
Some other successful methods which attempt to fine tune transformers have been recently proposed such as APLC-XLNet \citep{ye2020pretrained} which uses transfer learning of the popular transformer model XLNet \citep{yang2019xlnet}. As with previous methods, performing this directly is not possible, hence this methods uses Adaptive Probabilistic Label Clusters, which attempts a clustering based on separating the head and tail clusters, and then assigning head cluster as the root of a tree and tail clusters as leaves. 

\paragraph{}
Another method which uses transformers to perform extreme classification is LightXML \citep{jiang2021lightxml}. This paper points out that methods like AttentionXML and XTransformer suffer from static negative sampling, which highly reduces the ability of the model. The main advantage of LightXML over other methods using static negative sampling is that the model does not overfit to specific negative samples. LightXML performs dynamic negative sampling by using a generator-discriminator like method in which the label recalling is performed by a module (generator) and the label ranking is done by another module. This procedure allows each module to try and improve its own objective while enhancing the other, thus leading to better performance overall. 

\paragraph{}
The transformer-based approaches like X-Transformer \citep{chang2020taming} above and LightXML \citep{jiang2021lightxml} achieve state-of-the-art XMC results by clustering labels to reduce computational complexity. However, they still face significant memory and training time issues with larger label spaces.
\paragraph{}
XR-Transformer by \cite{zhang2021fast}, addresses these computational challenges by leveraging recursive shortlisting and hierarchical label trees (HLTs) \citep{prabhu2018parabel,matchxml}. XR-Transformer recursively fine-tunes pre-trained transformers, on progressively smaller label spaces, ultimately reducing the computational burden. This recursive process ensures that for any input, the number of candidate labels considered during training and inference is \(O(B)\), and the total number of labels is \(O(B \log_{B}(L))\), where \(B\) is the cluster size and \(L\) is the labels.
The key steps in XR-Transformer are as follows:
\begin{enumerate}
\item Labels are recursively clustered using balanced $k$-means, forming a hierarchical label tree (HLT). Label features \( Z \in \mathbb{R}^{L \times \hat{d}} \) are constructed using text vectorizers or Positive Instance Feature Aggregation (PIFA),
    $$
Z_{\ell}=\frac{{v}_{\ell}}{\left\|{v}_{\ell}\right\|} ; \quad \text { where } {v}_{\ell}=\sum_{i: y_{i, \ell}=1} \Phi\left({x}_i\right), \: \forall \ell \in[L]
$$ 
where \( \Phi : \mathcal{D} \to \mathbb{R}^d \) is the text vectorizer.

\item Coarse label vectors are obtained through max-pooling, creating a series of coarse-to-fine learning signals. To mitigate information loss when merging positive labels, XR-Transformer uses recursively constructed relevance matrices, which assign non-negative importance weights to each instance and label cluster.
\item Instead of training on the entire label space, it shortlists labels based on the relevance scores and the top-$k$ relevant clusters from the parent layer. This shortlisting approach, combined with multi-resolution learning, defines a series of learning objectives that progressively refine the model's ability to predict relevant labels efficiently.
\end{enumerate}

\paragraph{}
Another notable feature of XR-Transformer is its combination of statistical text features (such as TFIDF) and semantic transformer embeddings. This dual representation mitigates the information loss associated with text truncation in transformers and leverages the strengths of both statistical and deep learning approaches.
Empirical results show XR-Transformer significantly improves training efficiency and performance on six public XMC benchmarks datasets(Eurlex-4K, Wiki10-31K,
AmazonCat-13K, Wiki-500K, Amazon-670K, Amazon-3M). On the Amazon-3M dataset, it increased Precision@1 from 51.20\% to 54.04\% and reduced training time from 23 days to 29 hours, highlighting its scalability and efficiency without sacrificing accuracy.

\paragraph{}
\cite{chien2023pina} push label-aware extreme classification a step further with PINA (Predicted-Instance Neighbourhood Aggregation), a two-stage plug-in that wraps around any base XMLC model (they use XR-Transformer~\cite{zhang2021fast}) and injects graph-style context.  Earlier label-aware systems—ECLARE’s label-text features \citep{mittal2021eclare}, GalaXC’s joint document–label GNN \citep{saini2021galaxc}, NGAME’s batch-level negative mining \citep{dahiya2023ngame} either enrich the label side or mine harder negatives; PINA~\citep{chien2023pina} instead enriches the \emph{instance} side by treating XMLC as a neighbourhood-prediction task akin to NODEPRED~\citep{chien2022node}.  During pre-training PINA builds a biadjacency graph whose vertices are both documents and labels, then learns a bi-encoder (instance-text and label-text) with a graph-contrastive loss, directly leveraging the label descriptions that AttentionXML~\citep{you2019attentionxml} and XR-Transformer originally ignored.  In the subsequent augmentation stage the trained neighbour predictor retrieves the top-\(K\) nearest instances for every sample and aggregates their embeddings—essentially a graph-convolution step \citep{hamilton2018inductive} before the underlying classifier sees the features.  Because the neighbour predictor is frozen, the extra computation scales only with \(K\), so PINA preserves the logarithmic complexity of XR-Transformer.  On LF-AmazonTitles-1.3M this neighbourhood injection lifts Precision@1 by roughly 5 pp over the vanilla XR-Transformer, showing that side-information-driven feature aggregation can rival the label-aware gains of ECLARE while remaining model-agnostic and inexpensive at prediction time.

\paragraph{}

A general trend can be observed from papers like AttentionXML and XTransformer, where deep learning methods are not directly applied to the problem due to scalability issues. Instead, the deep learning model is applied to a smaller subproblem to obtain intermediate features which can be further used in the problem. Also, negative sampling is generally used for training the final model as training on all samples is not feasible. \cite{dahiya2021deepxml} comes up with a general framework for applying deep learning methods to XML called DeepXML. Consequently, they propose a new method called ASTEC using the existing method which achieves a new state of the art for some datasets in the domain. 

\paragraph{}
The paper defines DeepXML to be a framework with four modules. Each module is designed to be fairly of low complexity, allowing the entire method to scale to large datasets easily. Not all the methods need to have each of the modules depending on the utility, and the module algorithm can be switched to another algorithm easily without affecting the entire method.

\paragraph{}
The first module of the framework is for training intermediate representations for the features $Z^0$. Here, $Z^0$ should be as close as possible to the final representation of the features $Z$ which would be the feature representations if the entire problem was optimized directly.  Since features cannot be trained on the entire problem, a surrogate label selection procedure is followed to ensure that the problem is tractable. Some examples of surrogate label selection procedures could be label subset selection, label clustering or or label projection into a smaller space.

\paragraph{}
The second module of DeepXML is the label shortlisting or the negative sampling step, where for each sample, we select the most confusing of the ``hardest'' labels. We select $N_i \subseteq [L]$ for each sample $i$ such that $|N_i| = O(\log L)$ and these labels are most likely to be predicted as positive labels for the samples. This negative sampling is done based on the intermediate features learnt in the first module and any method can be used for this procedure such as graphs, trees, clustering or hashing.

\paragraph{}
The third module is based on transfer learning to obtain the final feature representations from the initial feature representations learnt in the first module. The parameters can be re-trained from the first module, but it must be ensured that the final feature representation $Z$ should not be too far from the initial representation.

\paragraph{}
In the fourth module, the final classifier is learnt. This classifier uses the feature representation from the third module and the negative samples from the second module. Since the number of instances for each label is not excessive due to the negative sampling, the model training time is limited. The third and fourth modules can be combined as required.

\begin{figure}[t!]
    \centering
    \includegraphics[scale=0.35]{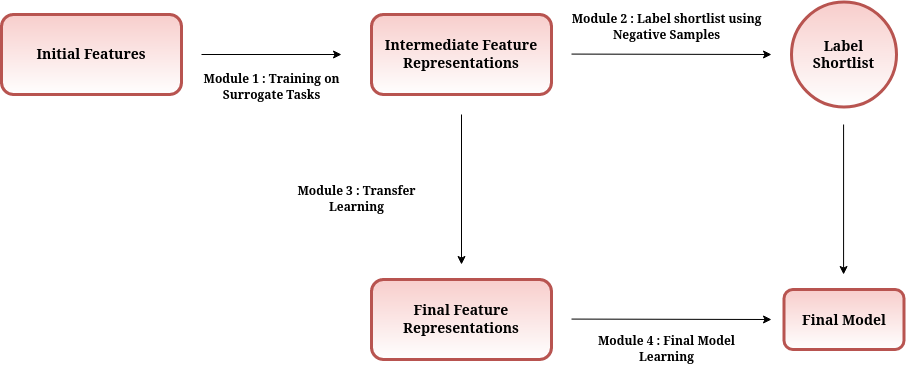}
    \caption{\label{fig:DeepXML} Overview of the DeepXML Framework}
\end{figure}

\paragraph{}
In the paper \cite{dahiya2021deepxml}, the method ASTEC is proposed which uses the DeepXML framework. For the first module, it uses a tree based clustering method to obtain the surrogate labels for training intermediate features. For the second module, an Approximate Nearest Neighbor Search (ANNS) data structure is used for shortlisting. Modules 3 and 4 are trained together where the transfer learning is done by a simple transfer matrix, while the final classifiers are simple one-vs-all classifiers. 

\paragraph{}
More methods using this framework have also been proposed and several existing ones, such as XTransformer can can be formulated to fit into this framework. \cite{mittal2021decaf} proposes a method DECAF using this framework which incorporates the label metadata. In another paper, \cite{mittal2021eclare} proposes ECLARE which utilizes label to label correlation as well as label text to give a new method following the DeepXML framework. GalaXC \citep{saini2021galaxc} is another method in this framework which uses graph neural networks on the combined document label space.

\paragraph{}
Recent advances have been made in the field by using the label features in addition to the instance features. Since in many of the XMLC datasets, the labels are also entities with some text information. This allows using the label text for learning better models. Different methods try to utilize the label features in different ways. For example, DECAF \citep{mittal2021decaf} directly incorporates label text into the classification procedure by embedding the instances and labels separately and then using a ranker to predict which label embedding is most relevant to the instance.

\paragraph{}
On the other hand, GalaXC \citep{saini2021galaxc} utilizes graph neural networks to learn embeddings jointly over documents and labels. The method uses convolutions of varying size to develop an attention mechanism for learning several representations for each node, where each document and label is a node.
Multiple hops over the joint label and document graph allow collection of information over related labels, other documents with these labels and so on. This is used for shortlisting and then prediction is finally made by using the label attention mechanism. 
ECLARE \citep{mittal2021eclare} also uses a similar method by accommodating for label metadata and label correlations. 

\paragraph{}
SiameseXML \citep{dahiya2021siamesexml} also uses label features, but it aims to learn joint embeddings for the labels and features by using the concept of Siamese Networks \citep{chen2020simple}, \citep{schroff2015facenet} where a triplet or contrastive loss is used with a hard negative mining procedure. The main goal is to learn common embeddings for instances and labels where instances are close to the respective positive labels and far from the other negative labels. This concept is adapted to the extreme classification scenario by using the DeepXML framework.

\paragraph{}
The paper \cite{dahiya2022ngame} suggests another method NGAME which uses Siamese Networks for extreme classification. A common method of using Siamese Networks is using minibatches. The set of candidate labels contains all the labels which appear in at least one of the other instances in the minibatch. The minibatch size, if constructed independently, cannot be too large as the set of labels would be to big. This leads to slower training. NGAME strategically constructs the mini-batches such that the samples are close, thus leading to informative negative samples as compared to random samples. This leads to better convergence and faster training.

\paragraph{}

The recent \textit{MatchXML} framework proposed in~\cite{matchxml} re-casts extreme multi-label classification as an explicit \emph{text–label matching} task. Earlier deep-XMC systems such as XR-Transformer \citep{zhang2021fast}, LightXML~\citep{LightXML} and X-Transformer~\citep{chang2019modular} embed labels with either sparse TF–IDF vectors or PIFA embeddings~\citep{Yu2019}; both inherit TF–IDF’s chief weakness—word-order is ignored, so fine-grained semantics are lost. MatchXML mitigates this by fusing those sparse features with dense semantic vectors.

\paragraph{}
First, MatchXML introduces {\em label2vec}, a Transformer-based encoder that produces compact, semantically rich label embeddings; these are organised into a hierarchical label tree (HLT) following~\cite{prabhu2018parabel}. Dense label2vec codes are markedly smaller than TF–IDF and capture meaning more faithfully, especially in large corpora. Second, training is formulated as a \textit{pairwise} problem: each mini-batch contains a document paired with a small set of positive and sampled negative labels. The encoder is fine-tuned with a contrastive loss that pushes true text–label pairs closer and unrelated pairs apart, thereby learning discriminative representations without scoring the entire label set.

\paragraph{}
MatchXML further enriches each document with fixed sentence-level embeddings from a pre-trained Sentence-Transformer. These static vectors, together with sparse TF–IDF and the dynamically fine-tuned Transformer features, give the model both lexical precision and contextual depth—beneficial for longer inputs and tail-label prediction.

\paragraph{}
Ablation studies confirm that (i) label2vec alone outperforms TF–IDF on large benchmarks, (ii) the mini-batch matching objective yields higher precision than flat softmax baselines, and (iii) adding static sentence embeddings provides an additional boost. Across six public datasets MatchXML delivers both higher accuracy and faster training than prior deep-learning XMC systems.

\paragraph{}
The majority of deep-learning XMC models—such as LightXML~\citep{LightXML} and XR-Transformer~\citep{zhang2021fast} implicitly optimise for \emph{head} labels because these dominate the training signal.  On the other hand, {\em BoostXML} tackles the opposite problem: it explicitly elevates \emph{tail-label} accuracy by embedding a gradient-boosting mechanism into a standard deep-XMC backbone.

\paragraph{}
Following the decoupling idea of~\cite{Kang2020Decoupling}, BoostXML splits learning into
(i)~a \emph{representation stage} that trains a BiLSTM–attention encoder end-to-end (identical to AttentionXML’s backbone) under a conventional binary-cross-entropy (BCE) loss, and  
(ii)~a \emph{classifier-adaptation stage} in which the encoder is frozen and the remaining layers are re-optimised via gradient boosting.

\paragraph{}
In the boosting step, at round~\(t\) the predictor is updated as
\[
F_t(x)=F_{t-1}(x)+\rho_t\,f_t(x),
\]
where \(f_t\) is a shallow shared MLP trained \emph{from scratch} on the residuals of all tail labels and \(\rho_t\) is a learned step size.  
Each weak learner uses only a few epochs and random initialisation (to keep it ``weak'') and is fitted with mean-squared error on residuals; both first- and second-order gradients steer the learner to labels currently mis-classified, thereby concentrating capacity on the long tail.

\paragraph{}
After adding \(f_t\), BoostXML performs a short corrective fine-tune of \emph{all} parameters—including the frozen encoder and earlier learners—under a BCE objective.  
This global adjustment mitigates drift, helps escape local minima, and dynamically regularises the effective boosting rate.

\paragraph{}
Evaluated on \textsc{Eur-Lex}, \textsc{Wiki10-31K}, \textsc{AmazonCat-13K}, \textsc{Wiki-500K}, and \textsc{Amazon-3M}, BoostXML surpasses state-of-the-art deep baselines (AttentionXML~\citep{AttentionXML}, LightXML~\citep{LightXML}) \emph{and} strong non-deep methods (AnnexML~\citep{AnnexML}, FastXML~\citep{FastXML}, PfastreXML~\citep{PfastreXML}).  
Gains are most pronounced on tail-label metrics while performance on head labels remains on par with the best existing systems, demonstrating that gradient boosting is an effective vehicle for re-balancing deep XMC toward the long tail.

For long-tail distribution a natural solution comes in mind is data augmentation technique to compensate the scarce data \citep{zhang2022adam} \citep{xu2023labelspecific}. There are some existing works such as \cite{zhou-etal-2022-flipda} and \cite{wang-etal-2022-promda} which resort to applying Pretrained Language Models (PLM) for data augmentation in low resource setting, but these approaches struggle in multi-label scenarios due to label co-occurrence \citep{wu2020distribution}. The paper \cite{xu2024tamingprompt} introduces a novel approach, Extreme Data Augmentation (XDA), that addresses these challenges and alleviate the low quality and long-tailed problem of augmented samples by employing three key mechanisms: \textit{Prompt, Filter} and \textit{Mask}.

\paragraph{}
Motivated by \cite{lester-etal-2021-power}, \cite{wang-etal-2022-promda}, employ the softprompt technique for fine tuning by adding a sequence of trainable vectors $ P^j=\{p_1^j,...,p_k^j\}$ at each transformer layer and update the parameters of soft-prompt only while keeping the all other PLM parameters fixed.  The pre-trained model $f_0$
  employs the original training set to learn a mapping from input texts 
$x$ to feature vectors, using a text encoder $\phi$ and classifier parameters 
$w_l$, with binary cross entropy (BCE) as the loss function,
\begin{equation}
    L_{\text{BCE}} = - \left( y \log(\sigma(f(x, l))) + (1 - y) \log(1 - \sigma(f(x, l))) \right)
\end{equation}
providing essential feedback signals for the sample filtering process. In filtering, check the semantic difference between an augmented sample $x_i^j$ and its original corresponding sample $x_i$, by using KL-divergence, the divergence score $S_{\text{Div}}^{i,j}$ of augmented sample $x_i^j$ is given by
\[S_{\text{Div}}^{i,j}=D_{\text{KL}}(p(\sigma(W_0^T \phi_0(x_i^j))) \, \| \, p(y_i)), \]
which gives us the diversity between these two distributions, here $\phi_0$ is the text encoder that maps $x$ to a feature vector, $W_0$ is the weight of the trainable classifier, and for augmented process we can consider high diversity between augmented samples. We also need to check the consistency score, because low quality samples will lead to label drift \citep{zhou-etal-2022-flipda},\citep{kamalloo-etal-2022-chosen}. The consistency score $S_{Con}^{i,j}$ of augmented sample $x_i^j$ is given by \[ S_{\text{Con}}^{i,j} =\frac{1}{k_i}| \text{Top}_{k_i}(f_0(x_i^j)) \cap \text{Top}_{k_i}(f_0(x_i))|,\]
where $f_0$ denotes the pretrained model.

\paragraph{}
By jointly considering $S_{\text{Div}}^{i,j}$ and $ S_{\text{Con}}^{i,j}$, high quality samples are selected with balanced diversity and consistency.
Now in the final step, directly using augmented samples might exacerbate the long-tailed problem, to address this issue divide the head and tail labels by some threshold and during training, augmented sample-label pairs associated with head-labels are masked. The loss function incorporates a mask indicator $m=\{m_l \}_{l=1}^L$, ensuring a balanced focus on tail-labels, 
\begin{equation*}
    L_{\text{Aug}} = -\sum_{l=1}^L m_l \left[ y_l \log (\sigma(f_0(x, l))) + (1 - y_l) \log (1 - \sigma(f_0(x, l))) \right],
\end{equation*}
where $m_l = 0$ if $l$ is a head-label and $m_l = 1$ if $l$ is a tail-label. During training, the encoder $\phi_0(\cdot)$ remains frozen, and only the classifier parameters $W$ are fine-tuned.
By adding $L_{\text{Aug}}$ to $L_{\text{BCE}}$, a more balanced loss is achieved, focusing on tail-labels without exacerbating the long-tail problem.
Experimental results on three benchmark datasets (Eurlex-4K, Wiki10-31K,  AmazonCat-13K) demonstrate the effectiveness  of our proposed XDA method, particularly in
improving the performance on tail-labels.

\paragraph{}
Dual-encoder (DE) models, effective in open-domain QA \citep{lee-etal-2019-latent}, are less explored in XMLC. The paper \cite{gupta2024dualencoders} introduces novel DE modifications for XMLC tasks. DE models, used in dense retrieval, map queries and documents into a shared embedding space for efficient fast similarity search \citep{johnson2017billionscale}. They excel with limited training data and single correct answers, using separate encoders for queries and documents. Existing  DE models underperform in XMLC due to inappropriate training losses and the need for extensive parameters.
OvA-BCE does not train effectively and
InfoNCE \citep{oord2019representation} disincentivizes confident predictions. However, these losses
exhibit several limitations when applied to extreme multi-label
classification (XMLC) tasks. Traditional contrastive losses like InfoNCE are suboptimal for multi-label settings as they enforce equal scoring for all positives, penalizing confident predictions.
To address these
limitations observed with the standard InfoNCE loss (XMLC)
tasks, a re-formulation called DecoupledSoftmax loss, 
and a soft top-$k$ operator-based loss are proposed.

\paragraph{}
The decoupled softmax loss is defined as:
\[ l(q_i, y_i; s) = -\sum_{j \in [L]} y_{ij} \cdot \log \frac{e^{s(q_i, d_j)}}{e^{s(q_i, d_j)} + \sum_{l \in [L]} (1 - y_{il}) \cdot e^{s(q_i, d_l)}}, \]
where \( s(q_i, d_j) \) is the model-assigned score for the query-document pair \((q_i, d_j)\).
Here \( y_{ij} \) is 1 if document \( j \) is a positive label for query \( q_i \), and 0 otherwise.
This formulation removes positive label correlation in the denominator, improving training for DE models in XMLC, provides more consistent and unbiased gradient feedback compared to standard softmax, essential for handling the imbalanced nature of XMLC datasets.

\paragraph{}
Soft Top-$k$ Loss:
Designed to optimize prediction accuracy within a fixed budget size $k,$ particularly relevant for top-k predictions in XMLC.
\begin{align*}
\ell(q_i, y_i; s) = - \frac{1}{L} \sum_{j \in [L]} y_{ij} \log z_{ij},
\end{align*}
where \( z_i = \text{SoftTop-}k(s_i) \) is the output of the soft top-$k$ operator applied to the score vector \( s_i = (s(q_i, d_1), \ldots, s(q_i, d_L)) \). The soft top-$k$ operator, taking a score vector as input, acts as a filter by assigning values near 1 to the top-$k$ scores and near 0 to the rest, ensuring differentiability for backpropagation.

\paragraph{}
The authors \cite{gupta2024dualencoders} compared their DE approach to SOTA XMLC methods like DEXA (Dual Encoder for eXtreme Classification Applications) \citep{dahiya2023-dexa}, NGAME (Neural Graph Attention for Multi-label Extreme classification) \citep{dahiya2023ngame}, and XR-Transformer~\citep{zhang2021fast}. Using the same distilbert-base model, their method outperformed these approaches on large-scale datasets (e.g., LF-Wikipedia-500K, LF-AmazonTitles-1.3M), achieving up to $2\%$  higher Precision@1 with $20\times$ fewer parameters.
The study shows that with the right loss formulation, DE models can achieve SOTA performance in XMLC, offering parameter-efficient, generalizable solutions. Proper training losses help DE models overcome XMLC's semantic gap. The study includes a memory-efficient, distributed implementation for large datasets using gradient caching \citep{gao-etal-2021-scaling}. 

\paragraph{}
The paper \cite{zhang-etal-2023-long} proposes a novel neural retrieval framework named DEPL (Dual Encoder with Pseudo
Label) for addressing the challenge of tail-label prediction in extreme multi-label text classification (XMTC). DEPL retrieval-based model leverages the semantic matching of document and label texts via a dual encoder model \citep{gao2021unsupervised} \citep{xiong2020approximate} \citep{luan2021sparse} \citep{karpukhin-etal-2020-dense},
which are generated automatically by a statistical model with BoW features. The primary goal is to enhance the mapping between input documents and system-enhanced label descriptions, thereby improving classification performance, particularly for rare labels.

\paragraph{}
In the DEPL framework, a BoW classifier (SVM) is first trained to extract top keywords from label embeddings based on token importance. These keywords are concatenated with original label names to create pseudo descriptions. A BERT-based dense retrieval model \citep{devlin2019bert} is then used to rank labels by semantic matching between document text and these enhanced label descriptions. Pseudo labels generated this way improve the retrieval process.

\paragraph{}
The paper relies on the Johnson-Lindenstrauss (JL) lemma from random matrix theory, which provides a mathematical foundation for connecting dense and sparse classifiers. This lemma helps establish a performance lower bound for the neural model under the assumption that neural embeddings behave similarly to random matrices.

\paragraph{}
The framework was extensively tested on large benchmark datasets, including EURLex-4K, AmazonCat-13K, Wiki10-31K and Wiki-500K. DEPL demonstrated significant improvements over strong baseline models, particularly in predicting tail labels. The experiments included ablation studies to understand the impact of pseudo label length on performance.

\paragraph{}
The study acknowledges that the assumptions made for theoretical analysis may not hold in practical applications. Additionally, comparisons with other models and techniques, such as reranking losses and regularization, were not included in this work.

The paper \cite{10.1145/3539597.3570392} introduces NGAME, a novel method for negative mining in extreme classification (XC) that addresses the inefficiencies of existing techniques.  This approach allows for larger mini-batch sizes, faster convergence, and higher accuracy in training deep XC models, particularly when using large transformer-based encoders. 

\paragraph{}
NGAME merges the tasks of mini-batch creation and negative mining, eliminating the need for a separate negative mining phase. This integrated approach results in mini-batches that naturally include informative negative samples, thereby accelerating the training process.
The method leverages in-batch sampling to provide high-quality hard negatives efficiently. During training, the method encourages data points and relevant labels to have similar embeddings. This similarity is used to distinguish between relevant and irrelevant labels. By using in-batch sampling, NGAME avoids the need for external structures or additional computations to find these negatives, thereby reducing memory and computational overheads. NGAME adapts the modular training pipeline from the DeepXML paper~\citep{Dahiya_2021}. By reparameterizing label classifiers and leveraging label embeddings, the method enhances training efficiency and effectiveness.
NGAME’s algorithm clusters data point embeddings periodically and forms mini-batches from these clusters. Hard negatives are identified based on the proximity of label embeddings to data point embeddings.
The paper evaluates NGAME on multiple benchmark datasets from the Extreme Classification Repository, covering various applications such as product recommendation and Wikipedia page prediction. The method is compared against several baselines, including both Siamese (\cite{pmlr-v139-dahiya21a},\cite{mittal2021decaf},\cite{mittal2021eclare}) and non-Siamese (\cite{jiang2021lightxml}, \cite{zhang2021fast},\cite{chang2020taming}) deep XC methods. The results demonstrate that NGAME achieves up to 16\% higher accuracy than state-of-the-art methods and shows significant improvements in click-through rates in live A/B tests on a popular search engine.
\paragraph{}
The paper \cite{10.1145/3580305.3599301} introduces DEXA, a framework addressing key limitations of existing encoder-based methods in Extreme Classification (XC). Models like NGAME (\cite{dahiya2022ngame}) and XR-Transformer(\cite{chang2020taming}) rely on embedding data points and labels into a shared space based solely on textual descriptions, which often fails in the presence of semantic gaps, especially in short-text applications. Modular training strategies used by these models train encoders independently of classifiers, resulting in suboptimal embeddings, particularly for underrepresented tail labels.
\paragraph{}
DEXA overcomes these issues by introducing auxiliary parameters shared across clusters of semantically related labels. These parameters act as correction terms, enriching label representations with latent information beyond text and reducing semantic distortions. By grouping labels into clusters using pre-trained embeddings and assigning shared auxiliary vectors, DEXA ensures representational adjustments for related labels, enhancing embedding quality. Integrating auxiliary parameters directly into encoder training eliminates the disjointed phases of methods like NGAME. Unlike XR-Transformer, DEXA achieves comparable or superior performance with a lightweight design, discarding auxiliary vectors after training to maintain efficiency.
\paragraph{}
DEXA demonstrates up to 6\% improvements in precision-based metrics and significant gains for tail labels on short-text datasets. On proprietary datasets like SponsoredSearch-40M, it achieves up to 15\% accuracy gains with smaller encoders, showcasing resource efficiency. It integrates seamlessly with encoders like DistilBERT and MiniLM, delivering consistent performance improvements without major architectural changes.
\paragraph{}
The paper \cite{Qaraei2024} addresses the challenge of training deep models for extreme multi-label classification (XMC) problems, where the output space can be extremely large. To address this issue the paper motivates the use of negative sampling and focuses on Maximum Inner Product Search (MIPS)  (\cite{auvolat2015clustering},\cite{8733051},\cite{shrivastava2014asymmetric}) to identify hard negatives as an alternative to meta-classifier-based approaches(\cite{Dahiya_2021},\cite{jiang2021lightxml}), aiming to reduce computational overhead while maintaining or improving performance.
\paragraph{}
The paper highlights two significant issues in training deep models using MIPS-based negative sampling. Starting training with only hard negatives leads to high-magnitude gradients, causing large, unstable updates. Large intervals between MIPS pre-processing lead to non-informative negative samples, as the embeddings remain relatively unchanged, causing repetitive selection of the same negatives.
\paragraph{}
To mitigate the instability from using hard  negatives, the paper proposes a hybrid negative sampling approach. Select a few hard negatives using MIPS and combine them with negatives sampled from a uniform distribution. This balances the gradient magnitudes and stabilizes the training process.
\paragraph{}
To improve the efficiency of the MIPS process the authors use  a clustering-based approximate MIPS (\cite{auvolat2015clustering},\cite{8733051}). By clustering the label space, the search for hard negatives is restricted to a few representative clusters, reducing computational complexity while maintaining high-quality negative samples. By choosing an optimal number of clusters $(K = \sqrt{L})$, the method balances the computational load and the quality of the negatives retrieved, leading to a scalable and efficient training process.
\paragraph{}
Two architectures has been used as the encoders for the proposed method: a shallow neural network with a single 
hidden layer and a BERT model \citep{devlin2019bert} keeping its all the hyperparameters same as \cite{jiang2021lightxml}.
Experiments conducted on the Eurlex dataset, which includes approximately 4000 labels, validate the proposed methods.The hybrid method reaches performance levels comparable to state-of-the-art models like LightXML \citep{jiang2021lightxml}, without the additional overhead of training and storing a meta-classifier.

\paragraph{Comparison between methods}
In Table~\ref{tab:deep_methods_comp}, we classify various deep learning based methods. 

\begin{table*}[t!]
\small
\centering
\caption{\textbf{Comparison of representative deep–learning approaches for extreme multi-label classification.}  
All methods fine-tune neural text encoders but differ markedly in how they handle the extreme label set (\(L\!\gg\!10^3\)).  
``PLT'' denotes probabilistic label tree; ``ANNS'' = approximate nearest-neighbour search; ``BN'' = bottleneck layer;  
``Dyn.\ neg.'' = dynamic (online) hard-negative mining.\label{tab:deep_methods_comp}}
\begin{tabular}{@{}p{2.0cm}|p{2.3cm}|p{2.2cm}|p{1.6cm}|p{1.6cm}|p{1.1cm}|p{1.4cm}@{}}
\toprule
\textbf{Method (Year)}          & \textbf{Encoder / Backbone}                  & \textbf{Label–space reduction / Retrieval}                & \textbf{Negative sampling}     & \textbf{Tail-label specific tricks} & \textbf{Train time\footnotemark} & \textbf{Key citations} \\
\midrule
XML--CNN (2017)                 & multi-kernel CNN $+$ BN                      & none (full sigmoid layer, BN shrinks params)              & random                          & —                                  & slow          & \cite{liu2017deep}           \\[2pt]

AttentionXML (2019)             & BiLSTM $+$ multi-head attention              & shallow \textit{wide} PLT                                 & static (per node)               & label-wise attention               & very slow     & \cite{you2019attentionxml}  \\[2pt]

X–Transf. (2020)            & BERT / RoBERTa (fine-tuned)                 & balanced $k$-means clusters $\to$ two-stage matcher       & static (cluster-based)          & cluster-aware fine-tune            & medium--slow  & \cite{chang2020taming}      \\[2pt]

APLC–XLNet (2020)               & XLNet                                         & adaptive prob.\ label clusters (tree)                     & static                           & head/tail cluster split           & medium--slow  & \cite{ye2020pretrained}     \\[2pt]

LightXML (2021)                 & BERT encoder                                 & PLT (depth 2) + ANNS recall                               & \textit{dynamic} generator–disc.& online hard negatives             & medium        & \cite{jiang2021lightxml}    \\[2pt]

XR–Transf. (2021)           & BERT / RoBERTa                               & recursive hierarchical label tree                         & dynamic (per level)             & TF-IDF $\,+\,$BERT hybrid         & fast          & \cite{zhang2021fast}        \\[2pt]

SiameseXML (2021)               & dual BERT encoders                           & joint doc/label embedding $+$ ANNS                        & hard (metric learning)          & label-text augmentation           & fast          & \cite{dahiya2021siamesexml} \\[2pt]

ASTEC / DeepXML (2021)          & any transformer (plug-in)                    & 4-stage pipeline (clustering \& ANNS shortlist)           & staged hard negatives           & transfer-learning for tails        & fast          & \cite{Dahiya_2021}          \\[2pt]

NGAME (2023)                    & Siamese encoder (BERT)                       & joint embedding + ANNS                                    & \textit{batch-aware} hard negs   & label–graph mini-batching         & very fast     & \cite{dahiya2023ngame}      \\[2pt]

DEXA (2023)                     & dual encoder (BERT/Distil)                   & joint embedding + ANNS                                    & hard (metric)                   & auxiliary cluster vectors          & fast          & \cite{10.1145/3580305.3599301} \\[2pt]

BoostXML (2022)                 & AttentionXML backbone                        & PLT (same as AttentionXML)                                & inherits PLT negatives           & gradient boosting on residuals     & medium        & \cite{boostxml}      \\[2pt]

MatchXML (2022)                 & transformer (doc–label pair encoder)         & none; learns pair-wise scoring                            & in-batch (contrastive)          & static sentence + TF-IDF labels    & medium        & \cite{matchxml}       \\
\bottomrule
\end{tabular}
\end{table*}

\footnotetext{Relative to contemporary transformer baselines; “fast’’ means $<2\times$ XR–Transformer on Amazon-3M, “medium’’ $2$–$5\times$, “slow’’ $>5\times$.}

\begin{itemize}
    \item XML-CNN although the first of the deep learning methods in XMLC, is no longer a competitive method in terms of performance. This is mainly because CNN's are not the best at capturing the information from text, and hence have been outperformed by most other methods.
    \item AttentionXML and XTransformer are both methods which require a lot of training time. However, they still remain relevant due to the fact that they perform significantly well on some datasets. In particular, AttentionXML performs well on datasets with high average feature text length.
    \item DeepXML based methods such as SiameseXML, ECLARE and NGAME are currently among the best performing methods in XMLC.
\end{itemize}

\paragraph{}
Over the last five years, deep learning based methods have outperformed all other methods in terms of performance. They have much higher accuracies than tree based methods, but can be slower in terms of train and test complexities and have higher model sizes.

\subsection{LLM-Assisted Methods}
\paragraph{}
The paper \cite{zhu2024icxml} addresses the challenges of Extreme Multi-Label Classification (XMC) in real-world scenarios, particularly focusing on zero-shot settings where new labels appear which were not present in prior training data.  Traditional retrieval-based methods often struggle due to the lack of lexical or semantic overlap between queries and labels. On the other hand, large language models (LLMs) face practical issues such as high computational costs and label generation impracticality. To solve these issues, this paper introduced a novel approach of ICXML, a two stage framework, first is generation of set of candidate labels and then rerank these generated labels.
\begin{itemize}
    \item \textbf{Stage 1:  Demonstration Generation:} Demonstrations should encompass both the inherent correlation between the input text and the task label as well as external knowledge that facilitates the model's learning process in relation to the input text. This goal can be achieved by two different strategies: \begin{itemize}
        \item Content based demonstration generation: Employ a LLM, $\phi$ to generate a set of $m$ demonstration inputs $Z_i=\{z_i^1,\ldots, z_i^m\},$ where $z_i^j \sim \phi(\text{PROMPT}(x_i,t_1))$, and $t_1$ is a task description. After this compute the scores between each demonstration inputs and label points corresponding to $x_i$ using zero shot retriever $\theta$ and then consider the top $n$ labels based on the score to make the pseudo demostration set. 
        \item Label-centric Demonstration generation: In this approach, initially select top $n$ labels using zero shot retriever $\theta$ for each test instance $x_i \in X,$ and then use LLM $\phi$ to generate $m$ pseudo demonstration inputs using these top labels.  
    \end{itemize} 
    
    \item \textbf{Stage 2: Label Reranking:} After the pseudo demonstration set generation, integrate them with each test set and guide the few shot learning process of the model $\phi$ and then for each generated label, fetch some desired number of top labels from the label set using zero-shot retriever that possess high semantic similarity with generated label.
    Now steer the LLM by giving a prompt using the shortlisted labels, test instance and more refined task description to select the most suitable set of labels.
\end{itemize}
The proposed method is evaluated on datasets LF-Amazon-131K and LF-WikiSeeAlso-320K. It outperforms baseline methods in various zero-shot settings, demonstrating its effectiveness and flexibility.
\paragraph{}

\cite{li-etal-2023-enhancing-extreme} tackle three long-standing bottlenecks of LLM-assisted extreme multi-label text classification (XMTC)—inflated model size, label-sparse regimes and slow, SME-driven evaluation—by swapping the usual monolithic SciBERT classifier \citep{beltagy2019} for a light-weight label-ranking architecture and coupling it with active learning and ChatGPT-based assessment. Their system mirrors the two-stage retrieval-then-re-rank pattern popularised by X-Trans \cite{chang2020taming} and MatchXML \citep{matchxml}: a Siamese Bi-Encoder (Dense Passage Retriever style \cite{karpukhin-etal-2020-dense}) encodes document and label texts separately, optimised with MultipleNegativesRanking loss and accelerated with HNSW indexing, delivering high-recall short-lists at negligible cost; a Cross-Encoder (BERT-family \cite{devlin-etal-2019-bert,zhuang-etal-2021-robustly}) then re-ranks those candidates token-interactively to boost precision. Unlike earlier LLM-assisted methods, their Bi-Encoder training enforces label uniqueness per batch to avoid easy negatives and their pipeline supports open-set growth: new labels can be appended without full retraining because the Bi-Encoder merely embeds them and the Cross-Encoder fine-tunes on a replay buffer that mixes old-and-new examples, preventing catastrophic forgetting (similarly to the rehearsal strategy in SiameseXML \citep{pmlr-v139-dahiya21a}). Data scarcity is mitigated by a cold-start, pool-based active-learning loop that greedily acquires unlabeled documents most similar to the new label vectors; each iteration yields SME-annotated positives which refresh both encoders. For evaluation they replace exhaustive manual judging with ChatGPT-assisted scoring \cite{ren2021survey}: GPT generates relevance justifications that SMEs quickly verify, cutting review time. Across four scientific-domain datasets the resulting BiCross-Encoder surpasses the SciBERT one-vs-all baseline in F-score while matching the retrieval speed of X-Trans and delivering large recall gains on unseen labels, showing that their retrieval-rank-AL triad effectively addresses scalability, label sparsity and evaluation latency in LLM-centred XMTC.

\paragraph{}
The dataset used is derived from Elsevier’s Compendex taxonomy, containing about 11,486 engineering labels and corpus 14M interdisciplinary documents. For active learning, 30 labels were selected, with 1,000 samples per label chosen using GPL (Generative Pretrained Labels) \citep{wang-etal-2022-gpl}. The pool had 30,000 samples, and the test set had 5,000 documents. Active learning improved new label performance by 15 points, reaching Recall@10 of 0.85 after 100 iterations, while maintaining performance on old labels. This involved 1-2 newly labeled samples per iteration. GPL by \cite{wang-etal-2022-gpl} is used to adapt a dense retrieval model to a specific domain without the need for labeled training data. This is particularly useful when dealing with a large, diverse corpus where labeled data may not be available for every domain or label of interest. The Bi-Encoder model, which is initially trained on a large corpus, is further fine-tuned using GPL. This fine-tuning process involves generating pseudo labels for documents, which helps the model learn to identify relevant documents more accurately within the specific domain of interest. By using GPL, the fine-tuned Bi-Encoder can effectively rank and retrieve the most relevant documents from a large corpus. This significantly reduces the number of documents that need to be considered in each iteration of active learning, making the process more efficient.

\subsection{Multi-modal Extreme Classification}

\paragraph{}
\cite{Mittal_2022} extend extreme multi-label classification to the truly multi-modal regime by introducing MUFIN, a retrieval-then-rank framework that can handle millions of labels whose descriptors include both images and text. Whereas earlier multi-modal XC systems relied exclusively on fixed embeddings and nearest-neighbour matching \citep{velioglu2024fashionfail,Revanur_2021,tan2019learning}, MUFIN keeps the successful DeepXML pipeline \citep{Dahiya_2021} but swaps in modality-aware encoders and a cross-modal attention classifier: a ViT-32 image encoder and an msmarco-distilBERT text encoder (both projected to 192 dimensions) are jointly fine-tuned, then a cross-attention layer learns how visual and textual cues complement each other when ranking candidate labels. Training follows the four-stage DeepXML schedule—self-supervised pre-training, augmented retrieval with hard-positive/negative mining, transfer, and label-aware fine-tuning—so the overall complexity still grows only logarithmically with label count, making MUFIN practical at Amazon scale. On the MM-AmazonTitles-300K benchmark it improves Precision@k by atleast 3\% over the strongest text-only, image-only and previous multi-modal baselines, demonstrating that a classifier-based, cross-modal architecture is superior to pure embedding retrieval when descriptors span both vision and language.

\subsection{Miscellaneous}
\paragraph{}
CascadeXML~\citep{kharbanda2022cascadexmlrethinkingtransformersendtoend} introduces a novel hierarchical framework for extreme multi-label classification (XMC) by leveraging hierarchical label trees (HLTs) \citep{prabhu2018parabel} and Transformer-based embeddings. Unlike traditional transformer-based models such as  AttentionXML \citep{you2019attentionxml}, and XR-Transformer \citep{zhang2021fast}, which require extensive computational resources,  CascadeXML combines the strengths of these approaches, thus creating an end to-end trainable multi-resolution learning pipeline which trains a single transformer model across multiple resolutions in a way that allows the creation of label resolution specific attention maps and feature embeddings.

\paragraph{}
CascadeXML operates in three key stages: candidate generation, label refinement, and final classification. The first stage utilizes a lightweight model, such as XR-Linear or Parabel by \cite{prabhu2018parabel}, to generate a shortlist of candidate labels, significantly reducing computational costs. The second stage refines these candidates using deep neural networks, ensuring improved recall, especially for tail labels. The final classification step employs a distilled Transformer-based model like LightXML \citep{jiang2021lightxml}, which efficiently predicts label probabilities from a much smaller subset. By processing only a reduced candidate set, CascadeXML dramatically cuts inference time compared to transformer-based models like X-Transformer and AttentionXML, making real-time applications feasible.
A key innovation in CascadeXML is its weighted loss function, which optimizes training by balancing prediction accuracy and computational efficiency across hierarchical levels. This ensures that the model does not disproportionately favor frequent labels while maintaining overall performance. Compared to methods like XR-Transformer and LightXML, CascadeXML significantly improves tail-label recall, enhancing propensity-scored precision (PSP@3 and PSP@5) by 5-8 \%. The model is also highly memory-efficient, avoiding the need for large ensembles that often make deep-learning-based XMC methods impractical at scale. 

\paragraph{}
One of the most significant advantages of CascadeXML is its drastic reduction in training time compared to existing XMC models. For instance, X-Transformer requires 23 days on eight GPUs, whereas CascadeXML achieves comparable performance in just 24 hours using a single Nvidia A100 GPU. This efficiency makes it an ideal choice for large-scale real-world applications that require frequent model retraining. At inference time, CascadeXML is approximately 1.5 times faster than LightXML and nearly twice as fast as XR-Transformer.


\begin{figure}[t!]
        \centering
        \includegraphics[width=108mm]{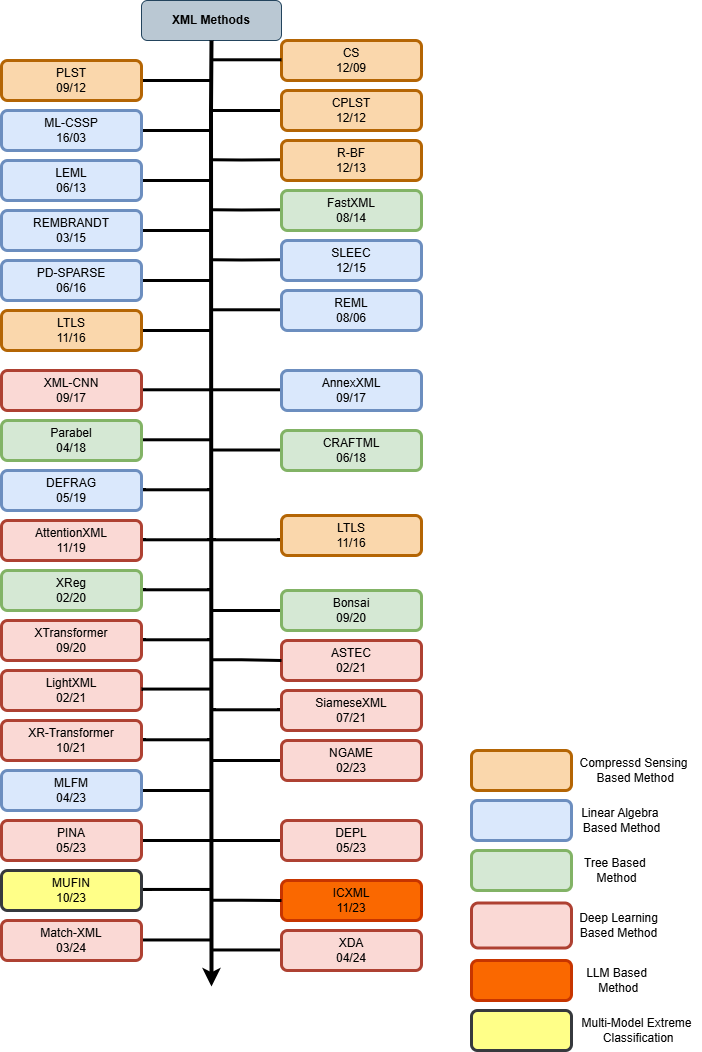}
        \caption{Timeline of Key Methods in XML Along with the Categories}
\end{figure}{}

\section{Applications of Extreme Classification}

Extreme Classification has several applications, especially in cases where the data is too large to be processed manually or semi-manually. 

\subsection{Document Tagging}
\paragraph{}
XML is highly useful for tagging documents with labels which can be selected multiple at a time. The features can be formulated as some representation of the document, while labels are the possible tags. For example, tagging of Wikipedia articles with relevant labels is a problem which is suited for XML due to the fact that there are a very large number of labels, some of which are very specific to the articles. This leads to a large tail label distribution which is typical in XML problems. 

\subsection{Product Recommendation}
\paragraph{}
Given a keyword or a ``search phrase'', the problem of determining which products to show given the product description and other information can also be converted to an extreme classification problem. The ``search phrase'' or the query can be treated as a feature while the different products act as labels. Amazon has several product recommendation datasets such as Amazon-670K, AmazonCat-13K and AmazonTitles-3M with large label spaces which serve as benchmarking datasets in XML \citep{mach_medini}.

\subsection{Advertising}
\paragraph{}
Finding the appropriate advertisements from the bid phrases given in an advertisement using the user search phrase is also a popular application of XML. Search engines such as Bing have adopted XML to highly improve their click-through rates \citep{dahiya2021deepxml}. An alternate probelem in advertizing, which is showing personalized ads using the search history of the user can also be modelled as an XML problem, with the user search history being included in the features and the advertisements as labels.

\subsection{Other Applications}
\paragraph{}
The paper \cite{valdeira2023extreme} develops a recommendation system (RS) for specialist doctor referrals, focusing on challenges like limited patient metadata and the cold-start problem. This issue arises when the system must recommend specialists for new patients who lack historical interaction data, hindering accurate recommendations due to insufficient data.
The paper addresses the cold-start problem by splitting the dataset to include both seen patients (with historical data) and new patients (without prior interactions). This approach evaluates recommendation methods under cold-start conditions, ensuring effective doctor recommendations for new patients lacking interaction history. Traditional RS methods like Collaborative Filtering (CF) and Content-Based (CB) approaches often struggle in healthcare due to their need for extensive data and explicit feedback \citep{Peito_2021}. The proposed solution utilizes XML methods, typically used in text classification, to encode patient and doctor features for recommendations.

\paragraph{}
The authors recast the doctor recommendation problem as a multi-label classification task, employing XML methods \citep{pmlr-v48-yenb16,sleec_bhatia,prabhu2014fastxml} for handling large label spaces. These methods predict relevant doctors for each patient based on their consultation history. A unified model utilizing patient history across specialties is proposed. The dataset includes patient-doctor consultations from a European healthcare provider, with demographic data but no explicit patient feedback or medical records. Patient history is converted into labels, and patient and doctor features are encoded in a TF-IDF-like manner. Various feature sets are tested to assess model performance.

\paragraph{}
The XML approach consistently outperforms state-of-the-art RS methods, especially for new patients, demonstrating significant improvements in standard recommendation metrics. It shows notable enhancements in recall metrics for both new and existing patients, indicating its promise as an alternative to traditional RS, particularly in settings with limited patient metadata. The study underscores XML's potential for creating effective and personalized doctor referral systems.

\paragraph{}
The dataset comprises 1,064 doctors, 1,003,809 patients, and 2,890,042 interactions. Data is split based on patient and visit time to prevent data leakage. Four feature groups were derived from interactions, with age normalized relative to the maximum training set age. DECAF (Deep Extreme Classification with Label Features), an XML model, outperformed benchmarks like SVD, BiVAE, and LightFM across PSnDCG@3, Recall@3, and Recall@10 metrics. DECAF demonstrates high effectiveness for specialist doctor recommendation, especially in cold-start scenarios with limited patient metadata.



\begin{figure}[t!]
    \centering
    \includegraphics[width=60mm]{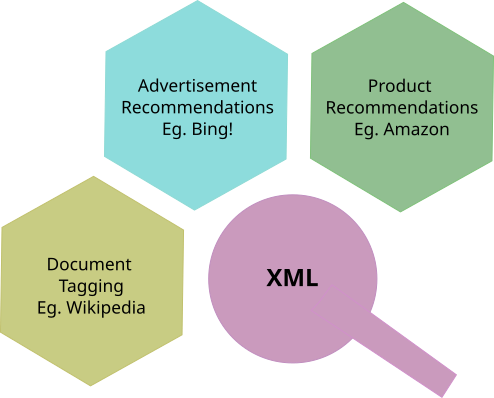}
    \caption{Practical Applications of XML}
\end{figure}{}


\acks{We would like to acknowledge support of various grants. First author Arpan and last author Pawan is funded by MAPG (Microsoft Academic Partnership Grant) and partially by Qualcomm Faculty Award. Preeti is funded by CSIR (Council of Scientific and Indistrial Research). Ankita and Kiran are funded by  UGC (University Grants Commision) Junior Research Fellowships for Doctoral program.}











\vskip 0.2in
\bibliography{xml}

\begin{thebibliography}{134}
\providecommand{\natexlab}[1]{#1}
\providecommand{\url}[1]{\texttt{#1}}
\expandafter\ifx\csname urlstyle\endcsname\relax
  \providecommand{\doi}[1]{doi: #1}\else
  \providecommand{\doi}{doi: \begingroup \urlstyle{rm}\Url}\fi

\bibitem[Auvolat et~al.(2015)Auvolat, Chandar, Vincent, Larochelle, and Bengio]{auvolat2015clustering}
Alex Auvolat, Sarath Chandar, Pascal Vincent, Hugo Larochelle, and Yoshua Bengio.
\newblock Clustering is efficient for approximate maximum inner product search, 2015.

\bibitem[Babbar and Sch\"{o}lkopf(2017)]{babbar2017dismec}
Rohit Babbar and Bernhard Sch\"{o}lkopf.
\newblock Dismec: Distributed sparse machines for extreme multi-label classification.
\newblock In \emph{Proceedings of the Tenth ACM International Conference on Web Search and Data Mining}, WSDM '17, page 721–729, New York, NY, USA, 2017. Association for Computing Machinery.
\newblock ISBN 9781450346757.
\newblock \doi{10.1145/3018661.3018741}.
\newblock URL \url{https://doi.org/10.1145/3018661.3018741}.

\bibitem[Balasubramanian and Lebanon(2012)]{moplms_balasubramanian}
Krishnakumar Balasubramanian and Guy Lebanon.
\newblock The landmark selection method for multiple output prediction.
\newblock In \emph{Proceedings of the 29th International Coference on International Conference on Machine Learning}, ICML’12, page 283–290, Madison, WI, USA, 2012. Omnipress.
\newblock ISBN 9781450312851.

\bibitem[Beltagy et~al.(2019)Beltagy, Lo, and Cohan]{beltagy2019}
Iz~Beltagy, Kyle Lo, and Arman Cohan.
\newblock {S}ci{BERT}: A pretrained language model for scientific text.
\newblock In Kentaro Inui, Jing Jiang, Vincent Ng, and Xiaojun Wan, editors, \emph{Proceedings of the 2019 Conference on Empirical Methods in Natural Language Processing and the 9th International Joint Conference on Natural Language Processing (EMNLP-IJCNLP)}, pages 3615--3620, Hong Kong, China, November 2019. Association for Computational Linguistics.
\newblock \doi{10.18653/v1/D19-1371}.
\newblock URL \url{https://aclanthology.org/D19-1371/}.

\bibitem[Bhatia et~al.(2016)Bhatia, Dahiya, Jain, Kar, Mittal, Prabhu, and Varma]{Bhatia16}
K.~Bhatia, K.~Dahiya, H.~Jain, P.~Kar, A.~Mittal, Y.~Prabhu, and M.~Varma.
\newblock The extreme classification repository: Multi-label datasets and code, 2016.
\newblock URL \url{http://manikvarma.org/downloads/XC/XMLRepository.html}.

\bibitem[Bhatia et~al.(2015)Bhatia, Jain, Kar, Varma, and Jain]{sleec_bhatia}
Kush Bhatia, Himanshu Jain, Purushottam Kar, Manik Varma, and Prateek Jain.
\newblock Sparse local embeddings for extreme multi-label classification.
\newblock In C.~Cortes, N.~Lawrence, D.~Lee, M.~Sugiyama, and R.~Garnett, editors, \emph{Advances in Neural Information Processing Systems}, volume~28. Curran Associates, Inc., 2015.
\newblock URL \url{https://proceedings.neurips.cc/paper_files/paper/2015/file/35051070e572e47d2c26c241ab88307f-Paper.pdf}.

\bibitem[Bi and Kwok(2013)]{cssp_bi}
Wei Bi and James~T. Kwok.
\newblock Efficient multi-label classification with many labels.
\newblock In \emph{Proceedings of the 30th International Conference on International Conference on Machine Learning - Volume 28}, ICML’13, page III–405–III–413. JMLR.org, 2013.

\bibitem[Blondel et~al.(2008)Blondel, Guillaume, Lambiotte, and Lefebvre]{louvain_blondel}
Vincent~D Blondel, Jean-Loup Guillaume, Renaud Lambiotte, and Etienne Lefebvre.
\newblock Fast unfolding of communities in large networks.
\newblock \emph{Journal of Statistical Mechanics: Theory and Experiment}, 2008\penalty0 (10):\penalty0 P10008, oct 2008.
\newblock \doi{10.1088/1742-5468/2008/10/p10008}.

\bibitem[Bloom(1970)]{bloom_bloom}
Burton~H. Bloom.
\newblock Space/time trade-offs in hash coding with allowable errors.
\newblock \emph{Commun. ACM}, 13\penalty0 (7):\penalty0 422–426, July 1970.
\newblock ISSN 0001-0782.
\newblock \doi{10.1145/362686.362692}.

\bibitem[Boutsidis et~al.(2009)Boutsidis, Mahoney, and Drineas]{cssp_boutsidis}
Christos Boutsidis, Michael~W. Mahoney, and Petros Drineas.
\newblock An improved approximation algorithm for the column subset selection problem.
\newblock In \emph{Proceedings of the Twentieth Annual ACM-SIAM Symposium on Discrete Algorithms}, SODA ’09, page 968–977, USA, 2009. Society for Industrial and Applied Mathematics.

\bibitem[Chang et~al.(2019)Chang, Yu, Zhong, Yang, and Dhillon]{chang2019modular}
Wei{-}Cheng Chang, Hsiang{-}Fu Yu, Kai Zhong, Yiming Yang, and Inderjit~S. Dhillon.
\newblock A modular deep learning approach for extreme multi-label text classification.
\newblock \emph{CoRR}, abs/1905.02331, 2019.
\newblock URL \url{http://arxiv.org/abs/1905.02331}.

\bibitem[Chang et~al.(2020)Chang, Yu, Zhong, Yang, and Dhillon]{chang2020taming}
Wei-Cheng Chang, Hsiang-Fu Yu, Kai Zhong, Yiming Yang, and Inderjit~S Dhillon.
\newblock Taming pretrained transformers for extreme multi-label text classification.
\newblock In \emph{Proceedings of the 26th ACM SIGKDD International Conference on Knowledge Discovery \& Data Mining}, pages 3163--3171, 2020.

\bibitem[Charikar et~al.(2002)Charikar, Chen, and Farach-Colton]{sketch_moses}
Moses Charikar, Kevin Chen, and Martin Farach-Colton.
\newblock Finding frequent items in data streams.
\newblock In \emph{Proceedings of the 29th International Colloquium on Automata, Languages and Programming}, ICALP ’02, page 693–703, Berlin, Heidelberg, 2002. Springer-Verlag.
\newblock ISBN 3540438645.

\bibitem[Chen et~al.(2020)Chen, Kornblith, Norouzi, and Hinton]{chen2020simple}
Ting Chen, Simon Kornblith, Mohammad Norouzi, and Geoffrey Hinton.
\newblock A simple framework for contrastive learning of visual representations.
\newblock In \emph{International conference on machine learning}, pages 1597--1607. PMLR, 2020.

\bibitem[Chen(2015)]{chen2015convolutional}
Yahui Chen.
\newblock Convolutional neural network for sentence classification.
\newblock Master's thesis, University of Waterloo, 2015.

\bibitem[Chien et~al.(2022)Chien, Chang, Hsieh, Yu, Zhang, Milenkovic, and Dhillon]{chien2022node}
Eli Chien, Wei-Cheng Chang, Cho-Jui Hsieh, Hsiang-Fu Yu, Jiong Zhang, Olgica Milenkovic, and Inderjit~S Dhillon.
\newblock Node feature extraction by self-supervised multi-scale neighborhood prediction, 2022.

\bibitem[Chien et~al.(2023)Chien, Zhang, Hsieh, Jiang, Chang, Milenkovic, and Yu]{chien2023pina}
Eli Chien, Jiong Zhang, Cho-Jui Hsieh, Jyun-Yu Jiang, Wei-Cheng Chang, Olgica Milenkovic, and Hsiang-Fu Yu.
\newblock Pina: Leveraging side information in extreme multi-label classification via predicted instance neighborhood aggregation, 2023.

\bibitem[Ciss\'{e} et~al.(2013)Ciss\'{e}, Usunier, Artieres, and Gallinari]{bloom_cisse}
Moustapha Ciss\'{e}, Nicolas Usunier, Thierry Artieres, and Patrick Gallinari.
\newblock Robust bloom filters for large multilabel classification tasks.
\newblock In \emph{Proceedings of the 26th International Conference on Neural Information Processing Systems - Volume 2}, NIPS’13, page 1851–1859, Red Hook, NY, USA, 2013. Curran Associates Inc.

\bibitem[Dahiya et~al.(2023{\natexlab{a}})Dahiya, Gupta, Saini, Soni, Wang, Dave, Jiao, Gururaj, Dey, Singh, Hada, Jain, Paliwal, Mittal, Mehta, Ramjee, Agarwal, Kar, and Varma]{dahiya2023ngame}
K.~Dahiya, N.~Gupta, Deepak Saini, A.~Soni, Y.~Wang, K.~Dave, J.~Jiao, K.~Gururaj, P.~Dey, A.~Singh, D.~Hada, V.~Jain, B.~Paliwal, A.~Mittal, S.~Mehta, R.~Ramjee, S.~Agarwal, P.~Kar, and Manik Varma.
\newblock Ngame: Negative mining-aware mini-batching for extreme classification.
\newblock In \emph{ACM International Conference on Web Search and Data Mining, Singapore}, March 2023{\natexlab{a}}.

\bibitem[Dahiya et~al.(2021{\natexlab{a}})Dahiya, Agarwal, Saini, Gururaj, Jiao, Singh, Agarwal, Kar, and Varma]{dahiya2021siamesexml}
Kunal Dahiya, Ananye Agarwal, Deepak Saini, K~Gururaj, Jian Jiao, Amit Singh, Sumeet Agarwal, Purushottam Kar, and Manik Varma.
\newblock Siamesexml: Siamese networks meet extreme classifiers with 100m labels.
\newblock In \emph{International Conference on Machine Learning}, pages 2330--2340. PMLR, 2021{\natexlab{a}}.

\bibitem[Dahiya et~al.(2021{\natexlab{b}})Dahiya, Agarwal, Saini, K, Jiao, Singh, Agarwal, Kar, and Varma]{pmlr-v139-dahiya21a}
Kunal Dahiya, Ananye Agarwal, Deepak Saini, Gururaj K, Jian Jiao, Amit Singh, Sumeet Agarwal, Purushottam Kar, and Manik Varma.
\newblock Siamesexml: Siamese networks meet extreme classifiers with 100m labels.
\newblock In Marina Meila and Tong Zhang, editors, \emph{Proceedings of the 38th International Conference on Machine Learning}, volume 139 of \emph{Proceedings of Machine Learning Research}, pages 2330--2340. PMLR, 18--24 Jul 2021{\natexlab{b}}.
\newblock URL \url{https://proceedings.mlr.press/v139/dahiya21a.html}.

\bibitem[Dahiya et~al.(2021{\natexlab{c}})Dahiya, Saini, Mittal, Shaw, Dave, Soni, Jain, Agarwal, and Varma]{Dahiya_2021}
Kunal Dahiya, Deepak Saini, Anshul Mittal, Ankush Shaw, Kushal Dave, Akshay Soni, Himanshu Jain, Sumeet Agarwal, and Manik Varma.
\newblock Deepxml: A deep extreme multi-label learning framework applied to short text documents.
\newblock In \emph{Proceedings of the 14th ACM International Conference on Web Search and Data Mining}, WSDM ’21. ACM, March 2021{\natexlab{c}}.
\newblock \doi{10.1145/3437963.3441810}.
\newblock URL \url{http://dx.doi.org/10.1145/3437963.3441810}.

\bibitem[Dahiya et~al.(2021{\natexlab{d}})Dahiya, Saini, Mittal, Shaw, Dave, Soni, Jain, Agarwal, and Varma]{dahiya2021deepxml}
Kunal Dahiya, Deepak Saini, Anshul Mittal, Ankush Shaw, Kushal Dave, Akshay Soni, Himanshu Jain, Sumeet Agarwal, and Manik Varma.
\newblock Deepxml: A deep extreme multi-label learning framework applied to short text documents.
\newblock In \emph{Proceedings of the 14th ACM International Conference on Web Search and Data Mining}, pages 31--39, 2021{\natexlab{d}}.

\bibitem[Dahiya et~al.(2022)Dahiya, Gupta, Saini, Soni, Wang, Dave, Jiao, Dey, Singh, Hada, et~al.]{dahiya2022ngame}
Kunal Dahiya, Nilesh Gupta, Deepak Saini, Akshay Soni, Yajun Wang, Kushal Dave, Jian Jiao, Prasenjit Dey, Amit Singh, Deepesh Hada, et~al.
\newblock Ngame: Negative mining-aware mini-batching for extreme classification.
\newblock \emph{arXiv preprint arXiv:2207.04452}, 2022.

\bibitem[Dahiya et~al.(2023{\natexlab{b}})Dahiya, Gupta, Saini, Soni, Wang, Dave, Jiao, K, Dey, Singh, Hada, Jain, Paliwal, Mittal, Mehta, Ramjee, Agarwal, Kar, and Varma]{10.1145/3539597.3570392}
Kunal Dahiya, Nilesh Gupta, Deepak Saini, Akshay Soni, Yajun Wang, Kushal Dave, Jian Jiao, Gururaj K, Prasenjit Dey, Amit Singh, Deepesh Hada, Vidit Jain, Bhawna Paliwal, Anshul Mittal, Sonu Mehta, Ramachandran Ramjee, Sumeet Agarwal, Purushottam Kar, and Manik Varma.
\newblock Ngame: Negative mining-aware mini-batching for extreme classification.
\newblock In \emph{Proceedings of the Sixteenth ACM International Conference on Web Search and Data Mining}, WSDM '23, page 258–266, New York, NY, USA, 2023{\natexlab{b}}. Association for Computing Machinery.
\newblock ISBN 9781450394079.
\newblock \doi{10.1145/3539597.3570392}.
\newblock URL \url{https://doi.org/10.1145/3539597.3570392}.

\bibitem[Dahiya et~al.(2023{\natexlab{c}})Dahiya, Yadav, Sondhi, Saini, Mehta, Jiao, Agarwal, Kar, and Varma]{10.1145/3580305.3599301}
Kunal Dahiya, Sachin Yadav, Sushant Sondhi, Deepak Saini, Sonu Mehta, Jian Jiao, Sumeet Agarwal, Purushottam Kar, and Manik Varma.
\newblock Deep encoders with auxiliary parameters for extreme classification.
\newblock In \emph{Proceedings of the 29th ACM SIGKDD Conference on Knowledge Discovery and Data Mining}, KDD '23, page 358–367, New York, NY, USA, 2023{\natexlab{c}}. Association for Computing Machinery.
\newblock ISBN 9798400701030.
\newblock \doi{10.1145/3580305.3599301}.
\newblock URL \url{https://doi.org/10.1145/3580305.3599301}.

\bibitem[Dahiya et~al.(2023{\natexlab{d}})Dahiya, Yadav, Sondhi, Saini, Mehta, Jiao, Agarwal, Kar, and Varma]{dahiya2023-dexa}
Kunal Dahiya, Sachin Yadav, Sushant Sondhi, Deepak Saini, Sonu Mehta, Jian Jiao, Sumeet Agarwal, Purushottam Kar, and Manik Varma.
\newblock Deep encoders with auxiliary parameters for extreme classification.
\newblock pages 358--367, 08 2023{\natexlab{d}}.
\newblock \doi{10.1145/3580305.3599301}.

\bibitem[Devlin et~al.(2018)Devlin, Chang, Lee, and Toutanova]{devlin2018bert}
Jacob Devlin, Ming-Wei Chang, Kenton Lee, and Kristina Toutanova.
\newblock Bert: Pre-training of deep bidirectional transformers for language understanding, 2018.

\bibitem[Devlin et~al.(2019{\natexlab{a}})Devlin, Chang, Lee, and Toutanova]{devlin-etal-2019-bert}
Jacob Devlin, Ming-Wei Chang, Kenton Lee, and Kristina Toutanova.
\newblock {BERT}: Pre-training of deep bidirectional transformers for language understanding.
\newblock In Jill Burstein, Christy Doran, and Thamar Solorio, editors, \emph{Proceedings of the 2019 Conference of the North {A}merican Chapter of the Association for Computational Linguistics: Human Language Technologies, Volume 1 (Long and Short Papers)}, pages 4171--4186, Minneapolis, Minnesota, June 2019{\natexlab{a}}. Association for Computational Linguistics.
\newblock \doi{10.18653/v1/N19-1423}.
\newblock URL \url{https://aclanthology.org/N19-1423}.

\bibitem[Devlin et~al.(2019{\natexlab{b}})Devlin, Chang, Lee, and Toutanova]{devlin2019bert}
Jacob Devlin, Ming-Wei Chang, Kenton Lee, and Kristina Toutanova.
\newblock Bert: Pre-training of deep bidirectional transformers for language understanding, 2019{\natexlab{b}}.

\bibitem[Dong et~al.(2011)Dong, Moses, and Li]{knng_dong}
Wei Dong, Charikar Moses, and Kai Li.
\newblock Efficient k-nearest neighbor graph construction for generic similarity measures.
\newblock In \emph{Proceedings of the 20th International Conference on World Wide Web}, WWW '11, page 577–586, New York, NY, USA, 2011. Association for Computing Machinery.
\newblock ISBN 9781450306324.
\newblock \doi{10.1145/1963405.1963487}.

\bibitem[Evron et~al.(2018)Evron, Moroshko, and Crammer]{evron2018efficient}
Itay Evron, Edward Moroshko, and Koby Crammer.
\newblock Efficient loss-based decoding on graphs for extreme classification.
\newblock \emph{Advances in Neural Information Processing Systems}, 31, 2018.

\bibitem[Friedland and Torokhti(2006)]{rank_friedland}
Shmuel Friedland and Anatoli Torokhti.
\newblock Generalized rank-constrained matrix approximations, 2006.

\bibitem[Gao and Callan(2021)]{gao2021unsupervised}
Luyu Gao and Jamie Callan.
\newblock Unsupervised corpus aware language model pre-training for dense passage retrieval, 2021.

\bibitem[Gao et~al.(2021)Gao, Zhang, Han, and Callan]{gao-etal-2021-scaling}
Luyu Gao, Yunyi Zhang, Jiawei Han, and Jamie Callan.
\newblock Scaling deep contrastive learning batch size under memory limited setup.
\newblock In Anna Rogers, Iacer Calixto, Ivan Vuli{\'c}, Naomi Saphra, Nora Kassner, Oana-Maria Camburu, Trapit Bansal, and Vered Shwartz, editors, \emph{Proceedings of the 6th Workshop on Representation Learning for NLP (RepL4NLP-2021)}, pages 316--321, Online, August 2021. Association for Computational Linguistics.
\newblock \doi{10.18653/v1/2021.repl4nlp-1.31}.
\newblock URL \url{https://aclanthology.org/2021.repl4nlp-1.31}.

\bibitem[Gupta et~al.(2024)Gupta, Khatri, Rawat, Bhojanapalli, Jain, and Dhillon]{gupta2024dualencoders}
Nilesh Gupta, Devvrit Khatri, Ankit~S Rawat, Srinadh Bhojanapalli, Prateek Jain, and Inderjit Dhillon.
\newblock Dual-encoders for extreme multi-label classification, 2024.

\bibitem[Halko et~al.(2011)Halko, Martinsson, and Tropp]{random_halko}
N.~Halko, P.~G. Martinsson, and J.~A. Tropp.
\newblock Finding structure with randomness: Probabilistic algorithms for constructing approximate matrix decompositions.
\newblock \emph{SIAM Review}, 53\penalty0 (2):\penalty0 217--288, 2011.
\newblock \doi{10.1137/090771806}.

\bibitem[Hamilton et~al.(2018)Hamilton, Ying, and Leskovec]{hamilton2018inductive}
William~L. Hamilton, Rex Ying, and Jure Leskovec.
\newblock Inductive representation learning on large graphs, 2018.

\bibitem[Hsu et~al.(2009)Hsu, Kakade, Langford, and Zhang]{cs_hsu}
Daniel~J. Hsu, Sham~M. Kakade, John Langford, and Tong Zhang.
\newblock Multi-label prediction via compressed sensing.
\newblock \emph{CoRR}, abs/0902.1284, 2009.

\bibitem[Jain et~al.(2016{\natexlab{a}})Jain, Prabhu, and Varma]{PfastreXML}
Himanshu Jain, Yashoteja Prabhu, and Manik Varma.
\newblock Extreme multi-label loss functions for recommendation, tagging, ranking and other missing label applications.
\newblock In \emph{Proceedings of the 22nd ACM SIGKDD International Conference on Knowledge Discovery and Data Mining}, KDD '16, page 935–944, New York, NY, USA, 2016{\natexlab{a}}. Association for Computing Machinery.
\newblock ISBN 9781450342322.
\newblock \doi{10.1145/2939672.2939756}.
\newblock URL \url{https://doi.org/10.1145/2939672.2939756}.

\bibitem[Jain et~al.(2016{\natexlab{b}})Jain, Prabhu, and Varma]{jain2016extreme}
Himanshu Jain, Yashoteja Prabhu, and Manik Varma.
\newblock Extreme multi-label loss functions for recommendation, tagging, ranking \& other missing label applications.
\newblock In \emph{Proceedings of the 22nd ACM SIGKDD International Conference on Knowledge Discovery and Data Mining}, pages 935--944, 2016{\natexlab{b}}.

\bibitem[Jain et~al.(2019)Jain, Balasubramanian, Chunduri, and Varma]{jain2019slice}
Himanshu Jain, Venkatesh Balasubramanian, Bhanu Chunduri, and Manik Varma.
\newblock Slice: Scalable linear extreme classifiers trained on 100 million labels for related searches.
\newblock In \emph{Proceedings of the Twelfth ACM International Conference on Web Search and Data Mining}, WSDM '19, page 528–536, New York, NY, USA, 2019. Association for Computing Machinery.
\newblock ISBN 9781450359405.
\newblock \doi{10.1145/3289600.3290979}.
\newblock URL \url{https://doi.org/10.1145/3289600.3290979}.

\bibitem[Jain et~al.(2010)Jain, Meka, and Dhillon]{svp_jain}
Prateek Jain, Raghu Meka, and Inderjit Dhillon.
\newblock Guaranteed rank minimization via singular value projection.
\newblock In \emph{Proceedings of the 23rd International Conference on Neural Information Processing Systems - Volume 1}, NIPS'10, pages 937--945, USA, 2010. Curran Associates Inc.
\newblock URL \url{http://dl.acm.org/citation.cfm?id=2997189.2997294}.

\bibitem[Jalan and Kar(2019)]{defrag_jalan}
Ankit Jalan and Purushottam Kar.
\newblock Accelerating extreme classification via adaptive feature agglomeration.
\newblock \emph{CoRR}, abs/1905.11769, 2019.

\bibitem[Jasinska and Karampatziakis(2016)]{jasinska2016log}
Kalina Jasinska and Nikos Karampatziakis.
\newblock Log-time and log-space extreme classification.
\newblock \emph{arXiv preprint arXiv:1611.01964}, 2016.

\bibitem[Jiang et~al.(2021{\natexlab{a}})Jiang, Wang, Sun, Yang, Zhao, and Zhuang]{LightXML}
Ting Jiang, Deqing Wang, Leilei Sun, Huayi Yang, Zhengyang Zhao, and Fuzhen Zhuang.
\newblock Lightxml: Transformer with dynamic negative sampling for high-performance extreme multi-label text classification.
\newblock \emph{CoRR}, abs/2101.03305, 2021{\natexlab{a}}.
\newblock URL \url{https://arxiv.org/abs/2101.03305}.

\bibitem[Jiang et~al.(2021{\natexlab{b}})Jiang, Wang, Sun, Yang, Zhao, and Zhuang]{jiang2021lightxml}
Ting Jiang, Deqing Wang, Leilei Sun, Huayi Yang, Zhengyang Zhao, and Fuzhen Zhuang.
\newblock Lightxml: Transformer with dynamic negative sampling for high-performance extreme multi-label text classification.
\newblock In \emph{Proceedings of the AAAI Conference on Artificial Intelligence}, volume~35, pages 7987--7994, 2021{\natexlab{b}}.

\bibitem[Johnson et~al.(2017)Johnson, Douze, and Jégou]{johnson2017billionscale}
Jeff Johnson, Matthijs Douze, and Hervé Jégou.
\newblock Billion-scale similarity search with gpus, 2017.

\bibitem[Johnson et~al.(2021)Johnson, Douze, and Jégou]{8733051}
Jeff Johnson, Matthijs Douze, and Hervé Jégou.
\newblock Billion-scale similarity search with gpus.
\newblock \emph{IEEE Transactions on Big Data}, 7\penalty0 (3):\penalty0 535--547, July 2021.
\newblock ISSN 2332-7790.
\newblock \doi{10.1109/TBDATA.2019.2921572}.

\bibitem[Johnson and Zhang(2014)]{johnson2014effective}
Rie Johnson and Tong Zhang.
\newblock Effective use of word order for text categorization with convolutional neural networks.
\newblock \emph{arXiv preprint arXiv:1412.1058}, 2014.

\bibitem[Joulin et~al.(2016)Joulin, Grave, Bojanowski, and Mikolov]{joulin2016bag}
Armand Joulin, Edouard Grave, Piotr Bojanowski, and Tomas Mikolov.
\newblock Bag of tricks for efficient text classification.
\newblock \emph{arXiv preprint arXiv:1607.01759}, 2016.

\bibitem[Kamalloo et~al.(2022)Kamalloo, Rezagholizadeh, and Ghodsi]{kamalloo-etal-2022-chosen}
Ehsan Kamalloo, Mehdi Rezagholizadeh, and Ali Ghodsi.
\newblock When chosen wisely, more data is what you need: A universal sample-efficient strategy for data augmentation.
\newblock In Smaranda Muresan, Preslav Nakov, and Aline Villavicencio, editors, \emph{Findings of the Association for Computational Linguistics: ACL 2022}, pages 1048--1062, Dublin, Ireland, May 2022. Association for Computational Linguistics.
\newblock \doi{10.18653/v1/2022.findings-acl.84}.
\newblock URL \url{https://aclanthology.org/2022.findings-acl.84}.

\bibitem[Kang et~al.(2020)Kang, Xie, Rohrbach, Yan, Gordo, Feng, and Kalantidis]{Kang2020Decoupling}
Bingyi Kang, Saining Xie, Marcus Rohrbach, Zhicheng Yan, Albert Gordo, Jiashi Feng, and Yannis Kalantidis.
\newblock Decoupling representation and classifier for long-tailed recognition.
\newblock In \emph{International Conference on Learning Representations}, 2020.
\newblock URL \url{https://openreview.net/forum?id=r1gRTCVFvB}.

\bibitem[Karpukhin et~al.(2020)Karpukhin, Oguz, Min, Lewis, Wu, Edunov, Chen, and Yih]{karpukhin-etal-2020-dense}
Vladimir Karpukhin, Barlas Oguz, Sewon Min, Patrick Lewis, Ledell Wu, Sergey Edunov, Danqi Chen, and Wen-tau Yih.
\newblock Dense passage retrieval for open-domain question answering.
\newblock In Bonnie Webber, Trevor Cohn, Yulan He, and Yang Liu, editors, \emph{Proceedings of the 2020 Conference on Empirical Methods in Natural Language Processing (EMNLP)}, pages 6769--6781, Online, November 2020. Association for Computational Linguistics.
\newblock \doi{10.18653/v1/2020.emnlp-main.550}.
\newblock URL \url{https://aclanthology.org/2020.emnlp-main.550}.

\bibitem[Khandagale et~al.(2020)Khandagale, Xiao, and Babbar]{khandagale2020bonsai}
Sujay Khandagale, Han Xiao, and Rohit Babbar.
\newblock Bonsai: diverse and shallow trees for extreme multi-label classification.
\newblock \emph{Machine Learning}, 109\penalty0 (11):\penalty0 2099--2119, 2020.

\bibitem[Kharbanda et~al.(2022)Kharbanda, Banerjee, Schultheis, and Babbar]{kharbanda2022cascadexmlrethinkingtransformersendtoend}
Siddhant Kharbanda, Atmadeep Banerjee, Erik Schultheis, and Rohit Babbar.
\newblock Cascadexml: Rethinking transformers for end-to-end multi-resolution training in extreme multi-label classification, 2022.
\newblock URL \url{https://arxiv.org/abs/2211.00640}.

\bibitem[Kim(2014)]{kim2014}
Yoon Kim.
\newblock Convolutional neural networks for sentence classification.
\newblock In Alessandro Moschitti, Bo~Pang, and Walter Daelemans, editors, \emph{Proceedings of the 2014 Conference on Empirical Methods in Natural Language Processing ({EMNLP})}, pages 1746--1751, Doha, Qatar, October 2014. Association for Computational Linguistics.
\newblock \doi{10.3115/v1/D14-1181}.
\newblock URL \url{https://aclanthology.org/D14-1181/}.

\bibitem[Kumar(2021)]{kumar2021DXML}
Pawan Kumar.
\newblock Dxml: Distributed extreme multilabel classification.
\newblock In Satish~Narayana Srirama, Jerry Chun-Wei Lin, Raj Bhatnagar, Sonali Agarwal, and P.~Krishna Reddy, editors, \emph{Big Data Analytics}, pages 311--321, Cham, 2021. Springer International Publishing.
\newblock ISBN 978-3-030-93620-4.

\bibitem[Lacoste-Julien and Jaggi(2015)]{lacoste2015global}
Simon Lacoste-Julien and Martin Jaggi.
\newblock On the global linear convergence of frank-wolfe optimization variants.
\newblock \emph{Advances in neural information processing systems}, 28, 2015.

\bibitem[Lacoste-Julien et~al.(2013)Lacoste-Julien, Jaggi, Schmidt, and Pletscher]{lacoste2013block}
Simon Lacoste-Julien, Martin Jaggi, Mark Schmidt, and Patrick Pletscher.
\newblock Block-coordinate frank-wolfe optimization for structural svms.
\newblock In \emph{International Conference on Machine Learning}, pages 53--61. PMLR, 2013.

\bibitem[Lee et~al.(2019)Lee, Chang, and Toutanova]{lee-etal-2019-latent}
Kenton Lee, Ming-Wei Chang, and Kristina Toutanova.
\newblock Latent retrieval for weakly supervised open domain question answering.
\newblock In Anna Korhonen, David Traum, and Llu{\'\i}s M{\`a}rquez, editors, \emph{Proceedings of the 57th Annual Meeting of the Association for Computational Linguistics}, pages 6086--6096, Florence, Italy, July 2019. Association for Computational Linguistics.
\newblock \doi{10.18653/v1/P19-1612}.
\newblock URL \url{https://aclanthology.org/P19-1612}.

\bibitem[Lester et~al.(2021)Lester, Al-Rfou, and Constant]{lester-etal-2021-power}
Brian Lester, Rami Al-Rfou, and Noah Constant.
\newblock The power of scale for parameter-efficient prompt tuning.
\newblock In Marie-Francine Moens, Xuanjing Huang, Lucia Specia, and Scott Wen-tau Yih, editors, \emph{Proceedings of the 2021 Conference on Empirical Methods in Natural Language Processing}, pages 3045--3059, Online and Punta Cana, Dominican Republic, November 2021. Association for Computational Linguistics.
\newblock \doi{10.18653/v1/2021.emnlp-main.243}.
\newblock URL \url{https://aclanthology.org/2021.emnlp-main.243}.

\bibitem[Li et~al.(2023)Li, Zhu, van~de Loo, Masip~Gomez, Yadav, Tsatsaronis, and Afzal]{li-etal-2023-enhancing-extreme}
Dan Li, Zi~Long Zhu, Janneke van~de Loo, Agnes Masip~Gomez, Vikrant Yadav, Georgios Tsatsaronis, and Zubair Afzal.
\newblock Enhancing extreme multi-label text classification: Addressing challenges in model, data, and evaluation.
\newblock In Mingxuan Wang and Imed Zitouni, editors, \emph{Proceedings of the 2023 Conference on Empirical Methods in Natural Language Processing: Industry Track}, pages 313--321, Singapore, December 2023. Association for Computational Linguistics.
\newblock \doi{10.18653/v1/2023.emnlp-industry.30}.
\newblock URL \url{https://aclanthology.org/2023.emnlp-industry.30/}.

\bibitem[Li et~al.(2024)Li, Zuo, Lin, and Wu]{boostxml}
Fengzhi Li, Yuan Zuo, Hao Lin, and Junjie Wu.
\newblock Boostxml: Gradient boosting for extreme multilabel text classification with tail labels.
\newblock \emph{IEEE Transactions on Neural Networks and Learning Systems}, 35\penalty0 (11):\penalty0 15292--15305, 2024.
\newblock \doi{10.1109/TNNLS.2023.3285294}.

\bibitem[Liu et~al.(2015)Liu, Zhang, Ye, Zhao, and Li]{liu2015mlrf}
Feng Liu, Xiaofeng Zhang, Yunming Ye, Yahong Zhao, and Yan Li.
\newblock Mlrf: multi-label classification through random forest with label-set partition.
\newblock In \emph{International conference on intelligent computing}, pages 407--418. Springer, 2015.

\bibitem[Liu et~al.(2017{\natexlab{a}})Liu, Chang, Wu, and Yang]{XML_CNN}
Jingzhou Liu, Wei-Cheng Chang, Yuexin Wu, and Yiming Yang.
\newblock Deep learning for extreme multi-label text classification.
\newblock In \emph{SIGIR '17}, 2017{\natexlab{a}}.

\bibitem[Liu et~al.(2017{\natexlab{b}})Liu, Chang, Wu, and Yang]{liu2017deep}
Jingzhou Liu, Wei-Cheng Chang, Yuexin Wu, and Yiming Yang.
\newblock Deep learning for extreme multi-label text classification.
\newblock In \emph{Proceedings of the 40th international ACM SIGIR conference on research and development in information retrieval}, pages 115--124, 2017{\natexlab{b}}.

\bibitem[Liu et~al.(2019)Liu, Ott, Goyal, Du, Joshi, Chen, Levy, Lewis, Zettlemoyer, and Stoyanov]{liu2019roberta}
Yinhan Liu, Myle Ott, Naman Goyal, Jingfei Du, Mandar Joshi, Danqi Chen, Omer Levy, Mike Lewis, Luke Zettlemoyer, and Veselin Stoyanov.
\newblock Roberta: A robustly optimized bert pretraining approach.
\newblock \emph{arXiv preprint arXiv:1907.11692}, 2019.

\bibitem[Loza~Menc{\'\i}a and F{\"u}rnkranz(2008)]{loza2008efficient}
Eneldo Loza~Menc{\'\i}a and Johannes F{\"u}rnkranz.
\newblock Efficient pairwise multilabel classification for large-scale problems in the legal domain.
\newblock In \emph{Joint European Conference on Machine Learning and Knowledge Discovery in Databases}, pages 50--65. Springer, 2008.

\bibitem[Luan et~al.(2021)Luan, Eisenstein, Toutanova, and Collins]{luan2021sparse}
Yi~Luan, Jacob Eisenstein, Kristina Toutanova, and Michael Collins.
\newblock Sparse, dense, and attentional representations for text retrieval, 2021.

\bibitem[McAuley and Leskovec(2013)]{mcauley2013hidden}
Julian McAuley and Jure Leskovec.
\newblock Hidden factors and hidden topics: understanding rating dimensions with review text.
\newblock In \emph{Proceedings of the 7th ACM conference on Recommender systems}, pages 165--172, 2013.

\bibitem[McAuley et~al.(2015{\natexlab{a}})McAuley, Pandey, and Leskovec]{mcauley2015inferring}
Julian McAuley, Rahul Pandey, and Jure Leskovec.
\newblock Inferring networks of substitutable and complementary products.
\newblock In \emph{Proceedings of the 21th ACM SIGKDD international conference on knowledge discovery and data mining}, pages 785--794, 2015{\natexlab{a}}.

\bibitem[McAuley et~al.(2015{\natexlab{b}})McAuley, Targett, Shi, and Van Den~Hengel]{mcauley2015image}
Julian McAuley, Christopher Targett, Qinfeng Shi, and Anton Van Den~Hengel.
\newblock Image-based recommendations on styles and substitutes.
\newblock In \emph{Proceedings of the 38th international ACM SIGIR conference on research and development in information retrieval}, pages 43--52, 2015{\natexlab{b}}.

\bibitem[Medini et~al.(2019)Medini, Huang, Wang, Mohan, and Shrivastava]{mach_medini}
Tharun Kumar~Reddy Medini, Qixuan Huang, Yiqiu Wang, Vijai Mohan, and Anshumali Shrivastava.
\newblock Extreme classification in log memory using count-min sketch: A case study of amazon search with 50m products.
\newblock In H.~Wallach, H.~Larochelle, A.~Beygelzimer, F.~d~

\bibitem[Mineiro and Karampatziakis(2014)]{rembrandt_mineiro}
Paul Mineiro and Nikos Karampatziakis.
\newblock Fast label embeddings for extremely large output spaces.
\newblock \emph{CoRR}, abs/1412.6547, 2014.
\newblock URL \url{http://arxiv.org/abs/1412.6547}.

\bibitem[Mishra et~al.(2023)Mishra, Dasgupta, Jawanpuria, Mishra, and Kumar]{lightDXML}
Istasis Mishra, Arpan Dasgupta, Pratik Jawanpuria, Bamdev Mishra, and Pawan Kumar.
\newblock Light-weight deep extreme multilabel classification.
\newblock In \emph{2023 International Joint Conference on Neural Networks (IJCNN)}, pages 1--8, 2023.
\newblock \doi{10.1109/IJCNN54540.2023.10191716}.

\bibitem[Mittal et~al.(2021{\natexlab{a}})Mittal, Dahiya, Agrawal, Saini, Agarwal, Kar, and Varma]{mittal2021decaf}
Anshul Mittal, Kunal Dahiya, Sheshansh Agrawal, Deepak Saini, Sumeet Agarwal, Purushottam Kar, and Manik Varma.
\newblock Decaf: Deep extreme classification with label features.
\newblock In \emph{Proceedings of the 14th ACM International Conference on Web Search and Data Mining}, WSDM '21, page 49–57, New York, NY, USA, 2021{\natexlab{a}}. Association for Computing Machinery.
\newblock ISBN 9781450382977.
\newblock \doi{10.1145/3437963.3441807}.
\newblock URL \url{https://doi.org/10.1145/3437963.3441807}.

\bibitem[Mittal et~al.(2021{\natexlab{b}})Mittal, Sachdeva, Agrawal, Agarwal, Kar, and Varma]{mittal2021eclare}
Anshul Mittal, Noveen Sachdeva, Sheshansh Agrawal, Sumeet Agarwal, Purushottam Kar, and Manik Varma.
\newblock Eclare: Extreme classification with label graph correlations.
\newblock In \emph{Proceedings of the Web Conference 2021}, WWW'21. ACM, April 2021{\natexlab{b}}.
\newblock \doi{10.1145/3442381.3449815}.
\newblock URL \url{http://dx.doi.org/10.1145/3442381.3449815}.

\bibitem[Mittal et~al.(2022{\natexlab{a}})Mittal, Dahiya, Malani, Ramaswamy, Kuruvilla, Ajmera, Chang, Agarwal, Kar, and Varma]{Mittal_2022}
Anshul Mittal, Kunal Dahiya, Shreya Malani, Janani Ramaswamy, Seba Kuruvilla, Jitendra Ajmera, Keng-Hao Chang, Sumeet Agarwal, Purushottam Kar, and Manik Varma.
\newblock Multi-modal extreme classification.
\newblock In \emph{2022 IEEE/CVF Conference on Computer Vision and Pattern Recognition (CVPR)}. IEEE, June 2022{\natexlab{a}}.
\newblock \doi{10.1109/cvpr52688.2022.01207}.
\newblock URL \url{http://dx.doi.org/10.1109/CVPR52688.2022.01207}.

\bibitem[Mittal et~al.(2022{\natexlab{b}})Mittal, Dahiya, Malani, Ramaswamy, Kuruvilla, Ajmera, Chang, Agarwal, Kar, and Varma]{mittal2022multi}
Anshul Mittal, Kunal Dahiya, Shreya Malani, Janani Ramaswamy, Seba Kuruvilla, Jitendra Ajmera, Keng-hao Chang, Sumeet Agarwal, Purushottam Kar, and Manik Varma.
\newblock Multi-modal extreme classification.
\newblock In \emph{Proceedings of the IEEE/CVF Conference on Computer Vision and Pattern Recognition}, pages 12393--12402, 2022{\natexlab{b}}.

\bibitem[Nam et~al.(2017)Nam, Loza~Menc\'{\i}a, Kim, and F\"{u}rnkranz]{MLC2Seq}
Jinseok Nam, Eneldo Loza~Menc\'{\i}a, Hyunwoo~J Kim, and Johannes F\"{u}rnkranz.
\newblock Maximizing subset accuracy with recurrent neural networks in multi-label classification.
\newblock In I.~Guyon, U.~V. Luxburg, S.~Bengio, H.~Wallach, R.~Fergus, S.~Vishwanathan, and R.~Garnett, editors, \emph{Advances in Neural Information Processing Systems 30}, pages 5413--5423. Curran Associates, Inc., 2017.

\bibitem[nan Chen and tien Lin(2012)]{cplst_chen}
Yao nan Chen and Hsuan tien Lin.
\newblock Feature-aware label space dimension reduction for multi-label classification.
\newblock In F.~Pereira, C.~J.~C. Burges, L.~Bottou, and K.~Q. Weinberger, editors, \emph{Advances in Neural Information Processing Systems 25}, pages 1529--1537. Curran Associates, Inc., 2012.

\bibitem[Naram et~al.(2022)Naram, Sinha, and Kumar]{naram22ReimannXML}
Jayadev Naram, Tanmay~Kumar Sinha, and Pawan Kumar.
\newblock A riemannian approach to extreme classification problems.
\newblock In \emph{Proceedings of the 5th Joint International Conference on Data Science \& Management of Data (9th ACM IKDD CODS and 27th COMAD)}, CODS-COMAD '22, page 54–62, New York, NY, USA, 2022. Association for Computing Machinery.
\newblock ISBN 9781450385824.
\newblock \doi{10.1145/3493700.3493714}.
\newblock URL \url{https://doi.org/10.1145/3493700.3493714}.

\bibitem[Pan et~al.(2018)Pan, Xu, Ruiz, Zhao, Pan, Sun, and Lu]{pan2018}
Junwei Pan, Jian Xu, Alfonso~Lobos Ruiz, Wenliang Zhao, Shengjun Pan, Yu~Sun, and Quan Lu.
\newblock Field-weighted factorization machines for click-through rate prediction in display advertising.
\newblock In \emph{Proceedings of the 2018 World Wide Web Conference on World Wide Web - WWW ’18}, WWW ’18. ACM Press, 2018.
\newblock \doi{10.1145/3178876.3186040}.
\newblock URL \url{http://dx.doi.org/10.1145/3178876.3186040}.

\bibitem[Pati et~al.(1993)Pati, Rezaiifar, and Krishnaprasad]{pati1993orthogonal}
Yagyensh~Chandra Pati, Ramin Rezaiifar, and Perinkulam~Sambamurthy Krishnaprasad.
\newblock Orthogonal matching pursuit: Recursive function approximation with applications to wavelet decomposition.
\newblock In \emph{Proceedings of 27th Asilomar conference on signals, systems and computers}, pages 40--44. IEEE, 1993.

\bibitem[Pavlovski et~al.(2023)Pavlovski, Ravindran, Gligorijevic, Agrawal, Stojkovic, Segura-Nunez, and Gligorijevic]{Pavlovski2023}
Martin Pavlovski, Srinath Ravindran, Djordje Gligorijevic, Shubham Agrawal, Ivan Stojkovic, Nelson Segura-Nunez, and Jelena Gligorijevic.
\newblock Extreme multi-label classification for ad targeting using factorization machines.
\newblock In \emph{Proceedings of the 29th ACM SIGKDD Conference on Knowledge Discovery and Data Mining}, KDD '23, page 4705–4716, New York, NY, USA, 2023. Association for Computing Machinery.
\newblock ISBN 9798400701030.
\newblock \doi{10.1145/3580305.3599822}.
\newblock URL \url{https://doi.org/10.1145/3580305.3599822}.

\bibitem[Peito and Han(2021)]{Peito_2021}
Joel Peito and Qiwei Han.
\newblock \emph{Incorporating Domain Knowledge into Health Recommender Systems Using Hyperbolic Embeddings}, page 130–141.
\newblock Springer International Publishing, 2021.
\newblock ISBN 9783030653514.
\newblock \doi{10.1007/978-3-030-65351-4_11}.
\newblock URL \url{http://dx.doi.org/10.1007/978-3-030-65351-4_11}.

\bibitem[Prabhu and Varma(2014{\natexlab{a}})]{FastXML}
Yashoteja Prabhu and Manik Varma.
\newblock Fastxml: a fast, accurate and stable tree-classifier for extreme multi-label learning.
\newblock In \emph{Proceedings of the 20th ACM SIGKDD International Conference on Knowledge Discovery and Data Mining}, KDD '14, page 263–272, New York, NY, USA, 2014{\natexlab{a}}. Association for Computing Machinery.
\newblock ISBN 9781450329569.
\newblock \doi{10.1145/2623330.2623651}.
\newblock URL \url{https://doi.org/10.1145/2623330.2623651}.

\bibitem[Prabhu and Varma(2014{\natexlab{b}})]{prabhu2014fastxml}
Yashoteja Prabhu and Manik Varma.
\newblock Fastxml: A fast, accurate and stable tree-classifier for extreme multi-label learning.
\newblock In \emph{Proceedings of the 20th ACM SIGKDD international conference on Knowledge discovery and data mining}, pages 263--272, 2014{\natexlab{b}}.

\bibitem[Prabhu et~al.(2018)Prabhu, Kag, Harsola, Agrawal, and Varma]{prabhu2018parabel}
Yashoteja Prabhu, Anil Kag, Shrutendra Harsola, Rahul Agrawal, and Manik Varma.
\newblock Parabel: Partitioned label trees for extreme classification with application to dynamic search advertising.
\newblock In \emph{Proceedings of the 2018 World Wide Web Conference}, pages 993--1002, 2018.

\bibitem[Prabhu et~al.(2020)Prabhu, Kusupati, Gupta, and Varma]{xreg_prabhu}
Yashoteja Prabhu, Aditya Kusupati, Nilesh Gupta, and Manik Varma.
\newblock Extreme regression for dynamic search advertising.
\newblock \emph{Proceedings of the 13th International Conference on Web Search and Data Mining}, Jan 2020.
\newblock \doi{10.1145/3336191.3371768}.

\bibitem[Qaraei and Babbar(2024)]{Qaraei2024}
Mohammadreza Qaraei and Rohit Babbar.
\newblock Meta-classifier free negative sampling for extreme multilabel classification.
\newblock \emph{Machine Learning}, 113\penalty0 (2):\penalty0 675--697, Feb 2024.
\newblock ISSN 1573-0565.
\newblock \doi{10.1007/s10994-023-06468-w}.
\newblock URL \url{https://doi.org/10.1007/s10994-023-06468-w}.

\bibitem[Ren et~al.(2021)Ren, Xiao, Chang, Huang, Li, Gupta, Chen, and Wang]{ren2021survey}
Pengzhen Ren, Yun Xiao, Xiaojun Chang, Po-Yao Huang, Zhihui Li, Brij~B. Gupta, Xiaojiang Chen, and Xin Wang.
\newblock A survey of deep active learning, 2021.

\bibitem[Rendle(2010)]{rendle2010}
Steffen Rendle.
\newblock Factorization machines.
\newblock In \emph{2010 IEEE International Conference on Data Mining}, pages 995--1000, 2010.
\newblock \doi{10.1109/ICDM.2010.127}.

\bibitem[Revanur et~al.(2021)Revanur, Kumar, and Sharma]{Revanur_2021}
Ambareesh Revanur, Vijay Kumar, and Deepthi Sharma.
\newblock Semi-supervised visual representation learning for fashion compatibility.
\newblock In \emph{Fifteenth ACM Conference on Recommender Systems}, RecSys ’21. ACM, September 2021.
\newblock \doi{10.1145/3460231.3474233}.
\newblock URL \url{http://dx.doi.org/10.1145/3460231.3474233}.

\bibitem[Saini et~al.(2021)Saini, Jain, Dave, Jiao, Singh, Zhang, and Varma]{saini2021galaxc}
Deepak Saini, Arnav~Kumar Jain, Kushal Dave, Jian Jiao, Amit Singh, Ruofei Zhang, and Manik Varma.
\newblock Galaxc: Graph neural networks with labelwise attention for extreme classification.
\newblock In \emph{Proceedings of the Web Conference 2021}, pages 3733--3744, 2021.

\bibitem[Schroff et~al.(2015)Schroff, Kalenichenko, and Philbin]{schroff2015facenet}
Florian Schroff, Dmitry Kalenichenko, and James Philbin.
\newblock Facenet: A unified embedding for face recognition and clustering.
\newblock In \emph{Proceedings of the IEEE conference on computer vision and pattern recognition}, pages 815--823, 2015.

\bibitem[Schultheis et~al.(2024)Schultheis, Wydmuch, Kotłowski, Babbar, and Dembczyński]{schultheis2024generalized}
Erik Schultheis, Marek Wydmuch, Wojciech Kotłowski, Rohit Babbar, and Krzysztof Dembczyński.
\newblock Generalized test utilities for long-tail performance in extreme multi-label classification, 2024.

\bibitem[Seshadri and Sundberg(1994)]{seshadri1994list}
Nambirajan Seshadri and C-EW Sundberg.
\newblock List viterbi decoding algorithms with applications.
\newblock \emph{IEEE transactions on communications}, 42\penalty0 (234):\penalty0 313--323, 1994.

\bibitem[Shrivastava and Li(2014)]{shrivastava2014asymmetric}
Anshumali Shrivastava and Ping Li.
\newblock Asymmetric lsh (alsh) for sublinear time maximum inner product search (mips), 2014.
\newblock URL \url{https://arxiv.org/abs/1405.5869}.

\bibitem[Siblini et~al.(2018)Siblini, Meyer, and Kuntz]{craftml_siblini}
Wissam Siblini, Frank Meyer, and Pascale Kuntz.
\newblock Craftml, an efficient clustering-based random forest for extreme multi-label learning.
\newblock In \emph{ICML}, 2018.

\bibitem[Sprechmann et~al.(2013)Sprechmann, Litman, Yakar, Bronstein, and Sapiro]{admm_sprechmann}
Pablo Sprechmann, Roee Litman, Tal~Ben Yakar, Alex Bronstein, and Guillermo Sapiro.
\newblock Efficient supervised sparse analysis and synthesis operators.
\newblock In \emph{Proceedings of the 26th International Conference on Neural Information Processing Systems - Volume 1}, NIPS'13, pages 908--916, USA, 2013. Curran Associates Inc.

\bibitem[Tagami(2017)]{AnnexML}
Yukihiro Tagami.
\newblock Annexml: Approximate nearest neighbor search for extreme multi-label classification.
\newblock In \emph{Proceedings of the 23rd ACM SIGKDD International Conference on Knowledge Discovery and Data Mining}, KDD '17, page 455–464, New York, NY, USA, 2017. Association for Computing Machinery.
\newblock ISBN 9781450348874.
\newblock \doi{10.1145/3097983.3097987}.
\newblock URL \url{https://doi.org/10.1145/3097983.3097987}.

\bibitem[Tai and Lin(2012)]{plst_tai}
Farbound Tai and Hsuan-Tien Lin.
\newblock Multilabel classification with principal label space transformation.
\newblock \emph{Neural Comput.}, 24\penalty0 (9):\penalty0 2508–2542, September 2012.
\newblock ISSN 0899-7667.
\newblock \doi{10.1162/NECO-a-00320}.

\bibitem[Tan et~al.(2019)Tan, Vasileva, Saenko, and Plummer]{tan2019learning}
Reuben Tan, Mariya~I. Vasileva, Kate Saenko, and Bryan~A. Plummer.
\newblock Learning similarity conditions without explicit supervision, 2019.

\bibitem[Valdeira et~al.(2023)Valdeira, Racković, Danalachi, Han, and Soares]{valdeira2023extreme}
Filipa Valdeira, Stevo Racković, Valeria Danalachi, Qiwei Han, and Cláudia Soares.
\newblock Extreme multilabel classification for specialist doctor recommendation with implicit feedback and limited patient metadata, 2023.

\bibitem[van~den Oord et~al.(2019)van~den Oord, Li, and Vinyals]{oord2019representation}
Aaron van~den Oord, Yazhe Li, and Oriol Vinyals.
\newblock Representation learning with contrastive predictive coding, 2019.

\bibitem[Vaswani et~al.(2017)Vaswani, Shazeer, Parmar, Uszkoreit, Jones, Gomez, Kaiser, and Polosukhin]{AttentionXML}
Ashish Vaswani, Noam Shazeer, Niki Parmar, Jakob Uszkoreit, Llion Jones, Aidan~N. Gomez, Lukasz Kaiser, and Illia Polosukhin.
\newblock Attention is all you need.
\newblock \emph{CoRR}, abs/1706.03762, 2017.
\newblock URL \url{http://arxiv.org/abs/1706.03762}.

\bibitem[Velioglu et~al.(2024)Velioglu, Chan, and Hammer]{velioglu2024fashionfail}
Riza Velioglu, Robin Chan, and Barbara Hammer.
\newblock Fashionfail: Addressing failure cases in fashion object detection and segmentation.
\newblock In \emph{2024 International Joint Conference on Neural Networks (IJCNN)}, page 1–8. IEEE, June 2024.
\newblock \doi{10.1109/ijcnn60899.2024.10651287}.
\newblock URL \url{http://dx.doi.org/10.1109/IJCNN60899.2024.10651287}.

\bibitem[Wang et~al.(2022{\natexlab{a}})Wang, Thakur, Reimers, and Gurevych]{wang-etal-2022-gpl}
Kexin Wang, Nandan Thakur, Nils Reimers, and Iryna Gurevych.
\newblock {GPL}: Generative pseudo labeling for unsupervised domain adaptation of dense retrieval.
\newblock In Marine Carpuat, Marie-Catherine de~Marneffe, and Ivan~Vladimir Meza~Ruiz, editors, \emph{Proceedings of the 2022 Conference of the North American Chapter of the Association for Computational Linguistics: Human Language Technologies}, pages 2345--2360, Seattle, United States, July 2022{\natexlab{a}}. Association for Computational Linguistics.
\newblock \doi{10.18653/v1/2022.naacl-main.168}.
\newblock URL \url{https://aclanthology.org/2022.naacl-main.168}.

\bibitem[Wang et~al.(2022{\natexlab{b}})Wang, Xu, Sun, Hu, Tao, Geng, and Jiang]{wang-etal-2022-promda}
Yufei Wang, Can Xu, Qingfeng Sun, Huang Hu, Chongyang Tao, Xiubo Geng, and Daxin Jiang.
\newblock {P}rom{DA}: Prompt-based data augmentation for low-resource {NLU} tasks.
\newblock In Smaranda Muresan, Preslav Nakov, and Aline Villavicencio, editors, \emph{Proceedings of the 60th Annual Meeting of the Association for Computational Linguistics (Volume 1: Long Papers)}, pages 4242--4255, Dublin, Ireland, May 2022{\natexlab{b}}. Association for Computational Linguistics.
\newblock \doi{10.18653/v1/2022.acl-long.292}.
\newblock URL \url{https://aclanthology.org/2022.acl-long.292}.

\bibitem[Weston et~al.(2013)Weston, Makadia, and Yee]{weston2013label}
Jason Weston, Ameesh Makadia, and Hector Yee.
\newblock Label partitioning for sublinear ranking.
\newblock In \emph{International conference on machine learning}, pages 181--189. PMLR, 2013.

\bibitem[Wu et~al.(2020)Wu, Huang, Liu, Wang, and Lin]{wu2020distribution}
Tong Wu, Qingqiu Huang, Ziwei Liu, Yu~Wang, and Dahua Lin.
\newblock Distribution-balanced loss for multi-label classification in long-tailed datasets.
\newblock \emph{CoRR}, abs/2007.09654, 2020.
\newblock URL \url{https://arxiv.org/abs/2007.09654}.

\bibitem[Xiong et~al.(2020)Xiong, Xiong, Li, Tang, Liu, Bennett, Ahmed, and Overwijk]{xiong2020approximate}
Lee Xiong, Chenyan Xiong, Ye~Li, Kwok-Fung Tang, Jialin Liu, Paul Bennett, Junaid Ahmed, and Arnold Overwijk.
\newblock Approximate nearest neighbor negative contrastive learning for dense text retrieval, 2020.

\bibitem[Xu et~al.(2016)Xu, Tao, and Xu]{reml_chang}
Chang Xu, Dacheng Tao, and Chao Xu.
\newblock Robust extreme multi-label learning.
\newblock In \emph{Proceedings of the 22nd ACM SIGKDD International Conference on Knowledge Discovery and Data Mining}, KDD ’16, page 1275–1284, New York, NY, USA, 2016. Association for Computing Machinery.
\newblock ISBN 9781450342322.
\newblock \doi{10.1145/2939672.2939798}.

\bibitem[Xu et~al.(2023)Xu, Xiao, Liu, Lu, Jing, and Yu]{xu2023labelspecific}
Pengyu Xu, Lin Xiao, Bing Liu, Sijin Lu, Liping Jing, and Jian Yu.
\newblock Label-specific feature augmentation for long-tailed multi-label text classification.
\newblock In \emph{Proceedings of the Thirty-Seventh AAAI Conference on Artificial Intelligence and Thirty-Fifth Conference on Innovative Applications of Artificial Intelligence and Thirteenth Symposium on Educational Advances in Artificial Intelligence}, AAAI'23/IAAI'23/EAAI'23. AAAI Press, 2023.
\newblock ISBN 978-1-57735-880-0.
\newblock \doi{10.1609/aaai.v37i9.26259}.
\newblock URL \url{https://doi.org/10.1609/aaai.v37i9.26259}.

\bibitem[Xu et~al.(2024)Xu, Song, Li, Lu, Jing, and Yu]{xu2024tamingprompt}
Pengyu Xu, Mingyang Song, Ziyi Li, Sijin Lu, Liping Jing, and Jian Yu.
\newblock Taming prompt-based data augmentation for long-tailed extreme multi-label text classification.
\newblock In \emph{ICASSP 2024 - 2024 IEEE International Conference on Acoustics, Speech and Signal Processing (ICASSP)}, pages 9981--9985, 2024.
\newblock \doi{10.1109/ICASSP48485.2024.10446315}.

\bibitem[Yang et~al.(2019)Yang, Dai, Yang, Carbonell, Salakhutdinov, and Le]{yang2019xlnet}
Zhilin Yang, Zihang Dai, Yiming Yang, Jaime Carbonell, Russ~R Salakhutdinov, and Quoc~V Le.
\newblock Xlnet: Generalized autoregressive pretraining for language understanding.
\newblock \emph{Advances in neural information processing systems}, 32, 2019.

\bibitem[Ye et~al.(2020)Ye, Chen, Wang, and Davison]{ye2020pretrained}
Hui Ye, Zhiyu Chen, Da-Han Wang, and Brian Davison.
\newblock Pretrained generalized autoregressive model with adaptive probabilistic label clusters for extreme multi-label text classification.
\newblock In \emph{International Conference on Machine Learning}, pages 10809--10819. PMLR, 2020.

\bibitem[Ye et~al.(2024)Ye, Sunderraman, and Ji]{matchxml}
Hui Ye, Rajshekhar Sunderraman, and Shihao Ji.
\newblock Matchxml: An efficient text-label matching framework for extreme multi-label text classification.
\newblock \emph{IEEE Transactions on Knowledge and Data Engineering}, pages 1--13, 2024.
\newblock \doi{10.1109/TKDE.2024.3374750}.

\bibitem[Yen et~al.(2017)Yen, Huang, Dai, Ravikumar, Dhillon, and Xing]{PPD_Sparse}
Ian~E.H. Yen, Xiangru Huang, Wei Dai, Pradeep Ravikumar, Inderjit Dhillon, and Eric Xing.
\newblock Ppdsparse: A parallel primal-dual sparse method for extreme classification.
\newblock In \emph{Proceedings of the 23rd ACM SIGKDD International Conference on Knowledge Discovery and Data Mining}, KDD ’17, page 545–553, New York, NY, USA, 2017. Association for Computing Machinery.
\newblock ISBN 9781450348874.
\newblock \doi{10.1145/3097983.3098083}.

\bibitem[Yen et~al.(2016{\natexlab{a}})Yen, Huang, Ravikumar, Zhong, and Dhillon]{pmlr-v48-yenb16}
Ian En-Hsu Yen, Xiangru Huang, Pradeep Ravikumar, Kai Zhong, and Inderjit Dhillon.
\newblock Pd-sparse : A primal and dual sparse approach to extreme multiclass and multilabel classification.
\newblock In Maria~Florina Balcan and Kilian~Q. Weinberger, editors, \emph{Proceedings of The 33rd International Conference on Machine Learning}, volume~48 of \emph{Proceedings of Machine Learning Research}, pages 3069--3077, New York, New York, USA, 20--22 Jun 2016{\natexlab{a}}. PMLR.
\newblock URL \url{https://proceedings.mlr.press/v48/yenb16.html}.

\bibitem[Yen et~al.(2016{\natexlab{b}})Yen, Huang, Ravikumar, Zhong, and Dhillon]{pdsparse_ian}
Ian En-Hsu Yen, Xiangru Huang, Pradeep Ravikumar, Kai Zhong, and Inderjit~S. Dhillon.
\newblock Pd-sparse : A primal and dual sparse approach to extreme multiclass and multilabel classification.
\newblock In \emph{ICML}, 2016{\natexlab{b}}.

\bibitem[Yih et~al.(2011)Yih, Toutanova, Platt, and Meek]{dssm_yih}
Wen-tau Yih, Kristina Toutanova, John~C. Platt, and Christopher Meek.
\newblock Learning discriminative projections for text similarity measures.
\newblock In \emph{Proceedings of the Fifteenth Conference on Computational Natural Language Learning}, pages 247--256, Portland, Oregon, USA, June 2011. Association for Computational Linguistics.

\bibitem[You et~al.(2019)You, Zhang, Wang, Dai, Mamitsuka, and Zhu]{you2019attentionxml}
Ronghui You, Zihan Zhang, Ziye Wang, Suyang Dai, Hiroshi Mamitsuka, and Shanfeng Zhu.
\newblock Attentionxml: Label tree-based attention-aware deep model for high-performance extreme multi-label text classification.
\newblock \emph{Advances in Neural Information Processing Systems}, 32, 2019.

\bibitem[Yu et~al.(2013)Yu, Jain, and Dhillon]{leml_yu}
Hsiang{-}Fu Yu, Prateek Jain, and Inderjit~S. Dhillon.
\newblock Large-scale multi-label learning with missing labels.
\newblock \emph{CoRR}, abs/1307.5101, 2013.

\bibitem[Yu et~al.(2019)Yu, Zhong, Dhillon, Wang, and Yang]{Yu2019}
Hsiang-Fu Yu, Kai Zhong, Inderjit~S. Dhillon, Wei-Cheng Wang, and Yiming Yang.
\newblock X-bert: extreme multi-label text classification using bidirectional encoder representations from transformers.
\newblock In \emph{NeurIPS 2019 Workshop on Science Meets Engineering of Deep Learning}, 2019.

\bibitem[Zhang et~al.(2022)Zhang, Liu, Chen, Lin, Wang, and Wang]{zhang2022adam}
Jiaxin Zhang, Jie Liu, Shaowei Chen, Shaoxin Lin, Bingquan Wang, and Shanpeng Wang.
\newblock Adam: An attentional data augmentation method for extreme multi-label text classification.
\newblock In \emph{Advances in Knowledge Discovery and Data Mining: 26th Pacific-Asia Conference, PAKDD 2022, Chengdu, China, May 16–19, 2022, Proceedings, Part I}, page 131–142, Berlin, Heidelberg, 2022. Springer-Verlag.
\newblock ISBN 978-3-031-05932-2.
\newblock \doi{10.1007/978-3-031-05933-9_11}.
\newblock URL \url{https://doi.org/10.1007/978-3-031-05933-9_11}.

\bibitem[Zhang et~al.(2021)Zhang, Chang, Yu, and Dhillon]{zhang2021fast}
Jiong Zhang, Wei-Cheng Chang, Hsiang-Fu Yu, and Inderjit Dhillon.
\newblock Fast multi-resolution transformer fine-tuning for extreme multi-label text classification.
\newblock In M.~Ranzato, A.~Beygelzimer, Y.~Dauphin, P.S. Liang, and J.~Wortman Vaughan, editors, \emph{Advances in Neural Information Processing Systems}, volume~34, pages 7267--7280. Curran Associates, Inc., 2021.
\newblock URL \url{https://proceedings.neurips.cc/paper_files/paper/2021/file/3bbca1d243b01b47c2bf42b29a8b265c-Paper.pdf}.

\bibitem[Zhang et~al.(2023)Zhang, Wang, Yang, Yu, Vu, and Lei]{zhang-etal-2023-long}
Ruohong Zhang, Yau-Shian Wang, Yiming Yang, Donghan Yu, Tom Vu, and Likun Lei.
\newblock Long-tailed extreme multi-label text classification by the retrieval of generated pseudo label descriptions.
\newblock In Andreas Vlachos and Isabelle Augenstein, editors, \emph{Findings of the Association for Computational Linguistics: EACL 2023}, pages 1092--1106, Dubrovnik, Croatia, May 2023. Association for Computational Linguistics.
\newblock \doi{10.18653/v1/2023.findings-eacl.81}.
\newblock URL \url{https://aclanthology.org/2023.findings-eacl.81}.

\bibitem[Zhou et~al.(2022)Zhou, Zheng, Tang, Jian, and Yang]{zhou-etal-2022-flipda}
Jing Zhou, Yanan Zheng, Jie Tang, Li~Jian, and Zhilin Yang.
\newblock {F}lip{DA}: Effective and robust data augmentation for few-shot learning.
\newblock In Smaranda Muresan, Preslav Nakov, and Aline Villavicencio, editors, \emph{Proceedings of the 60th Annual Meeting of the Association for Computational Linguistics (Volume 1: Long Papers)}, pages 8646--8665, Dublin, Ireland, May 2022. Association for Computational Linguistics.
\newblock \doi{10.18653/v1/2022.acl-long.592}.
\newblock URL \url{https://aclanthology.org/2022.acl-long.592}.

\bibitem[Zhu and Zamani(2024)]{zhu2024icxml}
Yaxin Zhu and Hamed Zamani.
\newblock Icxml: An in-context learning framework for zero-shot extreme multi-label classification, 2024.

\bibitem[Zhuang et~al.(2021)Zhuang, Wayne, Ya, and Jun]{zhuang-etal-2021-robustly}
Liu Zhuang, Lin Wayne, Shi Ya, and Zhao Jun.
\newblock A robustly optimized {BERT} pre-training approach with post-training.
\newblock In Sheng Li, Maosong Sun, Yang Liu, Hua Wu, Kang Liu, Wanxiang Che, Shizhu He, and Gaoqi Rao, editors, \emph{Proceedings of the 20th Chinese National Conference on Computational Linguistics}, pages 1218--1227, Huhhot, China, August 2021. Chinese Information Processing Society of China.
\newblock URL \url{https://aclanthology.org/2021.ccl-1.108}.

\bibitem[Zubiaga(2012)]{zubiaga2012enhancing}
Arkaitz Zubiaga.
\newblock Enhancing navigation on wikipedia with social tags.
\newblock \emph{arXiv preprint arXiv:1202.5469}, 2012.

\end{thebibliography}

\end{document}